\documentclass[acmsmall,natbib=false]{acmart}

\usepackage{xcolor}
\usepackage{xargs}
\usepackage{bbm}
\usepackage{optidef}
\usepackage{booktabs}
\usepackage{enumitem}
\usepackage{lscape}
\usepackage{longtable}

\usepackage[colorinlistoftodos,prependcaption,textsize=tiny]{todonotes}
\newcommandx{\mattialightnote}[2][1=]{\todo[linecolor=red,backgroundcolor=white,bordercolor=red,#1]{#2}}
\newcommandx{\mattianote}[2][1=]{\todo[linecolor=red,backgroundcolor=red!25,bordercolor=red,#1]{#2}}
\newcommandx{\bluenote}[2][1=]{\todo[linecolor=blue,backgroundcolor=blue!25,bordercolor=blue,#1]{#2}}
\newcommandx{\foscanote}[2][1=]{\todo[linecolor=green,backgroundcolor=green!25,bordercolor=green,#1]{FOSCA: #2}}
\newcommandx{\robernote}[2][1=]{\todo[linecolor=yellow,backgroundcolor=yellow!25,bordercolor=yellow,#1]{#2}}
\newcommandx{\claranote}[2][1=]{\todo[linecolor=purple,backgroundcolor=purple!25,bordercolor=purple,#1]{#2}}
\newcommandx{\andreanote}[2][1=]{\todo[linecolor=teal,backgroundcolor=teal!25,bordercolor=teal,#1]{#2}}
\newcommandx{\Andreanote}[2][1=]{\todo[linecolor=TealBlue,backgroundcolor=TealBlue!25,bordercolor=TealBlue,#1]{#2}}

\usepackage{dsfont}
\usepackage{makecell}

\newcommand{\Rho}{\mathrm{P}}

\definecolor{myred}{RGB}{188, 73, 67}
\definecolor{myblue}{RGB}{45, 98, 118}
\definecolor{mygreen}{RGB}{52, 146, 64}

\usepackage{multirow}
\usepackage{rotating}
\usepackage{geometry}

\theoremstyle{definition}

\usepackage{caption}
\usepackage{subcaption}

\usepackage[most]{tcolorbox}
\usetikzlibrary{arrows,positioning} 
\usepackage{etoolbox} % for \ifnumcomp
\usetikzlibrary{shapes,calc,decorations.pathreplacing,calligraphy}
\usepackage{dingbat}
\usepackage{listofitems} 
\usepackage{tikzsymbols}
\tikzset{>=latex} % for LaTeX arrow head
\colorlet{myred}{red!80!black}
\colorlet{myblue}{blue!80!black}
\colorlet{mygreen}{green!60!black}
\colorlet{mydarkred}{myred!40!black}
\colorlet{mydarkblue}{myblue!40!black}
\colorlet{mydarkgreen}{mygreen!40!black}

\tikzset{
    >=stealth',
    punkt/.style={
           rectangle,
           rounded corners,
           draw=black, very thick,
           text width=3.5em,
           minimum height=1em,
           text centered},
    pil/.style={
           ->,
           thick,
           shorten <=1pt,
           shorten >=1pt,}
}
\tikzstyle{node}=[circle,draw=myblue,minimum size=15,inner sep=0.2,outer sep=0.3]
\tikzstyle{connect}=[->,thick,mydarkblue,shorten >=1]
\tikzset{ % node styles, numbered for easy mapping with \nstyle
  node 1/.style={node,black,draw=black,fill=white},
  node 2/.style={node,black,draw=black,fill=white},
  node 3/.style={node,mydarkred,draw=myred,fill=myred!20},
  node 4/.style={node,mydarkblue,draw=myblue,fill=myblue!20},
  node 5/.style={node,black,draw=black,fill=white},
}
\def\nstyle{int(\lay<\Nnodlen?min(2,\lay):3)} % map layer number onto 1, 2, or 3

\AtBeginDocument{%
  \providecommand\BibTeX{{%
    \normalfont B\kern-0.5em{\scshape i\kern-0.25em b}\kern-0.8em\TeX}}}

\usepackage{array}
\usepackage{varwidth} % must be loaded after array

\acmJournal{JACM}
\acmYear{2024}

\usepackage[natbib,style=acmnumeric,giveninits=true,sorting=nty,sortcites=true]{biblatex}
\begin{document}

%\title{AI, Meet Human: A Survey on Hybrid Systems }
%\title{AI, Meet Human: A Survey on Hybrid Systems for Decision-Making Processes}
\title{AI, Meet Human: Learning Paradigms for Hybrid Decision-Making Systems}

\author{Clara Punzi}
\email{clara.punzi@sns.it}
\orcid{??}
\affiliation{%
  \institution{Scuola Normale Superiore}
  \streetaddress{Piazza dei Cavalieri, 7}
  \city{Pisa}
  \country{Italy}
  \postcode{56126}
}

\author{Roberto Pellungrini}
\email{roberto.pellungrini@sns.it}
\orcid{??}
\affiliation{%
  \institution{Scuola Normale Superiore}
  \streetaddress{Piazza dei Cavalieri, 7}
  \city{Pisa}
  \country{Italy}
  \postcode{56126}
}

\author{Mattia Setzu}
\email{mattia.setzu@unipi.it}
\orcid{??}
\affiliation{%
  \institution{University of Pisa}
  \streetaddress{Largo Bruno Pontecorvo, 3}
  \city{Pisa}
  \country{Italy}
  \postcode{56127}
}

\author{Fosca Giannotti}
\email{fosca.giannotti@sns.it}
\orcid{??}
\affiliation{%
  \institution{Scuola Normale Superiore}
  \streetaddress{Piazza dei Cavalieri, 7}
  \city{Pisa}
  \country{Italy}
  \postcode{56126}
}

\author{Dino Pedreschi}
\email{dino.pedreschi@unipi.it}
\orcid{??}
\affiliation{%
  \institution{University of Pisa}
  \streetaddress{Largo Bruno Pontecorvo, 3}
  \city{Pisa}
  \country{Italy}
  \postcode{56127}
}
\renewcommand{\shortauthors}{Punzi, Pellungrini, Setzu, et al.}

\begin{abstract}
Everyday we increasingly rely on machine learning models to automate and support high-stake tasks and decisions. This growing presence means that humans are now constantly interacting with machine learning-based systems, training and using models everyday. Several different techniques in computer science literature account for the human interaction with machine learning systems, but their classification is sparse and the goals varied. This survey proposes a taxonomy of Hybrid Decision-Making Systems, providing both a conceptual and technical framework for understanding how current computer science literature models interaction between humans and machines.
\end{abstract}

\begin{CCSXML}
<ccs2012>
   <concept>
       <concept_id>10010147.10010257.10010258</concept_id>
       <concept_desc>Computing methodologies~Learning paradigms</concept_desc>
       <concept_significance>500</concept_significance>
    </concept>
    <concept>
       <concept_id>10003120.10003121.10003126</concept_id>
       <concept_desc>Human-centered computing~HCI theory, concepts and models</concept_desc>
       <concept_significance>300</concept_significance>
    </concept>
 </ccs2012>
\end{CCSXML}

\ccsdesc[500]{Computing methodologies~Learning paradigms}
\ccsdesc[300]{Human-centered computing~HCI theory, concepts and models}

%    <concept>
%        <concept_id>10010147.10010257.10010258</concept_id>
%        <concept_desc>Computing methodologies~Learning paradigms</concept_desc>
%        <concept_significance>500</concept_significance>
%        </concept>
%    <concept>
%        <concept_id>10003120.10003121.10003126</concept_id>
%        <concept_desc>Human-centered computing~HCI theory, concepts and models</concept_desc>
%        <concept_significance>300</concept_significance>
%        </concept>
%    <concept>
%        <concept_id>10002951.10003227.10003241</concept_id>
%        <concept_desc>Information systems~Decision support systems</concept_desc>
%        <concept_significance>300</concept_significance>
%        </concept>
%  </ccs2012>

% \ccsdesc[500]{Computing methodologies~Learning paradigms}
% \ccsdesc[300]{Human-centered computing~HCI theory, concepts and models}
% \ccsdesc[300]{Information systems~Decision support systems}

%%
%% Keywords. The author(s) should pick words that accurately describe the work being presented. Separate the keywords with commas.
\keywords{hybrid systems, hybrid decision-making, cooperative AI}


\maketitle
\begin{refsection}
\section{Introduction}
    \label{sec:introduction}
    % Candidati titoli per paradigma 1:
    % \begin{itemize}
    %     \item human overseers
    %     \item human as overseers
    %     \item overseeing the machine
    %     \item machine learning-based decision-making
    %     \item human oversight of machines
    %     \item
    % \end{itemize}
    Advances in Artificial Intelligence (AI) systems have spurred a widespread use in a variety of domains and applications, achieving surprising performance on tasks previously thought to be out of reach for AI. 
    %Their effectiveness has grown exponentially over the past decades, and now AI systems are able to achieve surprising performances on tasks previously thought to be out of reach for artificial systems.
    \textcolor{black}{In high-stakes fields like medicine~\cite{rajpurkar2022ai}, finance~\cite{goodell2021artificial}, and law~\cite{surden2019artificial}, AI models now serve as decision-support systems for human experts. However, these systems are still susceptible to distrust~\cite{dietvorst2015algorithm}, misuse and disuse~\cite{lee2004trust} by human actors,
    with the risk of incurring in undesired patterns of user behavior, such as algorithmic aversion~\cite{dietvorst2015algorithm} and over-reliance~\cite{logg2019algorithm}.
    Indeed, selecting appropriate AI model architectures and learning paradigms is crucial not only for system accuracy, but also to ensure trustworthiness and mitigate unintended discriminatory outcomes.} %Building trust is essential for effective human-AI collaboration.
    
    \textcolor{black}{
    There has been a flourishing of methods that seek to prevent AI from supplanting human activities, instead supporting human-AI collaboration or ``human-in-the-loop'' approaches, such as explainable AI, which raises awareness of the decision-making process, and methods that focus on managing the uncertainty of the AI system.
    % Overall, the body of literature is huge. Therefore, we choose to provide a reading that facilitates the identification of the many interaction types and their components, while also offering a detailed description of certain interaction types coming from the most recent studies. 
    Such approaches foster synergistic and heterogeneous settings in which humans and AI jointly address given tasks, seeking to improve decision quality, reliability, transparency, and fairness. For instance, in medicine, an AI trained to detect diseases from imaging may defer certain cases to clinicians based on the uncertainty of the cases. In content moderation, AI may flag harmful content like misinformation or hate speech for human review, as humans excel at understanding context and nuance. Such abstention mechanisms or tiered structures may possibly improve hybrid system performance over standalone AI.}
    \textcolor{black}{Whether AI systems and humans carry out tasks independently or collaboratively, there exists a significant disparity between them, as they have fundamentally different ways of understanding, reasoning and representing the world. Leveraging both their complementary strengths could greatly benefit decision-making processes, as seen in the example of content moderation.}
    %Whether AI systems are standalone or collaborative, we find an inherent mismatch between the two parties: 
    %humans and AI systems have fundamentally different ways of understanding and representing the world. Therefore, human and AI reasoning are fundamentally different, and leveraging both their complementary strengths could greatly benefit decision-making processes, \textcolor{black}{as already noted in the example of content moderation.}
    %and their ways and abilities to reason about it are also drastically different as a consequence. 
    %This observation by itself does not have a positive or negative connotation, rather, it opens the doors for collaborative systems where both humans and AI systems are appropriately exploited to the fullest \textcolor{red}{, and where the collaborative system has greater abilities than the sum of its human and AI parts.}
    % \textcolor{red}{This observation does not have a positive or negative connotation, rather, it poses the question of how can we build AI systems where the collaboration with humans is leveraged to improve the overall usefulness of the whole system.}
    % \robernote{vedi se così ti piace}
    % to best exploit the strengths of both.
    Due to the heterogeneity of the agents involved, some being human, others being artificial, these
    % decision-making
    systems are often referred to as 
    %\emph{Hybrid Decision-Making Systems} or 
    \emph{Hybrid Systems} (HS).
    %for short. 
    In this work, we use ``Hybrid Decision-Making Systems'' and ``Hybrid Systems'' interchangeably; nevertheless, it must be noted that decision-making is indeed one specific purpose for which a HS are generally developed.
%    In the following, we are going to provide an overview of different learning paradigms for hybrid systems, their evolution over time, a taxonomy to properly categorize existing hybrid systems, and their strengths and weaknesses.

    %\paragraph{Two agents, two steps: hybrid systems.} 
    \textcolor{black}{From a macro perspective (a formal description will be provided in Section \ref{sec:generalModels}), a Hybrid System (HS) is defined by the task of interest, the agents involved, their computational steps, and their joint behaviour. 
    %In the following, we also use the term \emph{system} to denote the whole hybrid system, and \emph{machine (sub)system} and \emph{human (sub)system} to refer to the machine and human component of the system, respectively.
    Two archetypal agents\footnote{We define ``agents'' as entities inside a system that may act upon it, rather than as ``entities with agency'', meaning we do not inherently ascribe intents to agents.} of different nature are identified: the \textit{human}, endowed with peculiar reasoning abilities, domain, and commonsense knowledge, and the \textit{machine}, endowed with high computing power. The two agents are best suited to solve different tasks, with machines excelling in computation-heavy tasks and humans excelling in activities } 
    %In greater detail, within hybrid systems we 
    %identify two archetypal agents of different nature: the \emph{human}, which is endowed with peculiar reasoning abilities, domain, and commonsense knowledge; and the \emph{machine}, which is instead endowed with high computing power.
    The two agents,
    are best suited to solve different tasks.
    % Unlike machines, humans excel in tasks requiring (potentially) complex reasoning, domain knowledge~\cite{}, commonsense knowledge~\cite{}, contextualization~\cite{}, and general human touch~\cite{}.
    % On the other hand, machines excel in computation-heavy tasks and are uniquely capable of performing such tasks at scale.
    %Unlike humans, machines often fail in tasks 
    requiring lateral reasoning, domain knowledge~\cite{DBLP:conf/emnlp/CiosiciCLHFW21}, commonsense~\cite{DBLP:conf/acl/ChenSFC0X23}, contextualization, and general human touch~\cite{ferguson2020high}.
    %On the other hand, machines excel in computation-heavy tasks and are uniquely capable of performing such tasks at scale.
    \textcolor{black}{
    Thus, it is unsurprising that humans and machines often disagree ~\cite{DBLP:journals/corr/abs-1808-09123,kempt2022m}, and even when they concur, the outcomes may stem from contradictory arguments ~\cite{DBLP:conf/ijcai/RossHD17}.
    Traditional (standalone) AI models }
    %Further exacerbating this, traditional machine evaluation metrics 
    are often blind to this disagreement \textcolor{black}{and fail to capture it}, since they are designed to simply target the performance of the machine, rather than its internal reasoning.  
    %Whenever an agent attempts to solve a task, they are performing a \emph{computational step}, also known as \emph{computational turn}~\cite{hildebrandt2013privacy}, which is a \emph{human step} if the agent is human, and a \emph{machine step} if the agent is a machine.
    % Overall, we can then define a Hybrid System (HS) by specifying the task of interest, the agents involved, their computational steps, and their joint behaviour.
    % %, defines a \textit{Hybrid
    % % Decision-Making
    % %System}
    % % (HDMS).
    % %(HS).
    %In the following, we use the term \emph{system} to identify the whole hybrid system, and \emph{machine (sub)system} and \emph{human (sub)system} to refer to the machine and human component of the system, respectively.
    In contrast, hybrid systems aim to optimize overall task performance metrics, such as accuracy, by effectively combining human and machine sub-systems. 
    
    Pairing humans and machines has advantages: observing machine behavior helps humans develop a \emph{mental model}, i.e. an internal representation of how they perceive the machine, its capabilities and its limitations \cite{DBLP:conf/aaai/BansalW18}. If the machine can also observe and learn from human feedback, the system becomes bidirectional, enhancing mutual understanding. Such systems foster trust, increasing human reliance and engagement with the machine \cite{DBLP:conf/hcomp/BansalNKLWH19}, and encourage alignment through user control \cite{dietvorst2018overcoming}. Machines, meanwhile, can better leverage unique human abilities that are difficult for them to learn \cite{DBLP:conf/aies/TesoK19}. 
    \textcolor{black}{Notwithstanding, these collaborative frameworks also face novel challenges, such as defining }
    %are plagued by the same mistrust, misuse, and disuse issues of stand-alone AI systems, on top of novel and unique issues. Indeed, collaborative systems have to define 
    an effective communication channel between the AI system and the human decision-makers, and the capability to properly combine the two.
    Moreover, integrating coarse-grained and fine-grained control of AI systems has proven to be a non-trivial task in itself, indicating that steering AI systems to desired complex behaviors~\cite{DBLP:conf/iclr/Saparov023}, preventing them from learning spurious and undesired correlations~\cite{DBLP:journals/tkdd/KaufmanRPS12}, and ultimately aligning them with human values~\cite{DBLP:conf/nips/SolaimanD21} are far from a solved problem.
    
    \begin{tcolorbox}[colback=blue!5!white,colframe=blue!60!white,title=Example of a General Hybrid-Decision-Making System,fonttitle=\small]
    \small
        For clarity, we will use a running example to guide us during the survey.
        A dermatologist diagnoses skin lesions, diseases, and cancers, e.g., \textit{melanoma}, 
        %making them a high-stake decision maker.
        Detecting melanomas is a binary classification task in which imaging of a skin region is used to detect potentially malignant melanomas.
        Such a screening task lends itself particularly well to automation, especially in large population screenings, 
        %or when access to health professionals is limited.
        Advancements in AI-based medical imaging have produced models able to often achieve a good enough performance to properly aid health professionals in the field.
        However, the complexity and the high-stakes nature of the task still demands the dermatologist to play a role.
        Therefore, what is needed is for the physician to use the AI-based system to enhance their decision-making capabilities in a hybrid decision-making system.
    \end{tcolorbox}
    %
    % \paragraph{Hybrid systems and Decision-Making}
    %     \andreanote{@ANDREA}
    %     Here, Andrea gives an overview of hybrid systems form the congitive perspective\\

    \textcolor{black}{This survey provides a reasoned survey of the vast literature on hybrid systems and formalises the mechanisms of the machine learning models underpinning the human-AI collaboration, which mostly focusses on classification tasks with limited attention to regression and clustering. Therefore, we discuss hybrid system learning paradigms, their evolution over time, our proposed taxonomy, and their strengths and shortcomings.
   % As already mentioned, this study focusses on hybrid decision-making systems, namely those where humans and machines jointly solve a task, with a collaboration ranging from a mere oversight to a fully-fledged interaction.
% i.e., AI systems where the final decision of the human is supported by an AI systems, and there might be different forms of interactions, from mere observation of the recommendation till a synergistic collaboration.  our goal is to disectangle various form of this interaction. d
% that allow interaction between human and machines, and
% as such we do not specifically address systems where humans are not active agents nor they are pure teaching agents within an environment, such as knowledge injection systems, autonomous agents, reinforcement learning systems, and conversational XAI algorithms, where the interaction is limited to the exploration of the explanations.~\cite{DBLP:journals/corr/abs-2202-01875}.
% Moreover,
%We provide a reasoned survey and a formalization of the mechanisms of the machine learning models underpinning this collaboration, which mostly focuses on classification tasks with only a limited attention to other machine learning tasks, such as regression or clustering.
%We will prevalen\tly consider classification tasks, as they represent well the problem of high stake decisions.
The design of interactive user interfaces based on conversational agents is out of the scope of this survey, since the interactive step in such cases only concerns the stage of communication of ML outcomes and not the underlying learning architecture. We will further discuss the connection between our framework and Large Language Models (LLMs) in Section \ref{subsec:commandart}.
% Given their strong focus on interaction and little to no focus on integration of the heterogeneous agents into a cohesive system, we do not focus on conversational agents and autonomous large language models.
We refer to works in the literature on the topic~\cite{DBLP:journals/csur/YuZLHWJJ22,10.5555/3312046,llms_agents,DBLP:conf/ijcai/NobaniMM21}.}
    % In this paper, we focus on hybrid decision making systems with some degree of interaction between human and machines, and as such we do not specifically address systems where humans are not active agents nor they are pure teaching agents within an environment,
    % % \foscanote{FOSCA: siamo sicuri che non li tocchiamo, forse è poco chiaro cosa intendiamo qua}
    % such as knowledge injection systems, and reinforcement learning systems.
    % Given their strong focus on interaction and little to no focus on integration of the heterogeneous agents into a cohesive system, we do not focus on conversational agents, and interactive XAI algorithms.
    % % For the curious reader,
    % We refer to works in the literature on each of the aforementioned topics~\cite{DBLP:journals/csur/YuZLHWJJ22,10.5555/3312046,DBLP:conf/ijcai/NobaniMM21}.
    %
    \begin{figure}[t!]
        \centering
        \begin{subfigure}[b]{0.20\textwidth}
            \begin{tikzpicture}[node distance=0.5cm, auto,->,shorten >=1pt,font=\small]
                % nodes
                \node[draw=none,fill=none] (data) at (0,0) {Data};
                \node[shape=rectangle,draw=black] (machine) at (0,-1) {Machine};
                \node[shape=circle,draw=black] (human) at (0,-3) {Human};
                \node[draw=none,fill=none] (prediction) at (0,-4.5) {Prediction};
                
                % edges
                \path [->] (machine) edge node {Prediction} (human);
                \path [->] (data) edge node {} (machine);
                \path [->] (human) edge node {} (prediction);
            \end{tikzpicture}
            \caption{Human overseers.}
            \label{fig:paradigms:overseers}
        \end{subfigure}\hspace{1.5cm}
        \begin{subfigure}[b]{0.20\textwidth}
            \begin{tikzpicture}[node distance=0.5cm, auto,->,shorten >=1pt,font=\small]
                % nodes
                \node[draw=none,fill=none] (data) at (0,0) {Data};
                \node[shape=rectangle,draw=black] (policy) at (0,-1) {Orchestrator};
                \node[shape=circle,draw=black] (human) at (1,-3) {Human};
                \node[shape=rectangle,draw=black] (machine) at (-1,-3) {Machine};
                \node[draw=none,fill=none] (prediction_human) at (1,-4.5) {Prediction};
                \node[draw=none,fill=none] (prediction_machine) at (-1,-4.5) {Prediction};
                
                % edges
                \path [->] (data) edge node {} (policy);
                \path [->] (policy) edge node {} (human);
                \path [->] (policy) edge node {Data} (machine);
                \path [->] (human) edge node {} (prediction_human);
                \path [->] (machine) edge node {} (prediction_machine);
            \end{tikzpicture}
            \caption{Learn to Abstain.}
            \label{fig:paradigms:learn_to_abstain}
        \end{subfigure}\hspace{1.5cm}
        \begin{subfigure}[b]{0.20\textwidth}
        \centering
        \begin{tikzpicture}[node distance=0.5cm, auto,->,shorten >=1pt,font=\small]
            % nodes
            \node[draw=none,fill=none] (data) at (0,0) {Data};
            \node[shape=rectangle,draw=black] (machine) at (0,-1) {Machine};
            \node[shape=circle,draw=black] (human) at (0,-3) {Human};
            \node[draw=none,fill=none] (prediction) at (0,-4.5) {Prediction};

            % edges
            \path [->] (data) edge node {} (machine);
            \path [->] (human) edge[bend left] node {Correction} (machine);
            \path [->] (machine) edge[bend left] node {Prediction} (human);
            \path [->] (human) edge node {} (prediction);
        \end{tikzpicture}
        \caption{Learn together.}
        \label{fig:paradigms:learn_together}
        \end{subfigure}    
    \caption{Paradigms of hybrid systems, where human (circle) and machine (rectangle) steps alternate to form a cohesive system. In human overseers (\ref{fig:paradigms:overseers}), the machine performs a prediction, and the human accepts it or rejects it in favor of their own. In Learn to Abstain (\ref{fig:paradigms:learn_to_abstain}), an orchestrator assigns the prediction task to either of the two, which makes the prediction on their own. 
    \textcolor{black}{In Learning Together (\ref{fig:paradigms:learn_together}), the two agents engage in continuous interaction: the machine communicates its reasoning to the human, enabling the latter to understand the machine's internal mechanisms along with possibly rectifying any errors. The feedback loop continues until a human chooses to cease inspecting or interacting with the machine.}}
    \label{fig:paradigms}
    \end{figure}

        Since hybrid systems strive to synergistically combine humans and AI systems, we propose a taxonomy based on the level of integration, which distinguishes three learning paradigm families:
        %This allows us to identify three distinct families \textcolor{black}{of learning pardigms underlying} hybrid systems: 
        \emph{Human overseeing}, \emph{Learning to Abstain}, and \emph{Learning Together} (Figure~\ref{fig:paradigms}).
        Each of these systems furthers the control that the human can exert on the AI system, and as a consequence also the level of integration.
        % These paradigms evolve along the integration axis, each improving the human-machine integration of the previous.
        They follow in chronological order, with systems becoming increasingly integrated as research progresses.
        \emph{Human overseeing}, the simplest of the three, implements a pipeline with no integration between human and machine, the former overseeing the predictions of the latter.
        %Here, the two steps are one after the other.Since the human is aware of the results of the machine step, and is also able to overwrite it, compliance is maximal.
        Instead, \emph{Learning to Abstain} lets only one of the two types of agent perform the task independently, with an orchestrator determining which agent makes each prediction on a case base. Finally,
        \emph{Learning Together} systems establish a communication loop whereby human and machine agents engage across multiple interactions to share and expand their knowledge, aiming for a synergistic resolution of the task at hand.
        %more sophisticated alternation of the two steps, which create a true loop in which human and machine steps are repeated in sequence an arbitrary number of times.
        % \foscanote{ la caratteristica è il loop, evidenziamolo anche dopo}
        %Unlike the previous paradigms, the Learning Together paradigm features a communication loop between human and machine agents.
        
    The survey is organized as follows: in Section~\ref{sec:overview} we provide a general formulation of hybrid systems , and we identify the three learning paradigms: Human overseeing, Learning to Abstain, and Learning Together. In Sections~\ref{sec:hdms1},\ref{sec:hdms2}, and~\ref{sec:hdms3}, we survey and discuss the proposals in the literature of the three paradigms. 
    We  provide an overview of the open problems and a final discussion in Section~\ref{sec:gaps}.
    %and finally conclude in Section~\ref{sec:conclusion}.

% \vspace{-0.4cm}

    %
    \paragraph{\textbf{Paper selection criteria}}
    %We have used different selection criteria for papers based on the specific paradigm being analysed:
    \textcolor{black}{We have developed a taxonomy of three families of learning paradigms for hybrid systems by examining the scientific literature and adjusting the paper selection criteria based on the relevance of each paradigm to the objectives of this study.}
    %To precisely outline three paradigms, we have explored the scientific literature adapting the paper selection criteria to 
    %using different criteria for paper selection. The extent and depth of the research query align with the significance of the specific paradigm in relation to the scope of this study. 
    As for the \textbf{Paradigm 1}, we have we examined papers that have investigated the impact of human oversight on ML models through qualitative or quantitative analysis, but without completing a formal survey.
    %, only focusing on the computational, rather than interactive, aspect of oversight.
    On the contrary, in \textbf{Paradigm 2}
    %\begin{description}
    %\item[Paradigm 1.] We surveyed papers where the effect of human oversight over machine learning models has been studied qualitatively or quantitatively. As such, keywords such as \textit{"monitoring"}, \textit{"decision making"}, \textit{"user experience"} and \textit{"model surveillance"} where used.
    %\item[Paradigm 2.] 
     we conducted a comprehensive review based on the keyword \textit{``learning to reject''}, \textit{``learning to defer''}, \textit{``learning with a reject option''}, \textit{``selective classification''}, \textit{``deferral policy''}, \textit{``deferral function''} and \textit{``defer to expert''}.
     After a preliminary screening of top conference/journal papers and highly cited papers, where we gathered 150 papers focusing mainly on endowing algorithms with abstention mechanisms, we selected 37 papers representing the wide spectrum of solutions of Learning to Reject, and 63 papers encompassing the most significant findings in Learning to Defer.
   % \item[Paradigm 3.] %We surveyed papers that would integrate human contribution directly in the learning process of a ML model. 
    \textcolor{black}{Finally, in \textbf{Paradigm 3}, due to the novelty of the topic, and the heterogeneous use of the term ``hybrid'', which also finds wide application in the human-machine interfaces literature, we have gathered journals and papers matching keywords ``active learning'', ``human feedback'', ``learning with feedback'', ``interactive learning'', ``privileged learning'', ``human AI team'', ``human in the loop'', ``training * feedback'', ``human * advice'', filtering down to a subset of relevant manuscripts, then exploring the cited papers, repeating the process until no relevant manuscripts were found.
    % As such, terms like \textit{"interactive learning"}, \textit{"machine-human learning"}, \textit{"ai-human communication"}, \textit{"human in the loop"} were searched. 
    Our final selection highlighted 15 papers on this particular paradigm. }   
    The chosen papers represent a promising future path for research on Hybrid Decision-Making Systems.
%\end{description}
%For the first paradigm, we surveyed papers where the effect of human oversight over machine learning models has been studied qualitatively or quantitatively. As such, keywords such as \textit{"monitoring"}, \textit{"decision making"}, \textit{"user experience"} and \textit{"model surveillance"} where used; 
%ii) For the second paradigm, we surveyed papers in the "Learning to Reject" and "Learning to Defer" field. Here, we performed a comprehensive selection of papers, searching for \textit{"learning to reject"}, \textit{"learning to defer"}, \textit{"learning with a reject option"}, \textit{"selective classification"}, \textit{"deferral policy"}, \textit{"deferral function"} and \textit{"defer to expert"}. A first screening of top conference/journal papers, and most cited papers, we surveyed 150 papers, focusing mostly on the topic of deferral mechanism for their human component. We ended up selecting 37 papers for the Learning to Reject part, i.e. the most representative of the very wide spectrum of solutions in this field, and 53 papers for the Learning to Defer part, comprising of all the most relevant results in this particular field;
%iii) For the third paradigm, we surveyed papers that would integrate human contribution directly in the learning process of a ML model. As such, terms like \textit{"interactive learning"}, \textit{"machine-human learning"}, \textit{"ai-human communication"}, \textit{"human in the loop"} were searched. The papers presented constitute a prospective future direction of Hybrid Decision Making Systems research.
All papers where searched using Google Scholars and DBLP computer science bibliography, selecting papers with high citation counts and/or published in top journals or conferences, such as AAAI, IEEE, ACM, NeurIPS, etc.

\section{General formulation of Hybrid Systems}
% REPLACE \section{General formulation of Hybrid Decision-Making Systems}
    \label{sec:overview}
    \begin{table}[t!]
    \footnotesize
        %\label{tbl:symbols}
        \centering
        \begin{tabular}{@{} c l l p{9cm} @{}}
            \toprule
            %& \textbf{Symbol} & \textbf{Domain} & \textbf{Codomain} & \textbf{Description} \\ 
            & \textbf{Symbol} & \textbf{Definition} & \textbf{Description} \\ 
            \midrule
            \multirow{2}{*}{\begin{sideways} \footnotesize Agents\end{sideways}} 
               & $H$ & $\mathcal{H}$ & Human agent.\\ 
               & $M$ & $\mathcal{M}$ & Machine agent.\\ 
            \midrule
            \multirow{3}{*}{\begin{sideways}R. Variables\end{sideways}} 
            & $X$ & $\Omega \rightarrow \mathcal{X}$ & Feature matrix. $\mathcal{X}$ includes the empty element $\emptyset$.\\ 
            & $Y_{(\cdot)}$ & $\Omega \rightarrow \mathcal{Y}_{(\cdot)}$ & Label vector: $\mathcal{Y}_{(\cdot)}$ $\subset$ $\mathbb{R}$ (regression), $\mathcal{Y}_{(\cdot)}\subset$ $\mathbb{N}$ (classification) provided by a given agent $(\cdot)\in\{M,H\}$.\\ 
            %& $Y_{(\cdot)}$ & $\mathcal{Y}$ & Label vector provided by a given agent $(\cdot)$.\\ 
            & $Y^*$ & $\Omega \rightarrow \mathcal{Y}$ & Ground truth label vector. \\ 
            \midrule
            \multirow{6}{*}{\begin{sideways}Machines\end{sideways}}  % & $\theta$ & $\Theta$ & & Parametrization of a machine agent or system. \\
            & $f, f_\theta$ & $\mathcal{X} \rightarrow \mathcal{Y}_{(\cdot)}$ & Predictor function implemented by a machine agent ${(\cdot)}$, belongs to the family of functions $\mathcal{F}_M$. The parameter $\theta\in\Theta$ is omitted if not relevant. \\
            & $f^*$ & $\mathcal{X} \rightarrow \mathcal{Y}$ & Ground truth function. Belongs to the family of functions $\mathcal{F}_M$\\
            & $\mathscr{L}_{(\cdot)}$ & $\mathcal{Y} \times \mathcal{Y}_{(\cdot)} \rightarrow \mathbb{R}_{>0}$ & Real-valued loss function of an agent $(\cdot)\in\{M,H\}$.\\
            & $\widetilde{\mathscr{L}}_M$ & $\mathcal{Y} \times \mathcal{Y}_M  \rightarrow \mathbb{R}_{>0}$ & Real-valued surrogate loss function of a machine agent $M$. \\
            %It approximates a non-tractable loss $\mathcal{L}$. \\
%            & $\mathscr{R}$ & $\mathcal{F} \rightarrow \mathbb{R}$ & Risk of a given machine. \\
            & $\rho_{(\cdot)}$ & $\Rho_{(\cdot)} \rightarrow \{0,1\}$ & Deferral or rejection policy of agent $(\cdot)\in\{M,H\}$. \\
            %& $\pi$ & $\Pi$ & & Query policy of a machine agent. \\
            \midrule
            \multirow{4}{*}{\begin{sideways}Humans\end{sideways}}   & $Z_{(\cdot)}$ & $\mathcal{Z}$ & Set of artifact(s) only available to the human agent $(\cdot)$.\\ 
            & $h_{(\cdot)}$ & $\mathcal{X}\times\mathcal{Z} \rightarrow \mathcal{Y}_H$ & Predictor function implemented by the human agent $(\cdot)$. \\
            & $A$ & $\mathcal{A}$ & Hybrid artifact, enables communication among agents. \\
            %between human and machine agents. \\
            & $B$ & $\mathcal{P}(\mathcal{A})$ & Artifact bank, memory storing artifacts. \\
            \bottomrule
        \end{tabular}
        \caption{
        \textcolor{black}{Table of symbols. We will use lowercase letters for elements of a space, e.g., elements $x$ of $\mathcal{X}$. Moreover, $\mathbb{R}$ represents the set of real numbers, $\mathbb{Z}$ represents the set of integer numbers, $\Omega$ is the sample space (i.e., an arbitrary non-empty set), and $\mathcal{P}(A)$ is the power set of the artifacts $\mathcal{A}$.}
        %Elements of a space are indicated with lowercase letters, ordered samples of the space as uppercase letters, e.g., an element of the feature space $\mathcal{X}$ is $x$, and a $n$-large sample of $\mathcal{X}$ is indicated with $X = \{x_i\}_{i = 1}^n$.
        }
        \label{tbl:symbols}
    \end{table}
    \label{sec:generalModels}
        A Hybrid
        % Decision-Making
        System is composed of two types of agents: a machine agent $M$ and a human agent $H$.
        % Consider a probability space $(\Omega,\mathscr{F},\mathbb{P})$ and let $\mathcal{X}$, $\mathcal{Z}$, $\mathcal{Y}, \mathcal{Y}_M$, and $\mathcal{Y}_H$ be measurable spaces
        \textcolor{black}{Consider a probability space $(\Omega,\mathscr{F},\mathbb{P})$ such that $\Omega$ is a sample space (i.e., an arbitrary non-empty set), $\mathscr{F}$ is an event space (i.e., a $\sigma$-algebra over $\Omega$), and $\mathbb{P}$ is a probability measure $\mathbb{F}\rightarrow [0,1]$. Moreover, let  $\mathcal{X}$, $\mathcal{Z}$, $\mathcal{Y}, \mathcal{Y}_M$, and $\mathcal{Y}_H$ be measurable spaces
        such that $\mathcal{X}$ represents the machine feature space, $\mathcal{Z}$ the human expertise, that can be modeled as features, decision rules, etc.,
        %the \textcolor{black}{set of artificts only available to the human agent (e.g., features, decision rules, or domain knowledge)}, 
        and $\mathcal{Y}, \mathcal{Y}_M$, and $\mathcal{Y}_H$ the ground truth, machine and human label spaces for a certain task $T$, respectively. Full list of symbols can be found in Table \ref{tbl:symbols}.}

        \textit{\textbf{Machine model.}}
        Let $X:\Omega\rightarrow\mathcal{X}$ and $Y_M:\Omega\rightarrow\mathcal{Y}_M$ be random variables representing the input and output of a machine $M$, and let $Y^*:\Omega\rightarrow\mathcal{Y}$ be the random variable representing the true labels.
        In the classical setting of supervised learning, given independent and identically distributed pairs $\{(X_i,Y_i^*)\}_{i=1}^n\overset{\text{iid}}{\sim} (X,Y^*)$ drawn from the same unknown joint distribution over $X\times Y^*$, we aim to learn a predictor $f:\mathcal{X}\rightarrow \mathcal{Y}_M$ that approximates, as accurately as possible, the unknown function $f^*:\mathcal{X}\rightarrow\mathcal{Y}$ representing the true relationship between the features and the target. 
        Note that $f$ is chosen from a space of hypothesis $\mathcal{F}_M$ parameterized by $\theta\in\Theta$, however, in our work we often omit the parameter $\theta$ to simplify our notation.
        %Depending on the task $T$ being solved, $\mathcal{Y}_M$ can be, for instance: $\{0,1\}$ (binary classification), $\{1,\ldots,K\}$ (multiclass classification), or $\mathbb{R}$ (regression); while $f^*$ can either be such that $f^*(X)=Y$ or $f^*(X)=P(Y\mid X)$. 
        %The quality of the estimator $f$ is quantified by the machine loss $\mathscr{L}_M:\mathcal{Y}\times\mathcal{Y}_M\rightarrow\mathbb{R}_{>0}$. 
        The ultimate goal of the learning algorithm is to find an hypothesis $f\in\mathcal{F_M}$ that
        % More precisely, $f$ is chosen from a space of hypothesis $\mathcal{F}_M$ in order to minimize the \textit{expected risk}: 
        % \begin{equation}\label{eq:riskM}
        %     \mathscr{R}[f] = \mathbb{E}_{X,Y}[\mathscr{L}(Y,f(X))].
        % \end{equation}
        % \begin{equation}\label{eq:riskM}
        %     f^*= \arg \min_{f\in F_M}\mathscr{R}[f] = \arg \min_{f\in F_M}\mathbb{E}_{X,Y}[\mathscr{L}(Y,f(X))].
        % \end{equation}
        %Note that $\mathcal{F}_M$ is a family of functions parameterized by $\theta\in\Theta$, however, in our work we often omit the parameter $\theta$ to simplify our notation. 
        %In practical settings, the learning algorithm has only access to a limited number of samples. Hence, if the decision problem is not separable, $f$ is chosen to 
        minimizes the \textit{empirical expected risk}, which is defined on the training set as follows:
        \begin{equation}\label{eq:empRiskM}
            \widehat{\mathscr{R}}_n[f] = \frac{1}{n}\sum_{i=1}^n \mathscr{L}_M(Y_{M,i},f(X_i))
        \end{equation}
        The function $\mathscr{L}_M:\mathcal{Y}\times\mathcal{Y}_M\rightarrow\mathbb{R}_{>0}$ represents the machine loss, which quantifies how far the predictions of a hypothesis $f$ are from the true outcome.
        %Let $f^*$ and $\hat{f}^*$ be the minimizers of the expected risk and empirical risk, respectively (i.e., $f^*= \arg \min_{f\in \mathcal{F}_M}\mathscr{R}[f]$, $\hat{f}^*= \arg \min_{f\in \mathcal{F}_M}\widehat{\mathscr{R}}[f]$), and let $\mathscr{R}^*=\inf_{f\in \mathcal{F}_M} \mathscr{R}(f)$ be the Bayes (i.e., minimal) risk. Then, the aim of a learning algorithm is to find $\hat{f^*}$ that is close to $f^*$ with high probability. Whenever such an hypothesis $f$ exists, then it is said to be \textit{consistent}. We remind that a necessary and sufficient condition for nontrivial consistency of empirical risk minimization is given by uniform convergence in probability \cite{Vapnik91}, that is:
        % \begin{equation}
        %     \forall \epsilon>0 \lim_{n\rightarrow+\infty}P(sup_{f\in \mathcal{F}_M}(\mathscr{R}[f]-\widehat{\mathscr{R}}_n[f]>\epsilon)=0.
        % \end{equation}
        %It follows that a crucial choice for learning will be that of defining an appropriate set of functions $\mathcal{F}_M$ that ensure uniform convergence of risks.
        \textcolor{black}{In many cases, the formulation of the loss in equation \eqref{eq:empRiskM} may have a level of complexity that renders direct computation intractable or computationally undesirable.
        For instance, in the typical scenario of multivariate classification, the primary objective is to minimise the 0/1 loss, which aligns with minimising the misclassification rate. Nonetheless, even with this simple choice of loss function, optimisation remains complex in both scenarios of approximating it via a Bayes classifier, due to the curse of dimensionality or possible model mis-specification, and direct computation, owing to the discontinuity and discrete nature of the 0/1 loss, which renders it neither continuous nor differentiable, thus posing significant challenges for optimisation and becoming computationally intractable for numerous nontrivial function classes \cite{Goodfellow2016,Neykov2016}.}
        %In many cases, the formulation of the loss may have a level of complexity that renders direct computation infeasible due to the
        %However, when the underlying risk function is too complex, \eqref{eq:empRiskM} might not be directly computed because of the
        %curse of dimensionality or potential model mis-specification \cite{Neykov2016}; 
        %on the other hand, even when it has a simple representation, such as the zero-one loss in multivariate classification,
        %equation \eqref{eq:empRiskM} is often not continuous nor differentiable, hence extremely hard to optimize and computationally intractable for many nontrivial classes of functions \cite{Goodfellow2016}. 
        The approach typically employed to overcome this issue is to define and optimize a \textit{surrogate loss function} $\widetilde{\mathscr{L}}_M:\mathcal{Y}\times\mathcal{Y}_M\rightarrow\mathbb{R}_{>0}$, that is, a function with good computational guarantees (e.g., differentiability and convexity) that can be easily optimized and whose optimal values approximate well the minimizer of the original computationally hard loss function \cite{Bartlett2006,Lin2004}. 
        The exact formulation of a surrogate loss function is not straightforward, as it depends on the particular task at hand and the desired properties one seeks to guarantee. Refer to the Appendix for an extensive discussion about the fundamental mathematical properties that characterize surrogate losses.       
        % Note that, whenever the surrogate prediction space $\widetilde{\mathcal{Y}}_M$ is different from $\mathcal{Y}_M$, we also assume the existence of a decoding function $g:\widetilde{\mathcal{Y}}_M\rightarrow\mathcal{Y}_M$. 
        % The associated $\widetilde{\mathscr{L}}$-\textit{risk} $\mathscr{R}_{\widetilde{\mathscr{L}}}$ and \textit{empirical $\widetilde{\mathscr{L}}$-risk} $\widehat{\mathscr{R}_{\widetilde{\mathscr{L}}}}$ are defined as in \eqref{eq:riskM} and \eqref{eq:empRiskM} by replacing the loss $\mathscr{L}$ with its surrogate $\widetilde{\mathscr{L}}$. Analogously, the \textit{optimal $\widetilde{\mathscr{L}}$-risk} is defined as $\mathscr{R}_{\widetilde{\mathscr{L}}}^*=\inf_{f\in \mathcal{F}_M}\mathscr{R}_{\widetilde{\mathscr{L}}}(f)$
        % In this setting, the classifier is inferred from the decision function $\hat{f}_n$ that minimizes the empirical $\widetilde{\mathscr{L}}$-risk: 
        % \begin{equation}
        %     \hat{f}_n = \arg \min_{f\in \mathcal{F}_M}\widehat{\mathscr{R}}_{\widetilde{\mathscr{L}}}(f) = \arg\min_{f\in \mathcal{F}_M} \frac{1}{n}\sum_{i=1}^n\mathit{\widetilde{\mathscr{L}}(Y_i,f(X_i))}
        % \end{equation}

        \textit{\textbf{Human model.}}
        Let $Z:\Omega\rightarrow\mathcal{Z}$ and $Y_H:\Omega\rightarrow\mathcal{Y}_H$ be random variables representing, 
        respectively, the human expertise and the predictions of
        %the input and output of 
        a human agent $H$ for a certain task $T$, where $H$ is modeled as a predictor $h:\mathcal{X}\times\mathcal{Z} \rightarrow \mathcal{Y}_H$. Generally, we consider $\lvert \mathcal{X} \bigcap \mathcal{Z} \rvert \geq 0$, implying that there may exist shared information between the machine and the human, or there may not. 
        
        It is worth noting that there may be a divergence between the predictions made by human agents and the ground truth labels.  
        %Due to their specific skills, it is expected that the predictions made by human experts may diverge from the ground truth labels. 
        Indeed, different levels of background knowledge, experience, or personal biases can lead to distinct decision-making outcomes, resulting in both correct and incorrect predictions across various domains within the input space \cite{Lampert2016}. 
        Therefore, we additionally take into account a  loss function $\mathscr{L}_H:\mathcal{Y}\times\mathcal{Y}_H\rightarrow\mathbb{R}_{>0}$ as an indicator of the quality of human predictions.
        The observation that various types of errors are not only made by distinct human agents but also occur between humans and AI models provides support for the transition towards paradigms that incorporate hybrid combinations of human and machine predictions \cite{Kerrigan2021}.
    
    % \paragraph{Hybrid artifacts.}
    %     The core of a hybrid system is the \emph{hybrid artifact}, an artifact that enables human and machine to exchange information in a shared language.
    %     Different paradigms call for different artifacts: machine oversight requires \emph{oversight artifacts}, \textcolor{black}{learning to abstain models} require \emph{\textcolor{black}{abstaining} artifacts}, and learning together\mattialightnote{cambiare} systems require \emph{interaction artifacts}:
    %     %
    %     \begin{description}
    %         \item[Oversight artifacts.] Enable the human agent to oversee the machine by providing a trace of its behavior.
    %         \item[Socratic artifacts.] Enable the human agent to perform the task in place of a deferring machine.
    %         \item[Interaction artifacts.] Enable the human and machine agents to exchange information.
    %     \end{description}
    %     %

    %     We denote artifacts with $a$, sets of artifacts with $A$, and the space of artifacts with $\mathcal{A}$.
    %     Artifacts can be as simple as the prediction $y$ given by a machine~\cite{}, and as complex as logic theories~\cite{} and small knowledge bases~\cite{}.
    %     As the degree of integration of human and machine grows, so grows the complexity of the exchanges, and so tends to grow the sophistication of the artifacts.

\section{Peering into the Machine: Human Overseers}
\label{sec:hdms1}
        Human oversight~\cite{koulu2020proceduralizing} is probably the simplest and most straightforward form of hybrid system.
        In this first paradigm, machine and human agents are independent of each other, the former performing a task, and the latter \emph{verifying} its predictions.
        Informally, the human agents perform a straightforward task: given the machine computation and/or the input data, either accept or reject the computation.
        More formally, a \emph{human oversight policy} $\rho_H$ is a binary function that, given a machine $M$ implementing a function $f \in \mathcal{F}_M$, an overseeing human $H$ with additional expertise $Z \in \mathcal{Z}$, and some input data $x \in \mathcal{X}$, either \textit{accepts} or \textit{rejects} the prediction given by $M$.
        In its most general formulation, $\rho_H$ is formulated as follows:
        \begin{equation*}
            \rho_H: \mathcal{X} \times \mathcal{Y}_M \times \mathcal{Z} \times \mathcal{F}_M \rightarrow \{\text{accept}, \text{reject}\}.
            \label{eq:prdg1:def:gen}
        \end{equation*}
        In many cases, the human agent either does not have access or does not take into consideration the machine itself, thus the above formulation is often reduced to:
        \begin{equation*}
          \rho_H: \mathcal{X} \times \mathcal{Y}_M \times \mathcal{Z} \rightarrow \{\text{accept}, \text{reject}\}.
          \label{eq:prdg1:def}
        \end{equation*}
    \begin{tcolorbox}[colback=blue!5!white,colframe=blue!50!white,title=Example of melanoma detection: Human oversight,fonttitle=\small]
    \small
        In the paradigm of Human Oversight, the dermatologist is presented with an image, the prediction given by the machine, and has to decide whether to agree or not with the prediction of the machine, i.e., whether the given picture is a melanoma or not.
    \end{tcolorbox}
        
        Other than simply leveraging their own expertise $Z$, overseers often leverage external factors in their decision.
        In the simplest of cases, rejected patterns of misbehavior are limited to \textbf{machine-specific} failures in which the underlying context is of little or no impact.
        In these context-independent scenarios, the overseers aim to identify machine failures induced by machine-specific causes by tracking a set of \emph{subjects of monitoring} around which the overseeing policy will be centered.
        Machine-specific failures
        % are inherently machine-specific and
        may be induced by wildly different factors in each machine,
        hence 
        % here
        we abstract over the underlying causes, since the goal of machine oversight is to \textit{identify}, rather than \textit{diagnose}, undesired behavior.
        We tackle behavior correction later in Section~\ref{sec:hdms3}.
        Common subjects of monitoring are:

          \begin{itemize}
              \item \textit{Data shift.} Data shift is generally intended as a change in the data distribution~\cite{quinonero2008dataset}, and may be due to a change in feature distribution, i.e., \textit{covariate shift} or knowledge drift, or label distribution, i.e., \textit{prior shift};
              these shifts may occur frequently in real-world scenarios, namely due to data seasonality, sampling bias, or naturally-occurring distribution changes.
              Unlike
              % naturally-occurring
              spontaneous
              \textit{outlier} instances, dataset shifts are responsible for consistent and predictable failures in the model, thus they have to be accounted for.
    
              \item \textit{(Partial) Model performance.} Partially stemming from data shift, model performance is another primary subject of interest.
              Here, we distinguish between two classes of performance metrics: ``global metrics'', where the model is evaluated \textit{wholly}, and ``partial metrics'', where the model is evaluated \textit{partially} on a suitable subset of data.
              Data shift may indeed only affect a subset of data, hence global metrics may easily deceive the overseers.
    
              \item{\textit{Model uncertainty.}} Classical learning algorithms are trained to predict, setting \emph{confidence calibration} aside.
              Simply put,
              % algorithms
              they
              operate under a \emph{closed world assumption} where the only available option is to give a prediction, that is, the model has no notion of uncertainty nor of the unknown.
              This projects a false high confidence, seldom tricking the algorithm user into overestimating its competency~\cite{DBLP:journals/pacmhci/BucincaMG21}.
    
              % [CIT]\item \textit{Decision complexity.} Even with highly sophisticated and precise models, some decisions are inherently suitable for human rather than algorithmic reasoning~\cite{campbell20years,sadler2004intuitive,parikh1994intuition}.
              \item \textit{Decision complexity.} Even with highly sophisticated and precise models, some decisions are inherently suitable for human rather than algorithmic reasoning~\cite{parikh1994intuition}.
              In this context, it is crucial to anticipate which decision is which.
              % This trend is bound to increase with both time and performance of models as more and more tasks become easier to solve.
              
          \end{itemize}

        Operating as self-contained agents, machines often lack the decision context wherein their predictions are evaluated and applied.
        Failures in this domain are said to be \textbf{context-dependent}.
        Context can be of primary importance, and humans are far better suited than machines in understanding it and integrating it in their decision-making process.
        For a human, concerns such as fairness, legality, and explainability of the decision are strong contextual motivations that a machine does not necessarily take into proper account.
        Unsurprisingly, most of them are already being encoded in several legislatures, which strongly discourage or punish discriminative or otherwise illegal~\cite{babuta2021machine}, unexplainable~\cite{malgieri2017right} decisions and behaviors.
        Much of this stems from the current use of machine agents in ethically-charged contexts.
        For instance,
        % [CIT] machines are leveraged in monitoring and discouraging~\cite{babuta2021machine,babuta2018machine,calo2020automated} illegal activities, where they often yield unfair or biased predictions~\cite{ferguson2020high,geiger2021discriminatory}; furthermore, they are a critical component of speech regulation, an extremely dynamic use case where human scrutiny and decision autonomy are essential, yet they often regulate marginally- or fully-free~\cite{gillespie2018custodians,bambauer2018platforms} speech;
        machines are leveraged in monitoring and discouraging~\cite{babuta2021machine} illegal activities, where they often yield unfair or biased predictions~\cite{ferguson2020high}; furthermore, they are a critical component of speech regulation, an extremely dynamic use case where human scrutiny and decision autonomy are essential, yet they often regulate marginally- or fully-free~\cite{gillespie2018custodians} speech;
        they aid hiring in public and private companies, yet they are biased~\cite{DBLP:conf/fat/RaghavanBKL20}.

        % [CIT] What's more, context is often dynamic and loosely defined~\cite{nahmias2021oversight,yeung2019ai,busuioc2022ai}, and thus integrating it into the machine is an open challenge in and of itself.
        What's more, context is often dynamic and loosely defined~\cite{nahmias2021oversight}, and thus integrating it into the machine is an open challenge in and of itself.
        Jointly, machine-specific and context-specific failures offer a strong motivation for machine oversight. Yet, even though the \textit{why} is clear, \textit{how} machine oversight is to be implemented is still an open problem.
        Even worse, machine oversight poses a set of inherently human problems to face.

    \subsection{Monitoring pitfalls}
    \label{sec:hdms1:pitfalls}
      While technical solutions to machine-specific failures have already been developed, context-dependent failures cause a plethora of additional and more complex problems. Given that humans are usually the time bottleneck when it comes to decision-making, one cannot let the overseers monitor every prediction, thus one needs to understand \textit{when} to let the overseer monitor the machine.
      A conventional
      solution, which we explore in Section~\ref{sec:hdms2}, is to let the machine itself call the human into action.
      % \\
      Here, we focus instead on the overseers and their inherently human fallacies, which lead to some natural pitfalls of the whole monitoring process.
      
 \textbf{\textit{Human Executors and Skeptics.}}
      Overseers need to be aware of two possible cognitive biases: algorithmic \emph{aversion}~\cite{dietvorst2015algorithm} and \emph{overreliance}~\cite{logg2019algorithm}.
      Algorithmic aversion pushes the overseers towards excessively doubting the machine, thus introducing unnecessary monitoring in the decision-making process.
      Algorithmic aversion manifest itself independently of the performance of the machine~\cite{dietvorst2015algorithm,main2016cook}, and more strongly when the machine fails.
      In other words, every single perceived mistake of the machine compounds in increasing the rejection rate of the overseer.
      Unlike algorithm aversion, algorithm overreliance
      occurs when human agents under-monitor a machine system, and thus act as mere executors.
      The two biases are well-documented in the literature, and regardless of the machine system they human agents interface with, they are to be accounted for.
      % Algorithmic aversion\mattialightnote{here add after reading andrea's report}

    \textbf{\textit{Biased monitoring.}}
      Automation bias is particularly strong when human agents oversee fairness- related tasks where the task directly involves other humans.
      When monitoring decision on pre-held stereotypes, say on
      % minority
      vulnerable
      groups, overseers either avoid monitoring in the first place~\cite{alon2023human} or further confirm the stereotypes~\cite{DBLP:journals/automatica/Bainbridge83,ferguson2020high}.
      On similar reasoning, particularly in cases of fairness evaluation, human agents tend to under-monitor when they perceive affinity towards the monitor case at hand~\cite{DBLP:conf/eaamo/Grgic-HlacaLWR22}, or simply when they deem their reasoning more ``human'' than the machine's~\cite{lee2018understanding}.
      On an even more biological level, intrinsic demographic traits are also likely to play a role in the decision-making of the human agents~\cite{DBLP:conf/chi/0002HZ20,DBLP:journals/corr/abs-1712-09124}.
      % [CIT] To further increase the complexity of setting up a set of overseers, it is often the case that human agents, in part for the aforementioned reasons, have a low level of agreement on the correct task solution~\cite{DBLP:conf/sigecom/AlbachW21,DBLP:conf/www/Grgic-HlacaRGW18,DBLP:journals/corr/abs-2103-12016}.
      To further increase the complexity of setting up a set of overseers, it is often the case that human agents, in part for the aforementioned reasons, have a low level of agreement on the correct task solution~\cite{DBLP:conf/www/Grgic-HlacaRGW18}.
    
    \textbf{\textit{Failure to oversee and trust calibration.}}
      Even more worrying than biased monitoring is the failure to reject obvious machine failures.
      In a pilot study with legal experts,~\cite{englich2006playing} showed that, when assisted by a faulty machine agent, domain experts incorporate into their decision-making machine recommendations based on irrelevant or random factors, with extreme cases in which the domain experts were knowingly introducing random factors themselves -- a clear case of placebo effect where the mere presence of a machine prediction, regardless of its correctness, induces an almost blind trust in the human agent.
      % [CIT] Unsurprisingly, overseers have repeatedly shown to be unable to properly assess their ability to assess the performance of a given machine~\cite{DBLP:journals/corr/abs-1812-03219,green2019disparate,albright2019if}.
      Unsurprisingly, overseers have repeatedly shown to be unable to properly assess their ability to assess the performance of a given machine~\cite{DBLP:journals/corr/abs-1812-03219,albright2019if}.

    \subsection{Enhanced monitoring: the case for Explainable AI}
    \label{sec:hdms1:explanations}
        Overseeing a machine simply through its predictions and uncertainty provides minimal tools to a human agent, which can easily fall into one or more of the aforementioned pitfalls.
        To enhance their overseeing power, humans are often accompanied by \textit{explanation algorithms}, that is, algorithms able to further explain the predictions given by a machine.
        Explainable AI (XAI)~\cite{DBLP:journals/csur/GuidottiMRTGP19} is a recent field of research aiming to shed light on the prediction process of AI models by extracting human-understandable explanations.
        Explanations allow an overseer to peer into the machine and get a grasp of what features a machine is relying upon~\cite{DBLP:conf/nips/LundbergL17} and what rule-like logic it is following~\cite{DBLP:journals/expert/GuidottiMGPRT19} to make its predictions, what training instances have had a particular influence on the learning process~\cite{DBLP:conf/icml/KohL17}, and how one could change the input instance to achieve a different prediction~\cite{DBLP:journals/corr/abs-2303-09297}.
        Explanations have shown to empower the human agent into better understanding the machine, thus improving their ability to monitor it.
    % \paragraph{From overseeing to predicting}
    %   \foscanote{FOSCA: questa sezione non ha molto enso, potrebbe essere rimossa} Given the faults of machine oversight, and the inherent difficulties to tackle it, it is a sensible choice to minimize it by \textit{i)} reducing the number of predictions the human agents have to oversee, and \textit{ii)} maximize the effectiveness of the oversight on such instances.
    %   Classically, this translates into two hybrid learning paradigms, one an extension of the other:
    %   \emph{Learning to Reject} (L2R) tackles problem \textit{i)} by removing the \textcolor{black}{human} oversight policy $\rho_H$ and replacing it with a learnable \emph{machine rejection policy} $\rho_M$ which, like the human one, accepts or rejects the machine prediction;
    %   and \emph{Learning to Defer} (L2D) extends the Reject paradigm by both rejecting and directly designating the agent most apt to make the prediction.
\section{Learning to abstain}
  \label{sec:hdms2}
  To mitigate human failures in the monitoring process of the aforementioned HS paradigm, a potential approach involves developing an enhanced machine architecture that enables the machine learning model to refrain from predicting on certain instances. In broad terms, whenever the machine prediction can't be relied upon, it may be better to pay an extra costs and defer the prediction to a human agent~\cite{cortes2016l2r}. The ultimate goal of a Learning to Abstain system is thus to maximize the overall performance of the hybrid system by effectively identifying, for each single instance, the agent best suited for a given prediction.
  %The implementation of such paradigm leads to the definition of a hybrid decision-making framework characterized by a task division between the human and the machine. 
  This particular setting does not involve any formally defined interaction between human and machine agents. Instead, it can be better characterized as a policy in which a machine autonomously directs input data towards an agent, whether it be a human or an AI, based on an estimation of superior performance in each individual case. In other words, the algorithms in this paradigm produce models optimized to leverage the option to abstain w.r.t. their primary predictive task. The human interaction in this paradigm may come afterwards, e.g., operating on those instances rejected by the Learning to Abstain system.

  %\textcolor{red}{For instance, let us return to the example presented in Section \ref{sec:introduction} on content moderation in an online forum. This task is characterized by a frequent occurrence of inherent ambiguities, such as the classification of toxic and satirical content. While humans can readily solve these ambiguities by relying on their understanding of common sense and contextual knowledge, machines face significant challenges in accomplishing the same task due to the necessity of figuring out implicit meanings within words and sentences. Therefore, it may be preferable to delegate the moderation of the cases that are particularly difficult and uncertain for the machine to human workers, rather than depending solely on an automated system. In other words, we may think to perform content moderation within a hybrid system capable of performing two distinct tasks: \textit{i}) detecting and refraining from making predictions on "hard" (i.e., uncertain) instances that would (presumably) benefit from the assessment of a human worker, and \textit{ii}) independently handling all "easy" (i.e., predictable with high confidence) instances.}

  At the most general level, the machine learning systems that fall within this framework can be categorized into \textbf{Learning to Reject} (L2R)~\cite{Chow1970,cortes2016l2r} and \textbf{Learning to Defer} (L2D)~\cite{Madras18,Mozannar20} systems, depending on whether the model assumes a pre-specified cost for abstaining or instead is designed to work adaptively with the
  % [REPLACE] decision maker.
  human agent.
  Both classes of algorithms share the same set-up, namely an
  % [REPLACE] HDMS
  HS
  where the machine (or a system of machines) should learn not only a standard classification function but also a \textit{rejection function} (also called \textit{deferral function} in the L2D domain) under the optimization objective of maximizing the performance of the human-AI system as a whole. The deferral action incurs an additional cost which is offset by the payoff of an expected gain in performance resulting from querying a human expert, who can use information that the machine cannot rely on (e.g., professional experience and common sense). Although L2R and L2D have been tackled as separate and independent problems, a seminal paper by \citet{Madras18} demonstrated that L2R can be regarded as the specific case of L2D in which a fixed cost is allocated to each deferred instance. Consequently, they proposed a revised definition of L2D as an \textit{adaptive} L2R approach.
  Therefore, we devote most of our discussion to L2D (Section \ref{sec:L2D}) while giving a more general overview of L2R (Section \ref{sec:l2r}).
  %Moreover, L2R as a framework is older and more established in the literature, and there are already multiple surveys on the subject \cite{HendrickxSurv,ZhangSurv}.

\subsection{Learning to Reject}
\label{sec:l2r}
Learning to Reject (L2R) was first introduced by \citeauthor{Chow1970} \cite{Chow1970}, and its general formulation constitutes a base for more advanced learning to defer algorithms (Section \ref{sec:L2Dsingle}). 
\begin{tcolorbox}[colback=blue!5!white,colframe=blue!50!white,title=Example of melanoma detection: Learning to Reject,fonttitle=\small]
\small
In the L2R paradigm, the algorithm is trained observer the image of a nevus and evaluate weather to make a prediction or abstaining from it. 
For example, edge cases which require further analysis, or out of distribution images with peculiar melanoma shapes, sizes, colors, or even images of patients with peculiar skin complexion, e.g., due to skin disease or poor imaging. In such a scenario the dermatologist would again observe the output of the algorithm, that is a prediction of malign or benign melanoma, or a rejection of the image.

\end{tcolorbox}
In the literature, this area of research is referred to with several names: \textit{learning to reject} \cite{ZhangSurv}, \textit{selective classification} \cite{el_yaniv_foundation}, or \textit{machine learning with a reject option} \cite{HendrickxSurv}. The general idea is to provide machine learning algorithms with the means to learn when to abstain from actually making a certain prediction. Here, the focus is on optimizing machine learning algorithms to include a reject option, thus improving the final performance by refraining from predicting on critical instances.
Therefore, humans are not an active agent in this framework, as the focus is on finding the best combination of machine prediction and machine abstention.
However, in the context of Hybrid Systems, one could envision real life scenarios where L2R models can be used to focus the attention of a human overseer on the subset of instances that are rejected by the model.
Learning to Reject is a widely studied problem with a long history of research, already touched upon in several previous surveys. Therefore, we report a similar general definition to the work of \citeauthor{HendrickxSurv} \cite{HendrickxSurv} and \citeauthor{ZhangSurv} \cite{ZhangSurv}. 
The goal of L2R algorithms is to learn a model $f_{\rho_M}$ composed of two parts: a prediction function $f:\mathcal{X}\rightarrow \mathcal{Y}_M$ and a rejection policy $\rho_M:\mathcal{X} \rightarrow \{0,1\}$
%$\rho:\mathcal{X}  \times \mathcal{Y}_M \times \Theta \rightarrow \{0,1\}$. 
The composed system can be defined as $f_{\rho_M}:\mathcal{X} \rightarrow \mathcal{Y}_M \bigcup \{\emptyset\}$ such that:
\begin{equation}
\label{eq:l2rSystem}
f_{\rho_M}(x) = \begin{cases*} 
    \emptyset   & if $\rho_M(x)=1$\\
    f(x)       & otherwise 
             \end{cases*}
\end{equation}
That is, if the rejection policy $\rho_M$ rejects $x$, then no prediction is made. If instead $\rho_M$ accepts $x$, then the prediction function $f$ is applied to $x$ and the result $f(x)$ is obtained. The function $f$ is generally assumed to be a classifier.
Ideally, $\rho_M$ should be able to prevent misprediction of $f$ while conversely accepting examples for which a good prediction is more likely. The policy $\rho_M$ that actually performs the rejection operation is called \textit{rejection function} or \textit{rejector}. 
%\citeauthor{ZhangSurv} \cite{ZhangSurv} and \citeauthor{HendrickxSurv} \cite{HendrickxSurv} Essentially identify the latter as rejection options that rely on evaluating the confidence of the model.
The focus of a rejector, as mentioned, is to refuse those examples for which the classifier $f$ is expected to output an incorrect prediction. More generally, the focus of research in this field is to find the optimal balance between accuracy on accepted instances and number of rejected instances. We can generally divide the types of rejection into two macro-categories, depending on the type of data that are being rejected:

\begin{itemize}
    \item \textit{Novelties rejection}, i.e., rejections of examples that differ significantly from the data points in the training set $X$ of $f$ and therefore are likely to cause a misprediction.
    \item \textit{Ambiguities rejection}, i.e., rejections of examples in proximity of the decision boundary of $f$.
\end{itemize}

Novelties are examples that are, in general, very different from the ones that $f$ was trained on. Novelties are usually data that may be considered out of distribution (OOD), or simply particularly unusual instances.
Ambiguities are typically examples that can either be classified as any by $f$ as their probability of being a member of one is almost equal to the probability of belonging to the others \cite{GyorfiBayes}. Depending on the kind of mistake that needs to be addressed, different methods for learning the rejector can be used.
The rejector can be based only on feature observation or on model evaluation. Therefore, $\rho_M$ may have access to only the feature space $\mathcal{X}$ or to both $\mathcal{X}$ and the output of the predictor $f$, or even to the architecture and parametrization of $f$. In essence, a rejector can either be:

\begin{itemize}
    \item \textit{Independent}, i.e., the rejection function is learned regardless of the predictor, by observing solely the feature space.
    \item \textit{Dependent}, i.e., the rejection function is built by either querying the predictor or by relying on particular characteristics of the predictor.
\end{itemize}

\subsubsection{Independent rejectors} Independent rejectors are mostly designed to perform \textit{Novelties rejections}. Most of the works in this particular area frame the problem as \textit{Outlier detection} \cite{outlier_survey_2004} or \textit{Anomaly detection} \cite{anomaly_comprehensive_2020}. Another widely-used term is \textit{Open-set recognition} \cite{openset_surv}, where the focus is on optimizing the accuracy of a predictor both on in-distribution and out-of-distribution data. The basic idea is to recognize and handle those examples that present out-of-distribution characteristics. This is done independently from the predictor function $f$, i.e., the rejection mechanism can be applied directly to the data in an agnostic way w.r.t. the trained model $f$. 

Some of the earliest work in this area leverage statistical methods to find anomalies in the data. \citeauthor{Seo_Wallat_Graepel_Obermayer_2000} use the estimated posterior variance of Gaussian processes  as a testing point to understand whether to reject a data point or not \cite{Seo_Wallat_Graepel_Obermayer_2000}. Another similar technique was proposed by \citeauthor{Coles2001} \cite{Coles2001} which relies on statistical models studied in Extreme Value Theory to understand which examples may actually be considered outside expected distributions. This technique has been employed in many different tasks, e.g. facial recognition \cite{Scheirer11} and network analysis \cite{math11092171}.
% Recently  \citeauthor{ruddEVM} proposed a model for kernel-free variable bandwidth incremental learning called Extreme Value Machine \cite{ruddEVM}, based on Extreme Value Theory.
Recently,~\cite{ruddEVM} proposed the Extreme Value Machine, a model for kernel-free variable bandwidth incremental learning based on Extreme Value Theory.
% The Extreme Value Machine is proven to be able to perform Novelty rejection on an image classification task.
One limitation of
% the Extreme Value Machine
this model
is that it relies on the distances between the instances belonging to different classes in the training set, so that classes that are unreasonably distant from the majority classes in the training may be rejected in their entirety.
Some works address this limitation by relying on generalized Pareto distribution approximation to determine whether a new example is a normal or abnormal data point \cite{vignotto_extreme_2020}.

Other approaches rely on models trained to recognize the training distribution and perform anomaly detection as a separate task. One such method was proposed by \citeauthor{Coenen2020ProbabilityOD} and it relies on One-Class Support Vector Machines \cite{Coenen2020ProbabilityOD} specifically fitted to recognize the training distribution. Alternatively, Gaussian Mixture Models can be employed to estimate the training distribution, and thus detect novelties \cite{Landgrebe}. Recently \cite{fewshotanom} proposed a few-shot learning approach where margin-loss w.r.t. the training class is used to train a model to detect anomalies. 

While independent rejectors are mainly designed to detect anomalies, \citeauthor{AsifNets} \cite{AsifNets} propose a framework for generalized rejection based on jointly-trained Neural Networks. These networks are trained with a dual-penalization both on misprediction and incorrect rejection. Since the rejection network is trained independently, it can then be used with any other prediction model, and it can also perform ambiguity rejections.

\subsubsection{Dependent rejectors} Dependent rejectors are much more studied than independent ones. They are tied to the outputs or properties of the predictor and as such can be learned either after the learning phase of $f$ or simultaneously.  Adopting the same naming convention that we will use for Learning to Defer models (Section \ref{sec:L2D}) we call the former \textit{staged rejectors} and the latter \textit{joint rejectors}.

\textbf{\textit{Staged rejectors}} look at the predictor $f$ output to estimate the \textit{confidence} or \textit{uncertainty} of the model. The works that fall in this category are usually referred to as \textit{confidence based} or \textit{uncertainty based}. These approaches estimate uncertainty by \textit{i)} formulating a good confidence metric in and \textit{ii)} selecting an appropriate threshold for determining the actual rejection \cite{Dalitz2009RejectOA,Grandvalet2008,Jnior2017NearestND}.
Formally, the rejection of an instance can be seen as:
$\rho_M(x,f) = \mathbbm{1}_{c(x,f) > \tau}$
where $c(x,f)$ is the confidence metric, which depends both on the instance $x$ and the prediction function $f$, and $\tau$ is the threshold hyperparameter \cite{hansen_error_reject_1997,el_yaniv_foundation}. 
Many different metrics for confidence have been proposed in the literature. Confidence metrics can be
% divided essentially in two categories:
based on:

\begin{itemize}
    \item \textit{Hard predictions} where the exact class-output of predictor $f$,i.e., $y_M=f(x)$, is used to assess the confidence of the prediction. These works usually observe multiple predictions and consider the class-wise variance of these predictions as indication of uncertainty \cite{battisti_democracy,sousa_map}.
    \item \textit{Soft predictions} where some scoring output of $f$ is used as an estimation of $P(y_M|x)$  to determine the confidence. This approximation of confidence is more widely studied, and the array of solutions that leverage soft predictions is much varied \cite{destefano_shakespear,brinkrolf_lvq}.
\end{itemize}

The are also other methods that, while still exploiting some output of the predictor, have slightly different mechanisms to approximate the confidence.
For example, in the work of \citeauthor{tortorella_svm} \cite{tortorella_svm} the authors exploit directly the score of SVMs as a distance from the decision boundary.
In other works, the distance from the k-th nearest prototype is used as a proxy of confidence \cite{Johannes_interpretable_conf}.

The confidence threshold $\tau$ can either be global, or local: a global threshold is usually used for those predictors with an equally well calibrated confidence metric over the entire feature space \cite{capitaine_frelicot, ceccotti_vajda, Fischer_multiple_tau}.
Multiple local thresholds $\tau_1, \ldots, \tau_n$ \cite{Fischer_variance} are instead best suited for predictors with
% an accuracy that varies in certain cases
variable accuracy
\cite{pillai} or for different class-wise variance for the confidence metric. 

\textbf{\textit{Joint rejectors.}} Also sometimes called \textit{integrated rejectors} \cite{HendrickxSurv} are directly part of the prediction model, i.e., the rejection option is treated as an additional class to be learned by the predictor. In this setting, predictor $f$ and rejector $\rho_M$ are simultaneously learned with a single algorithm, specifically designed to learn both functions. This means that it is difficult to technically distinguish and separate the predictor from the rejector \cite{cortes2016l2r}. 
Some joint-learning approached rely on the definition of a specifically formulated objective function able to penalize both incorrect predictions and rejections. Several works resort to surrogate loss functions in place of a more natural discrete loss in order to more easily solve the problem via minimization \cite{Bartlett2008,RamaswamyReject}. Other works aim at specifically leverage some evaluation metric, for example \citeauthor{pugnana2023} \cite{pugnana2023} design a model-agnostic approach for probabilistic binary classifiers with a reject option specifically aimed at optimizing Area Under Curve. Other works directly allocate an extra class for rejection to any given predictor and assign a specific penalization cost for predicting such class \cite{Ziyin2019}.
Finally, several joint-learning algorithms for rejection have been developed specifically for certain kinds of machine learning models, for example for Support Vector Machines \cite{Grandvalet2008,LIN2018} and for Neural Networks \cite{Geifman2019SelectiveNetAD}.

\subsubsection{Cost model for rejection}

Approaches for L2R need to achieve a balance between predictive performance and rejection rate. In fact, classifier with a reject option can entirely trade performance for coverage of vice-versa. This means that, in theory, to achieve maximum performance, a predictor with a reject option can opt to reject all instances and therefore never actually make a misprediction. Otherwise, the predictor can opt to never reject in order to achieve maximum coverage over the data, thus completely forgoing the benefits of implementing a reject option in the first place. To avoid this, an appropriate cost model must be implemented. One foundational cost model can be found in the work of \citeauthor{cortes2016l2r}\cite{cortes2016l2r}:
\begin{equation}
\label{eq:l2rcmod}
cost(x) = \begin{cases*} 
    0,  & if $\rho_M(x)=0 \wedge f(x)=y^*$\\
    1,  & if $\rho_M(x)=0 \wedge f(x)\neq y^*$\\
    \mathcal{R} & if $\rho_M(x)=1.$
    \end{cases*}
\end{equation}
where $y^*$ is the ground truth for example $x$, and $\mathcal{R}$ is a fixed, predefined cost for rejection. 
If we adopt such a cost model, we can express the the learning objective of a L2R algorithm as follows:
\begin{equation}
\label{eq:l2robj}
%\min_{f,\rho_M} \sum_{x \in X}[(1-\rho_M(x))\mathbbm{1}_{f(x)\neq y^*}+\rho_M(x)\mathcal{R}]
\min_{f,\rho_M} \sum_{x \in X}[\mathbbm{1}_{\rho_M(x)=0}\mathbbm{1}_{f(x)\neq y^*}+\mathbbm{1}_{\rho_M(x)=1}\mathcal{R}]
\end{equation}

% In essence, we either pay a cost if we mispredict ($\mathbbm{1}_{f(x)\neq y^*}$) some instance $x$ that we accepted ($\rho_M(x)=0$) or we pay a cost $\mathcal{R}$ if we reject that instance ($\rho_M(x)\mathcal{R}$). 
% % Although this cost model makes perfect sense,
% Although sensible, this cost model
% has the major drawback of having to determine the rejection cost $\mathcal{R}$ beforehand. Rejection cost can be determined depending on the application domain, but this is not always an option \cite{Geifman2019SelectiveNetAD}.
% The work of \citeauthor{GeifmanNIPS17} \cite{GeifmanNIPS17} is among the most important in this field, as it tries to tackle the aforementioned problem. In their proposal, the authors propose a method called Selection with Guaranteed Risk Control that looks at the problem of rejection cost determination from a different angle, that is looking at the coverage of the model. Coverage of a model is defined as the portion of the data that the model accepts for prediction: $\frac{1}{n}\sum_{x \in X}\mathbbm{1}_{\rho_M(x)=0}$.
% Following Eq. \eqref{eq:l2robj} the learning objective can be expressed as a minimization of risk subject to a coverage constraint that forces the model to predict at least a certain portion of instances. Formally:

We incur a cost when mispredicting ($\mathbbm{1}_{f(x)\neq y^*}$) an accepted instance 
($\rho_M(x)=0$) or a cost $\mathcal{R}$ when rejecting it ($\rho_M(x)\mathcal{R}$).
While sensible, this approach requires predefining $\mathcal{R}$ which may not always be feasible \cite{Geifman2019SelectiveNetAD}. \citeauthor{GeifmanNIPS17} addressed this by introducing Selection with Guaranteed Risk Control, focusing on model coverage, defined as the fraction of data accepted for prediction: $\frac{1}{n}\sum_{x \in X}\mathbbm{1}_{\rho_M(x)=0}$. Their objective minimizes risk while ensuring the model predicts a minimum coverage. Formally:

\begin{equation}
\begin{aligned}
%\min_{f,\rho_M} \quad & \frac{\sum_{x \in X}(1-\rho_M(x))\mathbbm{1}_{f(x)\neq y^*}}{\sum_{x \in X}(1-\rho_M(x))} \\ \textrm{s.t.} \quad & \frac{1}{n}\sum_{x \in X}(1-\rho_M(x)) > \mathcal{C}
\min_{f,\rho_M} & \frac{\sum_{x \in X}\mathbbm{1}_{\rho_M(x)=0}\mathbbm{1}_{f(x)\neq y^*}}{\sum_{x \in X}(1-\rho_M(x))} & \textrm{s.t.} \frac{1}{n}\sum_{x \in X}\mathbbm{1}_{\rho_M(x)=0} > \mathcal{C}
\end{aligned}
\end{equation}
where $0<\mathcal{C}<1$ is a coverage threshold. The problem has thus shifted from having to determine cost $\mathcal{R}$ to having to select threshold $\mathcal{C}$ which is easier. It is important to note that if $\mathcal{C}=1$ the model reverts to a standard classifier with no rejection option. This coverage based formulation has been proven to be theoretically equivalent to the original cost model of Eq. \eqref{eq:l2rcmod} \cite{pmlr-v97-franc19a}. 

\subsubsection{Strengths and limitations of Learning to Reject}
In summary, the L2R paradigm allows for the development of ML models with a reject option, which is a foundational starting point for developing models able to interact with humans. Indeed, ML models equipped with the reject option can, in principle, reject exactly those instances that would yield a prediction error, and therefore call for human intervention only when strictly needed. However, the actual benefit of these techniques in a collaborative setting with humans has never been thoroughly investigated. Indeed, if human intervention is called only for those instances for which a decision is difficult, the human expert may find the same difficulties, and thus deem the model as not so useful for solving the task. Moreover, there are studies pointing to possible fairness issues when using classifiers with a reject option \cite{JonesSKKL21}.
\subsection{Learning to Defer}
\label{sec:L2D}
%\textcolor{black}{In this section we provide a thorough analysis of the L2D literature. In Section \ref{sec:l2dGeneral}, we begin by outlining the overall framework and the main dimensions along which L2D systems can be described. After that, we summarize the most noteworthy proposals categorized according to the adopted model architecture, namely \textit{staged} (Section \ref{sec:stagedL2D}) and \textit{joint learning} (Section \ref{sec:jointL2D}). In the latter case, we also separately analyze the settings of \textit{Multiple-Expert L2D} (Section \ref{sec:l2d:ME1J} and Section \ref{sec:l2d:MEnJ}. A few significant L2D systems that have been constructed using additional model architectures are briefly discussed in Section \ref{sec:otherL2D}. Finally, in Section \ref{sec:l2dLimitedLabels}, we address the realistic scenario of implementing a L2D system with limited human expert predictions.}
Learning to Defer (L2D) systems embed human knowledge directly into the training process of a ML model. The goal is to equip the model with the ability to call for human intervention on those instances where the human is likely to give accurate prediction and the machine is likely to fail.
\begin{tcolorbox}[colback=blue!5!white,colframe=blue!50!white,title=Example of melanoma detection: Learning to Defer,fonttitle=\small]
\small
    By comparing predictions made by the dermatologist on previous cases to the correct labels, an L2D system can be trained to determine which instances can be accurately predicted by AI and which are better handled by (a committee of) humans.
    For instance, in the case of melanoma detection, humans are expected to outperform machines on dubious or extremely difficult cases, mostly owing to their greater capacity for comprehending common sense and contextual information, as well as integrating external factors regarding the patient such as for example different skin complexions and other clinical data.
\end{tcolorbox}
In contrast to L2D, an L2R system learns its rejector policy only from the feature set $\mathcal{X}$ (e.g., the text of the message to be flagged) and, possibly, some properties of the predictor used by the AI system. L2D instead actively considers the human expertise in the task domain.
    \begin{table}[t!]
    \footnotesize
      \begin{tabular}{p{37mm} p{69mm} p{25mm}}
      %{ @{} l V{9cm} @{} }
        \toprule
        % & \textbf{Symbol} & \textbf{Values} & \textbf{Description} \\
        % \midrule
        \textbf{Architecture.} & \multicolumn{2}{p{95mm}}{\textit{Design according to which classifier and deferral policy are integrated.} }\\
        \midrule
        Staged learning      & Classifier and deferral policy are learnt in two subsequent steps. & \cite{Bansal2021IsTM,Raghu2019Med,Raghu2019TheAA,Wilder2020}\\
        Joint learning &  Classifier and deferral policy are learnt jointly. & \cite{Charusaie2022,Gao2021,Hemmer2022,Keswani2021,Liu2022,Madras18,Mozannar20,Mozannar2023,Pradier2021,Raman2021,Verma22,Verma2022Multi,Wilder2020}\\
        Others & Alternative solutions, e.g., iterative models. & \cite{De2020Reg,De2020Class,Okati2021}\\
        \bottomrule
        
        \textbf{Multiplicity.} & \multicolumn{2}{p{95mm}}{\textit{Number of human agents in the hybrid system.} } \\
        \hline
        Single  & One human agent. & \cite{Bansal2021IsTM,Charusaie2022,Liu2022,Madras18,Mozannar20,Mozannar2023,Okati2021,Pradier2021,Raghu2019Med,Raghu2019TheAA,Raman2021,Verma22,Wilder2020}\\
        Multiple (1 predicts) & One human is selected out of many. & \cite{Gao2021,Hemmer2022}\\
        Multiple ($j$ predict) &A subset of agents is selected out of many. & \cite{Keswani2021,Verma2022Multi}\\
        \bottomrule
        
        \textbf{Theoretical guarantees.} & \multicolumn{2}{p{95mm}}{\textit{The machine phase in which the interaction occurs.} }\\
        \midrule
        Fisher consistency  &  The optimization objective has the correct target. & \cite{Charusaie2022,Mozannar20,Verma22,Verma2022Multi}\\
        Classification-calibration &  Agents are given realistic uncertainty estimates. & \cite{Verma22,Verma2022Multi}\\
        Realizable consistency & The problem is well-defined under specific choices of the classifier and deferral hypothesis function spaces. & \cite{Mozannar2023}\\
        \bottomrule

        \textbf{Constraints.} & \multicolumn{2}{p{95mm}}{\textit{Additional conditions that the hybrid system should satisfy.} } \\
        \midrule
        Coverage  & Number of instances that can be deferred. & \cite{De2020Class,Mozannar20,Mozannar2023,Okati2021}\\
        Budget & Total cost to query human agents. & \cite{Raghu2019TheAA} \\
        Fairness & Metrics to guarantee algorithmic fairness. &  \cite{Keswani2021,Madras18,Mozannar2023}\\
        Others & Others, e.g., on the selection of human agents. & \cite{Keswani2021}\\
        \bottomrule
      \end{tabular}
      \caption{Properties of systems in the Learning to Defer paradigm.}
      \label{tbl:l2d:axes}
    \end{table}
\subsubsection{General formulation}
\label{sec:l2dGeneral}
%\paragraph{From L2R to L2D}
Keeping the same notation introduced in Section \ref{sec:generalModels}, we consider 
% [REPLACE] HDMS
Hybrid Systems
composed of a machine $M$ and a human $H$. Similarly to L2R, in L2D the machine $M$ is equipped with the possibility of abstaining from making a prediction. 
In addition, an L2D model also embeds a representation of the human agent $H$, thereby taking into account their estimated performance when assessing the act of deferral. By doing so, the human expertise $\mathcal{Z}$ is taken into account. 
%However, a representation of the human agent to whom the decision is deferred is here embedded in the model, which means that her (estimated) performance is considered when evaluating the action of deferral.
Nevertheless, note that the predictor $h:\mathcal{X}\times\mathcal{Z} \rightarrow \mathcal{Y}_H$ modeling $H$ is \textit{fixed},
%More in detail, a human agent $H$ is modeled as a \textit{fixed} predictor $h:\mathcal{X}\times\mathcal{Z} \rightarrow \mathcal{Y}_H$, with the same notation introduced in Section \ref{sec:generalModels}. Note that $h$ is \textit{fixed} 
meaning that L2D algorithms have no control nor visibility on the function $h$ itself; rather, they only have access to its image, that is, the set $\{Y_{H,i}\}_{i=1}^n\coloneqq \{h(x_i,z_i) \}_{i=1}^n$ of human predictions about the training data.

In analogy to Eq. \eqref{eq:l2rSystem}, a L2D system can be formulated as a function $f_{\rho_M}:\mathcal{X} \rightarrow \mathcal{Y}_M \bigcup \mathcal{Y}_H$ defined from the classifier $f:\mathcal{X}\rightarrow\mathcal{Y}_M$ and deferral policy $\rho_M:\mathcal{X}\rightarrow \{0,1\}$:  
%The quality of human predictions is quantified by the associated loss $\mathscr{L}_H:\mathcal{Y}\times\mathcal{Y}_H\rightarrow\mathbb{R}_{>0}$. %Without loss of generality, we assume that the images of $f$ and $h$ lie in the same output space $\mathcal{Y}$.
%In order to choose the best decision-maker for each instance, $M$ implements an additional \textit{deferral policy} $\rho_M:\mathcal{X}\rightarrow \{0,1\}$ such that, if $\rho_M(x)=1$, then $M$ defers the decision to $H$ incurring in an additional cost $c_H$, otherwise M computes $f(x)$ by paying a cost $c_M$, which usually accounts for the misclassification error only, and gives the final decision \cite{Mozannar20}. The composed model can be defined as $f_{\rho_M}:\mathcal{X} \rightarrow \mathcal{Y}_M \bigcup \mathcal{Y}_H$ such that:
%
\begin{equation*}
f_{\rho_M}(x) = \begin{cases*} 
    h(x)   & if $\rho_M(x)=1$\\
    f(x)       & otherwise 
             \end{cases*}
\end{equation*}
%
%Note that both the predictor $f$ and the deferral function $\rho_M$ are parameterized by $\theta\in\Theta$, that is, $f=f(\cdot\;;\theta)$, $\rho_M=\rho_M(\cdot\;;\theta)$ and consequently $\mathscr{L}_M = \mathscr{L}_M(\cdot\;;\theta)$. However, in order to not weight down the notation, in the following we omit the dependence of all these functions on $\theta$. 
%In its most general formulation, an L2D model is trained to learn optimal parameterized functions $f^*$ and $\rho_M^*$ that minimize the risk of the human-machine system as a whole:
%Equivalently, we seek the parameters $\theta^*$ for the classifier $f$ and the deferral function $\rho_M$ that solve the empirical risk minimization principle given a training set $\mathcal{D}=\{X_i,Y_i\,Y_{H,i}\}_{i=1}^n$:
% \begin{equation}\label{eq:riskL2d}
%     f^*, \rho^*= \arg \min_{f,\rho_M}\mathscr{R}_{\text{defer}}[f,\rho_M] = \arg \min_{f,\rho_M}\mathbb{E}_{X,Y,Z}\left[\mathscr{L}_{\text{defer}}(Y,Y_M,Y_H,\rho_M)\right]
% \end{equation}
The goal of L2D is to find the classifier-rejector pair $(\hat{f},\hat{\rho}_M)$ that minimizes the system loss $\mathscr{L}_{\text{defer}}$.
This
%where $\mathcal{L}_{\text{defer}}$ represent the L2D system loss. Such loss
can be expressed as the summation of the machine loss $\mathscr{L}_M$ and the human loss $\mathscr{L}_H$:
\begin{equation}
\label{eq:L2DsystemLoss}
% \begin{split}
%     \mathscr{L}_{\text{defer}}(Y,Y_M,Y_H,\rho_M) & \coloneqq \mathbbm{1}_{\rho_M(X)=0}\cdot\mathscr{L}_M(Y,Y_M)+\mathbbm{1}_{\rho_M(X)=1}\cdot\mathscr{L}_H(Y,Y_H)
%     %+ \lambda \mathcal{R}(Y,Y_H,Y_M,\rho_M) 
%     = \\
%      & =\underbrace{(1-\rho_M(X))}_{M\text{ predicts}}\underbrace{\mathscr{L}_M(Y,Y_M)}_\text{machine cost} +\underbrace{\rho_M(X)}_{\text{defer to }H}\underbrace{\mathscr{L}_H(Y,Y_H)}_\text{human cost}
%      %+ \underbrace{ \lambda \mathcal{R}(Y,Y_H,Y_M,\rho_M)}_{\text{regularization term}}
% \end{split} 
    \mathscr{L}_{\text{defer}}(Y^*,Y_M,Y_H,\rho_M) \coloneqq 
     \underbrace{\mathbbm{1}_{\rho_M(X)=0}}_{M\text{ predicts}}\underbrace{\mathscr{L}_M(Y^*,Y_M)}_\text{machine cost} +\underbrace{\mathbbm{1}_{\rho_M(X)=1}}_{\text{defer to }H}\underbrace{\mathscr{L}_H(Y^*,Y_H)}_\text{human cost} 
     %+ \underbrace{\lambda \mathcal{R}(Y^*,Y_M,Y_H,\rho_M)}_\text{regularization term}
\end{equation}
where $\mathbbm{1}$ denotes the indicator function. 
%$\lambda\geq 0$ is a regularization \textcolor{black}{parameter}, and $ \mathcal{R}(Y,Y_H,Y_M,\rho_M)$ is a regularization term added to help reduce overfitting or to model desirable priors, penalties, or constraints. 
Note that the optimization problem has $Y_M$ and $\rho_M$ as unique learnable parameters since both $Y_H$ and $Y^*$ are fixed (i.e., they belong to the training data). 
%In the subsequent discussion, we will assume, without loss of generality, that we are referring to the non-regularized scenario where $\lambda=0$. However, we will highlight some notable regularisation terms presented in the literature 
%Nevertheless, we will outline some common regularizers used in L2D 
%at the end of this section. 
%For this reason, in the following we simplify the notation of the system loss by suppressing the constant terms and referring to the predictors instead of their image, that is, we write $\mathcal{L}_{\text{system}}(m,r)$ instead of $\mathcal{L}_{\text{system}}(Y,Y_H,Y_M,r(X))$.
In general, the individual losses $\mathscr{L}_M$ and $\mathscr{L}_H$ can take several forms to account for different "costs", such as the misprediction error as in the 0-1 loss, or the cost of querying the human agent.
% are defined instance-wise as follows:
% \begin{equation*}
%     \mathscr{L}_M(y^*,f(x))=\mathbbm{1}_{f(x)\neq y^*}+c_M(y,f(x)) 
% \end{equation*}
% \begin{equation*}
%     \mathscr{L}_H(y^*,y_H)=\mathbbm{1}_{y_H\neq y^*}+c_H(y^*,y_H).
% \end{equation*}
% In both cases, the first part accounts for the misprediction errors, while the second \textcolor{black}{represents a cost function $c_{(\cdot)}$ which}
% accounts for any additional penalty term, such as the cost of querying a human expert in case of deferral. 

Notably, when there exists a constant $\mathcal{R}> 0$ such that $\mathscr{L}_H(y^*,y_H)=\mathcal{R}$ for all $(y^*,y_H)\in (Y^*,Y_H)$, then the loss \eqref{eq:L2DsystemLoss} matches the rejection loss described in \cite{cortes2016l2r} for the L2R framework 
%(the mathematical proof is given in 
\cite{Madras18}:
%)

\begin{equation}
\label{eq:l2rLoss}
     \mathscr{L}_{\text{reject}}(Y^*,Y_M,\rho_M) \coloneqq \mathbbm{1}_{\rho_M(X)=0}\mathscr{L}_M(Y^*,Y_M)+\mathcal{R}\mathbbm{1}_{\rho_M(X)=1} 
     %+ \lambda \mathcal{R}(Y^*,Y_M,\rho_M)
\end{equation}
Note that Eq. \eqref{eq:l2rLoss} and Eq. \eqref{eq:l2robj} are equivalent under the assumption of using the 0/1 cost model for prediction/rejection.

\textbf{\textit{Optimization constraints.}}
Depending on the specific context of use, the application of specific constraints may be necessary for hybrid systems. This objective is commonly accomplished by incorporating regularization terms into the system loss or by imposing specific bounding conditions. Examples of such constraints include: 

\begin{itemize}
    \item Coverage or triage level \cite{De2020Class,Mozannar20,Mozannar2023,Okati2021}: the number of instances that can be deferred.
    \item Fairness metrics \cite{Keswani2021,Madras18,Mozannar2023}: for instance, the Minimax Pareto Fairness criterion \cite{Martinez20} or the equalized odds metric with respect to a protected attribute. 
    \item Budget \cite{Raghu2019TheAA}: total cost that can be allocated to query human agents.
\end{itemize}

\textbf{\textit{Model architectures.}}
\label{sec:stagedVsJoint}
L2D systems typically adhere to either of two general designs, referred to as \textit{staged learning} and \textit{joint learning}, which vary in terms of when the classifier and rejector are learned. In the former case, the algorithmic process starts by learning the classifier and only subsequently fits the deferral policy on top of it.
On the other hand, in a joint learning setting the classifier and rejector are learnt simultaneously through the direct minimization of the system loss \eqref{eq:L2DsystemLoss}.
While most of the proposals documented in the literature can be categorized as staged or joint learning models, a few exceptions also exists that do not fit in either category (Section \ref{sec:otherL2D}).

\textbf{\textit{Number of human agents.}}
L2D systems can be characterized along another dimension, that is, the size of the pool of human agents to which the decision can be deferred to. 
Whenever the number of these is greater than one, term \textit{Multiple-Expert L2D} (L2D-ME) is used, as opposed to to \textit{Single-Expert L2D} (L2D-SE or L2D), which considers one human only. An example where L2D-ME modeling may be more suitable is in a medical setting, where a critical decision regarding a complex case could be made either by an automated classifier or by one or more doctors chosen from a team of experts with potentially diverse expertise and opinions.
%Thus far, our discourse about L2D has primarily revolved around scenarios in which a single human is involved. Nevertheless, there are various situations in which a deferral system is expected to be enacted within a context that involves a \textit{set} of multiple experts. For example, such a scenario may occur within a medical context, wherein a crucial decision regarding a complex case could be made either by an automated classifier or by one or more doctors selected from a pool of experts with potentially varying expertise and bias. 
As compared to L2D-SE, the formalization of L2D-ME makes the deferral function more complex in nature. Specifically, it should not only determine \textit{when} to defer, but also to \textit{which} human agent(s) \cite{Verma2022Multi}. Furthermore, 
%the latter problem can appear in two distinct manners More precisely, 
the deferral policy can be designed in a manner that allocates predictions to either 
%, the optimal performance of the hybrid system can be attained either by choosing the 
\textit{one} single agent or to a
%by selecting the best 
\textit{subset} of agents from the available pool.
%Section \ref{sec:l2d:ME1J} and \ref{sec:l2d:MEnJ}, respectively.

% \paragraph{Limitations.}
% As introduced in Section \ref{sec:generalModels}, a first difficulty in approximating the optimal classifier- rejector pair is the non-convex shape of the loss function. Such a problem is handled by optimizing instead a \textit{surrogate} loss function $\widetilde{\mathscr{L}}$. For a formal discussion of the properties of surrogate losses in context of L2D, refer to Section \ref{sec:surrogate}.

% A second limitation of the L2D framework is already evident from the formulation of the optimization problem, namely, it requires \emph{all} predictions of the human expert for \emph{all} instances in the training set \cite{Leitao2022}. This condition can hardly be satisfied in real-world scenarios \cite{Hemmer2022,Charusaie2022}.

\subsubsection{Staged learning architectures} \label{sec:stagedL2D}
%\paragraph{General formulation.}
In L2D models characterized by staged learning architectures, the classifier $f$ and deferral function $\rho_M$ are learnt separately. Specifically, algorithms in this category first fit a classifier on the training dataset, then they learn a second model that predicts the probability that the human makes a mistake on the same dataset, and finally they defer based on which has the lowest error probability instance-wise. 
%In other words, a staged learning architecture involves two steps: one first estimates the optimal classifier given the training data, and then $\hat{f}=\arg\min_{f\in\mathcal{F}_M}\hat{\mathscr{R}}_n[f]$ and then the optimal rejector $\hat{\rho}_M$ by minimizing the joint 0-1 loss keeping the classifier fixed, that is: $\hat{\rho}_M=\arg\min_{\rho_M\in\mathcal{D}}\mathscr{R}_{\text{defer}}[\hat{f},\rho_M]$. 

%\paragraph{Models.}
%A naive baseline for L2D via staged learning is the \textit{Selective Prediction} suggested in \cite{Mozannar2023}. Here, the first step consists in fitting the classifier on the whole training set to minimize the cross-entropy loss, while an optimal threshold for deferral is successively learnt on the validation set based on the confidence of the class predicted by the classifier so to maximize the system accuracy score. At test time, an instance is deferred to the human expert if the classifier confidence is lower than such a optimal threshold. Note that \textit{confidence} simply refers here to the highest of the prediction probabilities estimated for all classes. 

For instance, \citet{Raghu2019TheAA} developed a basic heuristic for L2D consisting of two independent models trained on the full dataset: a multiclass classifier representing the machine agent, and a binary classifier representing the correctness of the human agent. At inference time, an instance is deferred to the human if the predicted classifier error probability is higher than that of the human. In case of coverage constraints, then the samples whose difference between human and classifier error probability is higher are chosen first. 
Interestingly, the authors also suggest a reduction of L2D-ME to L2D-SE by modeling the human subsystem in terms of average disagreement between human agents on each single prediction. This approach has been further developed in \cite{Raghu2019Med}.
%The task of accurately estimating the disagreement among human experts has been further addressed by \citet{Raghu2019Med}, where the authors introduce and compare two methods for predicting uncertainty scores describing the strength of expert disagreement on each instance. 
%Finally, as pointed out in \cite{De2020Reg}, the method developed in \cite{Raghu2019TheAA} suffers from two major limitations in addition to those mentioned in Section \ref{sec:stagedVsJoint} regarding staged learning in general: the classifier is trained on the full training set as in full automation, and the algorithm does not enjoy theoretical guarantees. 

Another common baseline for staged learning is the model proposed by \citet{Bansal2021IsTM}, who described a staged learning setting aimed to maximize the expected utility of the system, which is measured in terms of the accuracy of the final decision, the cost of deferring, and the individual accuracy of both the human and machine component.  Differently from other L2D models, 
%the authors claimed 
this method has been claimed to be user-initiated, since the action of deferral is triggered through an (over-simplified) threshold-based policy that represents the humans’ mental model of the AI.
%Indeed, the action of deferral is triggered through a threshold-based policy which represents the humans’ mental model of the AI system. It is important to acknowledge that the description of this mental model is oversimplified, as it solely considers the minimum value of predictive confidence that the user is willing to rely on to accept the prediction. Furthermore, the authors assumed the human agent is rational, in the sense that they will accept or reject the machine prediction based on the option that will yield the highest expected utility. 

Finally, a third relevant staged learning method known as \textit{fixed value of information approach} has been proposed in \cite{Wilder2020}. It consists in training independently three probabilistic models describing, respectively, the distribution of the label given the input data, the human predictions given the input data, and the label given both the input data and human predictions. At inference, the deferral policy evaluates the estimated expected utility of the classifier in two scenarios: when the human is not consulted and when the human is queried, while also taking into account the distribution of human predictions and a constant cost for querying the human. %Nevertheless, empirical evidence indicates that a similar approach, outlined in Section \ref{sec:jointL2D}, exhibits superior performance due to its implementation of a joint learning architecture.

As noted by \citet{Charusaie2022}, the staged learning approach presents some important advantages: first of all, it is suited for convenient implementation, since already known appropriate algorithms can be adopted to solve the two stages separately. Secondly, theoretical and experimental results suggest that it outperforms the joint learning approach in realistic scenarios where only a limited portion of data is labeled by the human agent. In these cases, the classifier $f$ can still be optimized over the full dataset; on the other hand, in joint learning $f$ can be learnt from the subset of human labeled data only, thus leading to a reduction int performance dependent on the proportion of unlabeled data.  
However, \cite{Charusaie2022} also pointed out that staged learning is sub-optimal with respect to joint learning and provide both theoretical and experimental results showing the existence of a performance gap between the two approaches in terms of model complexity.   

\subsubsection{Joint learning architectures}
\label{sec:jointL2D}
%\paragraph{General formulation.}
\begin{figure}[t!]
    \centering
    \resizebox{0.75\textwidth}{!}{%
\begin{tikzpicture}[x=2.2cm,y=1.4cm]
  \message{^^JNeural network without text}
  \readlist\Nnod{4,2,4} % array of number of nodes per layer
  
  \message{^^J  Layer}
  \foreachitem \N \in \Nnod{ % loop over layers
    \def\lay{\Ncnt} % alias of index of current layer
    \pgfmathsetmacro\prev{int(\Ncnt-1)} % number of previous layer
    \message{\lay,}
    \foreach \i [evaluate={\y=\N/2-\i; \x=\lay; \n=\nstyle;}] in {1,...,\N}{ % loop over nodes
      
      % NODES
      \ifnum\lay=3
        \ifnum\i=1
            \node[node \n] (N\lay-\i) at (\x,\y) {$g_1$};
        \else
            \ifnum\i=2
                \node[node \n] (N\lay-\i) at (\x,\y) {$\dots$};
            \else
                \node[node \n] (N\lay-\i) at (\x,\y) {$g_n$};
            \fi
        \fi
      \else
        \ifnum\lay=1
            \node[node \n] (N\lay-\i) at (\x,\y) {};
        \else
            \ifnum\i=1
                \node[node \n, label={[label distance=1.45cm, align=center]
                above:{\large\textbf{Classifier-Deferral} \\ \large\textbf{AI system}}
                }] (N\lay-\i) at (\x,\y) {};
            \else
                \node[node \n] (N\lay-\i) at (\x,\y) {};
            \fi
        \fi
      \fi
      
      % CONNECTIONS
      \ifnum\lay>1 % connect to previous layer
        \foreach \j in {1,...,\Nnod[\prev]}{ % loop over nodes in previous layer
          \draw[connect,white,line width=1] (N\prev-\j) -- (N\lay-\i);
          \draw[connect] (N\prev-\j) -- (N\lay-\i);
          %\draw[connect] (N\prev-\j.0) -- (N\lay-\i.180); % connect to left
        }
      \fi % else: nothing to connect first layer 
    }
  }

  % Defer output node
  \node[node 4, label=\rotatebox{-90}{$<$}] (N3-4) at (3, -2.0) {$g_\bot$};

  % Human defer node
  \node[shape=rectangle,draw=black,inner ysep=2mm] (N4-1) at (4, -0.0) {Machine};
  \node[shape=rectangle,draw=black,inner ysep=2mm] (machine_predict) at (5, -0.0) {Prediction};
  
  \node[node 5, inner ysep=2mm] (N4-2) at (4, -2.0) {Human};
  \node[shape=rectangle,draw=black, inner ysep=2mm] (human_predict) at (5, -2.0) {Prediction};
  % Connection
  \draw[connect] (N3-4) -- (N4-2);
  % \draw[connect] (N3-1) -- (N4-1);
  % \draw[connect] (N3-2) -- (N4-1);
  % \draw[connect] (N3-3) -- (N4-1);

  \draw[connect] (N4-2) -- (human_predict);
  \draw[connect] (N4-1) --  node[above=2.3cm,midway,color=black]{\large\textbf{Prediction}} (machine_predict);

  \draw [pen colour={red}, decorate, decoration = {calligraphic brace, amplitude=15pt}, ultra thick] (3.3,1.3) --  (3.3,-1.3) node [red,midway,xshift=-0.1,rotate=-90] {argmax};
  
    % Post
    \node[shape=rectangle,draw=black,align=left,inner ysep=3mm,     label={[label distance=1cm, align=center]above:{
    \large\textbf{Input data}\\e.g., social media post}}] (outer) at (-1, -0.5) {
    \hspace{10mm}@Abc
    \\[2mm]
        \begin{tikzpicture}[align=center]
            \node[shape=rectangle,draw=black,align=center, inner ysep=3mm, inner xsep=2mm] (message) 
            %at (-1, -0.5) 
            {``Goddam tarantula NEXT to\\my bed and a stinking bug\\ON my bed?! suicidal fr no lie''};
        \end{tikzpicture}
    \\ \hspace{34mm}x \Heart y\leftthumbsup \vspace*{-2mm}};
    \node[node 1, align=center, inner ysep=2mm] (user) at (-1.9, 0.25) {\baselineskip=5pt. .\\ $\smile$\par };
    
  %\node[shape=rectangle,draw=black,align=center] (message) at (-1, -0.5) {``Goddam tarantula NEXT to\\my bed and a stinking bug\\ON my bed?! suicidal fr no lie''};
  
  \node[shape=circle,draw=white,align=center] (phantom_connect) at (0.8, -0.5) {};
  \draw[connect] (outer) -- (phantom_connect); 
\end{tikzpicture}
    }
    \caption{Overview of the joint learning architecture for the Single-Expert Learning to Defer (L2D-SE) setting, illustrated in the application of flagging online contents for moderation. Adapted from \cite{Mozannar2023}.}
    \label{fig:l2dJoint}
\end{figure}
In L2D systems characterized by a joint learning architecture the classifier $f$ and deferral function $\rho_M$ are learnt simultaneously. 
%Typically, this is achieved by minimizing a unique surrogate loss function representing the system as a whole after a reduction to a cost-sensitive loss \cite{Mozannar20}. 
In order to implement this design, the task is shaped as a $K+1$ multiclass problem over an augmented label space $\mathcal{Y}^{\emptyset} \coloneqq \mathcal{Y}_M\cup\{\emptyset\}$, where $\mathcal{Y}_M=\{1,\ldots,K\}$. %where we remind that the class $\emptyset$ correspond to the action of deferral. 
In particular, we set $\textbf{g}=(g_1,\ldots,g_K,g_\emptyset)$ to be the set of real-valued scoring functions 
$g_i:\mathcal{X}\rightarrow\mathbb{R}$, such that $g_\emptyset$ returns the human predictions $Y_H$, while the classifier and rejector are defined, respectively, as:
\begin{equation}
    f(x) = \arg\max_{i\in\mathcal{Y}_M} g_i(x)
    \quad \quad \quad \quad \quad
    \rho_M(x)= \begin{cases*} 
    1   & if $\max_{i\in\mathcal{Y}_M} g_i(x)\leq g_\emptyset (x)$\\
    0       & otherwise.
             \end{cases*}
\end{equation}

%\mathbbm{1}_{\max_{i\in\{1,\ldots,K\}}g_i(x)\leq g_\emptyset (x)}$$
%samples are classified according to $\hat{y} = \arg\max_{i\in\{1,\ldots,K, \emptyset\}} g_i(x)$, where deferral occurs whenever $g_\emptyset$ is the maximum argument.  
In general, the optimal classifier-rejector pair is found by minimizing the system loss expressed in Eq. \eqref{eq:L2DsystemLoss}.
%after having set the cost for machine prediction equal to the machine loss $\mathscr{L}_M$ and the cost for human prediction equal to the human loss $\mathscr{L}_H$. 
However, the function \eqref{eq:L2DsystemLoss} is often computationally hard to optimize.  
%a first difficulty in approximating the optimal classifier- rejector pair is the non-convex shape of the loss function. 
Such a problem is addressed by replacing \eqref{eq:L2DsystemLoss} with a \textit{surrogate loss} $\widetilde{\mathscr{L}}_\text{defer}$ that is easy to optimize and is chosen to guarantee specific properties with respect to the original loss $\mathscr{L}_\text{defer}$ (refer to the Appendix for a formal discussion of desirable properties of surrogate loss functions in the context of L2D).
Hence, in case of appropriate choices of the human and machine surrogate loss functions, the joint learning HS gives theoretical guarantees for optimal performance.

%At inference time, for a given sample, the prediction probabilities of all classes, including $\emptyset$, are first computed; then, the machine prediction, computed as the class with maximum probability among all classes excluded $\emptyset$, is compared with the prediction probability of $\emptyset$, thus defining a deferral policy according to which the decision is deferred to the human whenever the difference between the human and classifier prediction is greater or equal than rejection threshold, which is usually set to $0$ if not differently optimized (e.g., in the realizable surrogate loss approximation).

%The advantage of joint learning models with respect to staged learning ones is that, for appropriate choices of the human and machine loss functions, the system surrogate loss satisfies the properties described in Section \ref{sec:surrogate} and thus provides theoretical guarantees for optimal performance. However, as already pointed out, this approach suffers from the limitation of requiring access to a fully human labeled dataset, which is often unrealistic \cite{Charusaie2022}.

\textbf{\textit{Single-Expert L2D.}}
\label{sec:L2Dsingle}
Most of the literature on joint learning models for L2D-SE focuses on the \textit{cost-sensitive formulation} of the problem over an augmented label space $\mathcal{Y}^\emptyset$ developed by \citeauthor{Mozannar20} \cite{Mozannar20} and illustrated in Figure \ref{fig:l2dJoint}.
This setting considers random costs $\textbf{c}=\{c_1,\ldots,c_{K+1}\}\in\mathbb{R}_{>0}^{K+1}$ where each component $c_i$ represents the cost of predicting the label $i\in\mathcal{Y}^\emptyset$. In this section, we review and categorize the most relevant proposals according to their statistical properties. 

\textit{Fisher Consistency} (FC) has been described as a minimal requirement that surrogate loss functions should satisfy to achieve reasonable performance, since it posits that, if an estimator were computed using the complete population instead of a sample, it would yield the true value of the estimated parameter \cite{Lin2004}. 
    To the best of our knowledge, FC approximation of the 0-1 loss in the L2D setting have been implemented (up to adaptations) only by \citet{Mozannar20,Charusaie2022,Verma22}.
    In particular, the surrogate loss $\mathscr{L}_{CE}^{\alpha}$ \cite{Mozannar20} consist of a generalization of the cross-entropy loss with the costs corresponding to multiclass misclassification, where the cost of the $K+1$ class represents the action of deferral, and $\alpha\in\mathbb{R}_{>0}$ is a weighting parameter that modulates deferral. When $\alpha=1$, $\mathscr{L}_{CE}^{\alpha}$ has FC and can be expressed as:
    \begin{equation}
        \mathscr{L}_{CE}^{1}(\textbf{g};x,y^*,y_H)\coloneqq -\log\left(\frac{\exp(g_{y^*}(x))}{\sum_{y\in\mathcal{Y^\emptyset}}\exp(g_{y}(x))} \right) - \mathbbm{1}_{y_H=y^*} \log\left(\frac{\exp(g_\emptyset(x))}{\sum_{y\in\mathcal{Y^\emptyset}}\exp(g_{y}(x))} \right) 
    \end{equation}
    intuitively, the first term maximizes the scoring function associated with the true label, while the second maximizes the rejection (scoring) function but only if the human’s prediction is correct. 
    Notably, \citet{Charusaie2022} \textcolor{black}{presents a unified framework that allows the use of any consistent multiclass loss for constructing a consistent surrogate for L2D, thus generalizing prior work \cite{Cao2024,Mozannar20,Verma22}.}
    %Namely, given any consistent surrogate loss function $\widetilde{\mathscr{L}}$ of the 0-1 loss for multiclass classification, then the surrogate 
    %$$\widetilde{\mathscr{L}'}(\textbf{c},f(\textbf{x}))\coloneqq\sum_{i=1}^{K+1}(\max_j c_{(j)}-c_{(i)})\widetilde{\mathcal{L}}(i,f(\textbf{x}))$$ 
    %is a also a consistent surrogate for cost-sensitive learning and thus for L2D.

    A few adaptations have been proposed to enhance the L2D algorithms based on the surrogate $\mathscr{L}_{CE}$, with the aim of better capturing specific properties. These include: 
    
    \begin{itemize}[leftmargin=8mm, topsep=2mm]
        \item \textit{Learning to Defer with Uncertainty} (LDU), where the deferral policy accounts for the epistemic uncertainty of the model (i.e., the uncertainty resulting from limited data availability and lack of knowledge about the system of interest) \cite{Liu2022};
        \item The customization of the model to suit the expertise of particular human agent, which however requires the availability of supplementary data that has been annotated by that particular human \cite{Raman2021}.
        \item \textcolor{black}{\textit{Label-smoothing-free loss} (LSF). \citet{Narasimah2022} demonstrated that consistent loss functions experience \textit{underfitting} when the additional cost of deferring to the expert is non-zero, as this scenario introduces a label smoothing term, which results in a flattened training distribution. \citet{Liu2024} propose a novel loss formulation which tackles this issue while preserving statistical consistency. Moreover, the authors demonstrate that current representative surrogate losses for L2D \cite{Cao2024,Mozannar20,Verma22} can be devoid of label smoothing by plugging their base multiclass losses into their suggested loss formulation, which is as follow:}
        $$\mathscr{L}_\psi^{LSF}(\textbf{g};x,y^*,y_H)=\psi(\textbf{g}(x), y^*)) + c\mathbbm{1}_{y_H\neq y^*} \min_{y\in \mathcal{Y}_M}(\textbf{g}(x), y)+(1-c) \mathbbm{1}_{y_H = y^*} \psi(\textbf{g}(x), K+1))$$
    \end{itemize}
     
\textit{Confidence calibration} refers to the property of an estimator (e.g., a probabilistic classifier) to produce a predictive distribution that is consistent with the empirical frequencies observed from realized outcomes \cite{Dawid1982}. \citet{Verma22} proposed a surrogate loss $\mathscr{L}_{OvA}$ \cite{Verma22} that satisfies both Fisher consistency and classification-calibration. This solution consists in solving the L2D problem via a One-vs-All classification method which breaks down the original $K+1$ classes into $K+1$ binary classifier models. The resulting objective function to be optimized is thus a surrogate loss function composed of different logistic loss components, each accounting for the error on one of the $K+1$ different classes. 
        \begin{equation*}
        \mathscr{L}_{OvA}(\textbf{g};x,y^*,y_H)\coloneqq \phi[g_{y^*}(x)]+\sum_{\substack{y\in\mathcal{Y}_M\\ y\neq y^*}}\phi[-g_y(x)] + \phi[-g_\emptyset(x)] + \mathbbm{1}_{y_H=y^*} (\phi[g_\emptyset(x)]-\phi[-g_\emptyset(x)])
    \end{equation*}
    where $\phi:\{0,1\}\times\mathbb{R}\rightarrow\mathbb{R}_{>0}$ is a binary surrogate loss (e.g., the logistic loss).
    Experimental findings show that $\mathscr{L}_{OvA}$ results in better calibrated models w.r.t. ones trained with $\mathscr{L}_{CE}$, with competitive performance w.r.t. other L2D baselines \cite{Bansal2021IsTM,Mozannar20,Okati2021,Raghu2019TheAA}. 

    % Boundedness
\textcolor{black}{A \textit{bounded} function $f$ is a function defined on a bounded set $X$ of real or complex values, meaning there exists a real integer $N$ such that $|f(x)|\leq N \;\forall x\in X$. While exploring the confidence calibration properties of $\mathscr{L}_{CE}$, \citet{Verma22} found that this loss is also unbounded, as it can exceed values larger than one, hence failing to adequately calibrating the expert correctness. To tackle this issue, \citet{Cao2024} offer a statistically consistent \textit{asymmetric}\footnote{An \emph{asymmetric} learning setting here refers to one characterized by unequal misclassification costs or training data imbalance \cite{Scott2012}} softmax-based surrogate loss that generates reliable estimates without the miscalibration and unboundedness issues that characterize $\mathscr{L}_{CE}$. It has the following formulation:}
    $$\mathscr{L}^{ASM}_\psi(\textbf{g};x,y^*,y_H)=-\text{log}(\psi_y(g(x))) - \mathbbm{1}_{y_H\neq y^*} \text{log}(1 - \psi^{K+1}(x)) -\mathbbm{1}_{y_H = y^*} \text{log}(\psi^{K+1}(x)).$$
    \textcolor{black}{where $\psi$ is an asymmetric softmax function obtained from the standard softmax function by adding an asymmetry w.r.t. the additional $k+1$ class of the augmented label space. Moreover, the authors also discuss the possibility of extending $\mathscr{L}_\psi^{ASM}$ to the multi-expert L2D setting.\\
    A different formulation of bounded loss function has been given by \citet{Wei2024} while studying the possibility of adopting an alternative Bayes optimality definition, that is, the minimiser of the Bayes risk for which any surrogate loss function is demonstrated to be consistent. Differently from the one proposed in current representative L2D models \cite{Mozannar20}, this definition accounts for dependence patterns between humans and models. Specifically, it introduces a novel deferral principle that assesses deferral according to the dependence pattern identified in training data, hence bypassing the necessity for confidence estimation.
    Motivated by their formulation of \textit{dependent Bayes optimality}, \citet{Wei2024} also present a novel \textit{Dependent Cross-Entropy} (DCE) loss that is consistent and capable of inducing a bounded confidence estimator for the expert.
}

$(\mathcal{F}_M,\Rho_M)$-\textit{Realizable consistency} is a property that refines the notion of FC by addressing the optimization process over restricted hypothesis classes $\mathcal{F}_M$ and $\Rho_M$ for the predictor and deferral function. For instance, the surrogate $\mathscr{L}_{RS}$ \cite{Mozannar2023} is differentiable, non-convex, and realizable $(\mathcal{F}_M,\mathcal{\Rho}_M)$-consistent for classes $\mathcal{F}_M$ and $\mathcal{\Rho}_M$ closed under scaling:
    \begin{equation*}
        \mathscr{L}_{RS}(\textbf{g};x,y^*,y_H)\coloneqq -2\log \left( \frac{\exp(g_{y^*}(x))+\mathbbm{1}_{y_H=y^*}\exp(g_\emptyset(x))}{\sum_{y\in\mathcal{Y^\emptyset}}\exp(g_{y}(x))}\right)
    \end{equation*}

    %In the attempt to overcome such issues, \citet{Mozannar2023} also design the surrogate $\mathscr{L}_{RS}$, which they proved to be optimal in the realizable setting of L2D with halfspaces, to perform well with non-linear predictors, and which they found it could be efficiently minimized to a local optimum. Importantly, $\mathscr{L}_{RS}$ is differentiable, non-convex, and realizable $(\mathcal{F}_M,\mathcal{D})$-consistent for classes $\mathcal{F}_M$ and $\mathcal{D}$ closed under scaling; however, the authors could not prove or disprove whether it is also Fisher-consistent. Finally, in order to avoid underfitting the target, \citet{Mozannar2023} also suggest a variant $\mathscr{L}_{RS}^{\alpha}$ of the realizable surrogate function which uses a hyperparameter $\alpha \in [0, 1]$ to write a convex combination of $\mathscr{L}_{RS}$ and the cross entropy loss for the classifier; as a result, points that are deferred to the human can still contribute to the classifier training and thus improve the sample efficiency. However, since $\mathscr{L}_{RS}^{\alpha}$ is no longer realizable consistent, a threshold to binarize the rejection score is also learned at training time. 
    
    \noindent Other formulations of joint learning L2D-SE also exist. 
    %are based on a \textit{Mixture-of-Experts} (MoE) setting involving a human and a machine expert. 
    This is the case of:
    \begin{itemize}
        \item the seminal paper by \citet{Madras18}, which presents a framework for addressing the L2D problem using a Mixture-of-Experts (MoE) approach, with the deferral policy acting as a gating function. The classifier and deferral function are learned together by negative log-likelihood minimization over the augmented label space $\mathcal{Y}^\emptyset$. An alternative version of the algorithm is also introduced, wherein a regularization component is added to the system loss to account for a fairness. Unfortunately, this method was proven to not have FC \cite{Mozannar20}.
        \item \textit{Preferential MoE} \cite{Pradier2021}, a variant of \cite{Madras18} where human knowledge is encoded in the form of decision rules that should be followed as much as possible, that is, whenever they are applicable and do not decrease the system performance. The algorithm first checks the applicability of the available rules and, in case of positive response, a deferral function selects whether to rely on the human or machine prediction based on their performance. 
        %The Preferential MoE problem is formulated as a two-player game, one that optimizes the gating function with the objective of maximizing the number of human-based decisions, and the other that optimizes the performance of the classifier by minimizing the system loss function. 
        %The authors suggest two algorithms with theoretical guarantees to solve such a game: a log-barrier method applied to a combined formulation of the two objective functions, and a projected gradient descent that minimizes each objective alternately. Notably, 
        Notably, the deferral function is chosen to be interpretable (e.g., a linear classifier or decision tree) guaranteeing transparency in the selection of the human agent, and also highlighting reasons for forgoing the human-based rules.

        \item In the \textit{joint value of information method} \cite{Wilder2020}, the three probabilistic models already introduced in the \textit{fixed value of information method} described in the Section \ref{sec:stagedL2D} are trained together through a single neural network which includes a final Platt calibration layer that guarantees the estimation of meaningful expected utilities. Experimental findings show that joint learning yields greater advantages compared to analogous staged learning method.

        \item The \textit{Mixed Integer Linear Program} (MILP) \cite{Mozannar2023} is a scheme to exactly minimize the mis-classification error of the HS.    
        %analyzed the problem of solving L2D using halfspaces, that is, by finding halfspace $f$ and halfspace $\rho_M$ that approximately minimize the system error, which they proved to be computationally hard even in the realizable setting. Hence, they designed an exact algorithm based on mixed integer linear program \textit{MILP} formulation, and a novel surrogate loss function $\mathscr{L}_{RS}$ that both obtain better empirical performance than other existing surrogate approaches. 
        %In particular, \textit{MILP} provides an exact solution to the linear L2D problem 
        It comes with generalization bounds and allows to provably and easily integrate any linear constraints on the variables. However, it suffers from two limitations: it is computationally expensive and it does not generalize to non-linear predictors.
    \end{itemize}

\textbf{\textit{Multiple-Expert L2D (1 out of $\mathbf{J}$).}}
\label{sec:l2d:ME1J}
In this first scenario of L2D-ME, the goal of the multi-expert deferral policy $\rho^{ME}_M$ is to choose either the classifier or \textit{exactly one} human agent from the set of $J$ available ones. Hence, $\rho^{ME}_M$ takes the form of $\rho^{ME}_M:\mathcal{X}\rightarrow\{0,1,\ldots,J\}$, where $\rho^{ME}_M(x)=0$ means that the classifier decides, while $\rho^{ME}_M(x)=j$ for $j\neq0$ indicates that the decision is deferred to the $j^{\text{th}}$ human agent.\citet{Hemmer2022} adopt a mixture of experts (MoE) approach whith the deferral policy serving 
    %in order to tackle the L2D-ME problem. The authors propose a method where a classifier is trained simultaneously with a deferral function. The latter is responsible for determining whether instances should be assigned to human experts or the classifier. Therefore, it serves 
    as a gating function that assigns each instance either to the predictor or one specific human agent. The joint learning of the classifier and deferral function is carried out through a surrogate loss function based on the negative log-likelihood of the system. However, subsequent work \cite{Verma2022Multi} has proven this surrogate to be not FC and proposed instead 
%    \textbf{\citet{Verma2022Multi}} have successfully addressed the theoretical fallacies identified in \cite{Hemmer2022}. They have proposed 
    two surrogate loss functions, namely one based on cross-entropy and one on the One-vs-All classification, which are consistent with the 0-1 loss in the L2D-ME setting and extend their single-expert analogue. The experimental findings indicate that the OvA-trained model frequently achieves superior performance compared to both the cross-entropy variant and the MoE baseline \cite{Hemmer2022}. Additionally, it exhibits better calibration in terms of the correctness of agents' decisions.
\textcolor{black}{The properties of realizable consistency with respect to a certain hypothesis space have been investigated within the context of L2D-ME \citet{Mao2023Staged,Mao2024Joint} as well. In particular, the authors present novel families of surrogate losses that are underpinned by \textit{realizable consistency bounds} \cite{Awasthi2022}, indicating the existence of upper bounds on the target estimation loss formulated in relation to the surrogate estimation loss, rendering them more advantageous as they are hypothesis set-specific and non-asymptotic.
    }
Finally, \citet{Gao2021} study the problem of L2D-ME in a \textit{bandit feedback} setting (i.e., a sequential dynamic allocation problem). Specifically, 
%    extended their L2D framework based on bandit feedback in order to select the most appropriate human decision-maker for each deferred instance. 
    the deferral policy is specifically learned using a supervised learning model that has been previously trained on historical data that reflect human decisions and corresponding outcomes in order to maximize the complementarity of the machine and human agents. However, by doing so, it is assumed that the human agents who generated the historical data are the same individuals who will be assigned decisions at inference time.
    %The findings from experiments conducted on both simulated and real data indicate that the utilization of such a "personalized" routing algorithm leads to superior performance when compared to the staged and joint L2D algorithms with bandit feedback previously introduced, which only considered a single expert.
    
\textbf{\textit{Multiple-Expert L2D ($j$ out of $J$).}}
\label{sec:l2d:MEnJ}
% \begin{figure}[t!]
%     \centering
%     \includegraphics[width=\textwidth]{media/L2D_ME.png}
%     \caption{Overview of the Multi-Expert Learning to Defer (L2D-ME) setting. Adapted from \cite{Keswani2022}.}
%     \label{fig:l2dMe}
% \end{figure}
In the second scenario of L2D-ME, the multi-expert deferral policy is defined as $\rho^{ME}_M:\mathcal{X}\rightarrow\{0,1\}^{J+1}$. For each input $x\in\mathcal{X}$, the goal is to choose the committee of agents $C(x) \subseteq \{0,\ldots,J\}$, possibly including the classifier, who are likely to make the most accurate decision for $x$. Hence, the $i^{th}$ vector component of the deferral policy will be defined as $\rho^{ME}_M(x)_{(i)}=1$ for all $i\in C(x)$, and $\rho^{ME}_M(x)_{(i)}=0$ for all $i\notin C(x)$. In the event that the designated committee comprises multiple agents, the resulting outcome will be an aggregated decision. 
This setting has been firstly addressed by \citet{Keswani2021}, who proposed a joint loss functions obtained by linearly combining the losses associated to the classifier and deferral function via context-dependent hyperparameters. The authors proved that the combined loss is convex with respect to the classifier and deferral function whenever the loss associated to the former is convex; under such assumption, it can be optimized using the projected-gradient descent algorithm. Additionally, the authors outlined a few adaptations of the L2D-ME framework to account for potential real-world constraints and requirements:
    \begin{itemize}
        \item \textit{Fair learning}: this variant takes into account the possibility of performance discrepancies that may occur with respect to individuals belonging to different protected categories. 
        %These disparities may arise due to the utilization of a biased classifier or the involvement of biased human agents, who may eventually be given greater weight by the system. 
%        the authors propose this variant as a means to address potential disparities in performance among different protected attribute types. These disparities may arise due to the utilization of a biased classifier or the involvement of biased human experts, who may eventually be given greater weight by the system. Two solutions are suggested: (1) the \textit{joint balanced framework}, where group-specific weights are assigned to the samples; and (2) the \textit{joint minimax-fair framework}, where instead the worst error rate across all groups is minimized based on the Minimax Pareto Fairness mechanism \cite{Martinez20}.
        \item \textit{Sparse Committee Selection}: this variant enables the deferral function to exclusively choose a limited number of agents on a per-instance basis. 
        %The initial step involves constructing a probability distribution over the experts based on the weights assigned to them by the deferral function. Subsequently, the maximum number of experts is sampled from this distribution. An alternative approach entails simply choosing the experts who are assigned the highest weights by the deferral function.
        \item \textit{Dropout}: this variant aims to reduce the dependence on a single agent and achieve a more equitable distribution of workload.
%        in order to reduce the dependence on a single expert and achieve a more equitable distribution of workload, the authors suggest the implementation of a dropout mechanism which involves randomly excluding an expert's prediction during the training process with a fixed probability. As a result, the expert's weight is not updated in correspondence of those input samples for which it was excluded.
        \item \textit{Regularized versions}: additional constraints can be added to the joint framework as regularizers of the loss function. For instance, this solution can be employed in cases where specific costs associated with individual human agent consultations are provided.
        \end{itemize}     
    Subsequent work \cite{Keswani2021} further developed this setting to adapt to \textit{closed} deferral pipeline, wherein the human agents of the HS also provided the training labels. 
    %(see Figure \ref{fig:l2dMe})
    This is achieved through an online framework in which
% \citet{Keswani2022} design a closed-deferral pipeline for L2D-ME. The setting is the same as in \cite{Keswani2021}, but here the pipeline is trained in an online manner: 
    input samples are received in a continuous stream. After each prediction is made, which involves aggregating the outputs of agents in the chosen committee, the sample are utilized to retrain the classifier and deferral function. 
    %They propose training procedures under different assumptions, namely: (1) ground truth labels are available, (2) ground truth class labels are unavailable, but there is access to only (noisy) expert labels obtained when the pipeline defers to the human experts.

    Alternatively, \citet{Verma2022Multi} suggest to use Conformal Inference \cite{Shafer2008} to find ensembles of agents $C(x)$ that include the best agent with high marginal probability. The size of $C(x)$ is computed dynamically as a function of the input $x$, thereby ensuring optimal utilization of agent queries. The authors propose two test statistics for the estimation of $C(x)$: a naive score function that sums up the correctness scores of all agents who correctly predict the given instance, and a regularized statistics that employ conformal risk control \cite{Angelopoulos2022} to increase the robustness to noise. The experimental findings demonstrate that the latter approach yields a nearly flawless identification of the appropriate number of agents. Moreover, the conformal approach exhibits superior performance in system accuracy compared to a fixed-size ensemble of agents.  

   % Finally, as noted by \citet{Keswani2022}, deferral pipelines, in which predictions are assigned to either a classifier or one or more human experts, can be considered as \textit{task allocation} problems. Therefore, state-of-the-art techniques in the domain of task allocation within multi-agent systems can be leveraged and adapted to the framework of L2D-ME.  

\subsubsection{Further model architectures}
\label{sec:otherL2D} 
While most of the proposals documented in the literature can be categorized as staged or joint learning models, a few exceptions also exists. A notable example often used as a baseline in the L2D literature is the method proposed by \citet{Okati2021}, namely, an iterative algorithm that optimizes the classifier and triage policy alternately. At each iteration, the optimization process is carried out for the classifier on instances where it outperforms the human agents, while the remaining data points are optimized for the triage policy. The authors show that their method converges to a local minimum. Nevertheless, subsequent experimental studies have shown that this method exhibits lower performance in comparison to other L2D algorithms \cite{Mozannar2023}. Additionally, similar algorithms have been implemented 
%a few studies have been conducted 
to address the issue of L2D for model-specific settings, namely Support Vector Machines \cite{De2020Class} and Ridge Regression \cite{De2020Reg}. \textcolor{black}{A comprehensive framework for L2D in regression tasks with theoretical guarantees was provided by \citet{Mao2024}, covering both staged and joint model architectures. \citet{Yannis2024} recently introduced a staged L2D approach for multi-task settings (classification and regression) that integrates expertise from multiple experts, ensuring Bayes-consistency and realizable consistency for any surrogate with consistent bounds.
A key limitation of L2D is assuming that the human expert available at test time matches the one who provided the training data, which is rarely true in practice \cite{Leitao2022}. Instead, a more realistic approach considers that all potential experts share decision-making similarities. Based on this, \citet{Tailor2024} proposed the \textit{Learning to Defer to a Population} framework, an L2D-SE system capable of deferring to unobserved human predictions during training within a defined population. During testing, it assumes an expert is selected from this population, requiring the L2D system to make deferral decisions despite uncertainty about specific expert behavior, using meta-learning on a limited context set representing expert capabilities.
}

\subsubsection{L2D with limited human predictions} \label{sec:l2dLimitedLabels}
    
A significant drawback of L2D is the requirement of human predictions, alongside ground truth labels, for every instance within the training set \cite{Leitao2022}. Ideally, the L2D system has to be trained on human labels belonging to the same human that will then interact with the system itself. By doing so, the L2D system will learn to complement that specific human \cite{Hemmer2023}. Due to the significant computational and human costs, it is likely that the implementation of the L2D algorithm would be impractical for most real-world scenarios. 
To address implementing L2D-SE algorithms with limited human predictions, \citet{Charusaie2022} proposed \textit{Disagreement on Disagreements} (DoD), an active learning scheme for training a classifier-rejector pair with minimal human queries. DoD operates in two phases: (i) a standard active learning algorithm (e.g., CAL \cite{Cohn1994}) identifies predictor disagreement sets, and humans are queried on these instances to learn their error boundary; (ii) a consistent classifier-rejector pair is then learned from pseudo-labeled data. Alternatively, \citet{Hemmer2023} presented a three-step approach using limited human predictions to generate synthetic labels. First, an \textit{embedding model} maps instances into feature representations. Next, an \textit{expertise predictor model} approximates human capabilities using semi-supervised learning. Finally, the model produces synthetic predictions for unlabeled instances, usable in L2D algorithms. Empirical results show that few human predictions per class suffice for effective synthetic generation.
\textcolor{black}{In the multi-expert L2D context, \citet{Alves2024} have tackled the challenge of limited human data availability by introducing the \textit{Deferral under Cost and Capacity constraints Framework} (DeCCaF). This innovative L2D-ME approach utilises supervised learning to estimate the likelihood of human error with reduced data necessities (e.g., requiring merely one expert prediction per instance) and employs constraint programming to globally minimise error costs while adhering to workload restrictions. In particular, DeCCaF incorporates a component which simultaneously models the behaviour of the human team and forecasts the likelihood that deferring to a certain expert would provide a correct decision.}

\subsubsection{Strength and limitations of Learning to Defer}
In contrast to algorithms that operate under oversight (see Section \ref{sec:hdms1}), Learning to Abstain Hybrid Systems are trained not to predict when their performance is weak. As a result, 
%the prediction generated by the machine is enriched by a new piece of information corresponding to the outcome of the deferral policy. In other words, 
when using an L2R or L2D algorithm to make decisions, one can expect to receive two kinds of evidence: the machine's prediction concerning the action of deferral, and, if the AI does not abstain, the result of the prediction task. 
L2D algorithms improve upon L2R by incorporating a representation of human knowledge directly in the training process.
%The utilization of L2D algorithms offers an improvement over L2R by incorporating a certain level of "adaptability" into the deferral policy. This is achieved by attempting to model the human decision-making behavior and using this representation as a discriminating factor for training the deferral policy. Specifically, a collection of predictions that represent decisions made by human decision-makers (e.g., historical or prototypical decisions) is matched against both the AI predictions and the ground truths in order to learn their error boundary. 
%of human decision-makers
%and matched against both the AI predictions and the ground truths corresponding to the same input data. 
In such a way, the deferral policy is trained to adapt to both the AI model and the human decision-maker, ideally the same that will employ the Hybrid System. 
%The ideal scenario would involve the adoption of \textit{closed} deferral pipelines, wherein the human predictions utilized during training are sourced from the same decision-makers who will employ the system at inference.
Recent empirical investigations involving human subjects yielded evidence for the additional advantages that abstaining systems bring to Hybrid Systems. \citet{Hemmer2023UserStudy} found that such algorithms improve both \textit{human task performance} compared to a human or an AI working alone, 
%(and, consequently, human-AI team performance), 
and \textit{human task satisfaction} compared to a human working alone. 
%Interestingly, this improvement is observed regardless of whether or not humans are aware that the instances they receive have been deferred by a machine. Furthermore, the authors found that the effects of deferral on task performance and task satisfaction can be attributed to an enhancement in humans confidence regarding their capability to successfully accomplish the task. 
A different study \cite{Papenmeier2023} also investigated the effects of employing abstaining Hybrid Systems on the human
%human performance, as well as their 
perception of AI performance and credibility. The results indicate that users are frequently influenced by the system's recommendation also on ambiguous instances, even without conscious awareness, and thus support the adoption of L2D algorithms. 
%Additionally, the authors noted that the use of abstaining algorithms does not have a detrimental effect on the perceived performance and trustworthiness of the system.
%While many studies have shown that abstaining algorithms may exhibit superior performance compared to both humans and AI working in isolation,
However, L2D also comes with several limitations ~\cite{Leitao2022}. Most importantly, these include: data availability issues, which are primarily due to the need of human predictions in addition to the ground truth for all instances within the training set and all human agents involved in the Hybrid System; %concerns about the applicability of methods, particularly in non-stationary settings where the outcomes can be influenced by the predictions or dataset shifts can occur; 
and fairness concerns that may stem from the introduction of bias by both human and machine agents, as well as from the abstention mechanism itself \cite{Jones21}. Although there have been suggestions to deal with such issues,
%limited human predictions or to incorporate fairness regularization in the L2D framework, 
these proposals still do not offer a straightforward solutions. %for addressing this issue.

In terms of the human's role in the Hybrid System, Learning to Abstain marginally enhances the paradigm of human oversight over machines (Section \ref{sec:hdms1}), as the deferral is exclusively a machine-side operation and there is no direct human-side interaction considered in the design of the algorithms.

  \section{Learning Together: Humans Teaching Machines, and Machines Teaching Humans}
\label{sec:hdms3}
  The next natural step in hybrid
  % [REPLACE] decision-making
  systems
  is a two-way collaboration in which human agents are not mere executors or overseers, but can directly \emph{interact} with the machine to best infuse their human decision-making abilities directly into the machine.
  While Learning to Defer aims to identify which agent is best suited for a given
  % [REPLACE] decision,
  prediction,
  there is little to no effort in integrating the decision-making abilities of one agent to the other.
  In such a setting, the
  % [REPLACE] decision
  prediction
  capability of the overall system is at best but a sum of the
  % [REPLACE] decision
  prediction
  abilities of its agents, that is, the system is not synergic.  
  Usually, in machine learning machines learn directly \textit{from data}, while in a Learning Together system they aim to learn \textit{from other agents}.
  To reach the ultimate goal of learning from other human agents, we need to build on top of a foundational stepping stone that enables machines to learn from other machines, the \emph{Teacher-Student} paradigm.
    % Bidirectional learning between human and machines improve on existing 

  \subsection{Machines Learning from Other Machines: The Teacher-Student Model}
    % Seamless Human-Machine collaboration has been a long-standing goal of Artificial Intelligence systems, dating back to the 80s~\cite{DBLP:journals/nn/VapnikV09}.
    % We can draw a direct parallel to standard
    At the basis of machine-to-machine learning systems is the \emph{Teacher-Student} paradigm.
    In this approach, we identify two agents: a \emph{Teacher}, whose goal is to train, and a \emph{Student}, whose goal is to learn from the Teacher.
    Among Teacher-Student models, four are of particular note:
    Learning with Privilege~\cite{DBLP:journals/nn/VapnikV09},
    % Knowledge Distillation~\cite{DBLP:conf/kdd/BucilaCN06,DBLP:journals/corr/HintonVD15,DBLP:journals/corr/RomeroBKCGB14,DBLP:conf/cvpr/YimJBK17},
    % [CIT] Knowledge Distillation~\cite{DBLP:conf/kdd/BucilaCN06,DBLP:journals/corr/HintonVD15,DBLP:journals/corr/RomeroBKCGB14},
    % Knowledge Distillation~\cite{DBLP:conf/kdd/BucilaCN06,DBLP:journals/corr/HintonVD15},
    Knowledge Distillation~\cite{DBLP:conf/kdd/BucilaCN06},
    Transfer Learning~\cite{DBLP:journals/pieee/ZhuangQDXZZXH21},
    and Active Learning~\cite{settles2009active}.
    % \textcolor{black}{aimed at Machine-Machine collaboration}
    % include Learning with Privilege~\cite{DBLP:journals/nn/VapnikV09}, Knowledge Distillation~\cite{DBLP:conf/kdd/BucilaCN06,DBLP:journals/corr/HintonVD15,DBLP:journals/corr/RomeroBKCGB14,DBLP:conf/cvpr/YimJBK17}, and Transfer Learning~\cite{DBLP:journals/pieee/ZhuangQDXZZXH21},
    % most of which
    % \textcolor{black}{which, however, tackle}
    % all focused on a
    % machine-to-machine settings where all agents are machines.
    
    \textit{\textbf{Learning with Privilege.}}
      Initially introduced by~\citeauthor{DBLP:journals/nn/VapnikV09}~\cite{DBLP:journals/nn/VapnikV09}, Learning with Privilege is defined as a learning paradigm in which a teaching agent $f_T$, i.e., a Teacher, provides the learning agent $f_S$, i.e., the Student, with additional ``privileged'' information about the input data which is not present in the data itself.
      % Formally,\mattialightnote{ask matematiclara if she wants to mathematically formalize this too}
      The additional data is considered privileged because only available at training time.
      In our formulation, this is roughly equivalent to integrating a proxy $\widetilde{Z}$ of the human knowledge $Z$ into the training data.
      At learning time, a privileged estimator implements a function
      \begin{equation*}
        \mathcal{F}_M: \mathcal{X} \times \mathcal{Z} \rightarrow \mathcal{Y_M},
      \end{equation*}
      which directly integrates the human knowledge into the model.
      Integration relies on a machine-specific encoding function that allows to map the input data $X \in \mathcal{X}$ and the human knowledge $Z \in \mathcal{Z}$ into the input space $\mathcal{X}$.
      At inference time, the privileged information is unavailable, and the encoding function reduces to its projection $(x,\{\emptyset\})\mapsto x$,
      %transitions into an identity map $\mathcal{X} \times \{\emptyset\} \rightarrow \mathcal{X}$, 
      thus allowing the machine to seamlessly operate either with or without privileged knowledge.

    % On the other side,
    \textbf{\textit{Knowledge distillation.}}
      Knowledge distillation
      % is a form of Privileged learning in which the goal is
      aims to distill a given Teacher into a more suitable Student, usually to reduce the running cost of inference~\cite{DBLP:conf/kdd/BucilaCN06}.
      Unlike Privileged learning, here the privileged information is exclusively derived from a machine, usually involving its internal state, which the Student tries to emulate.
      Formally, at training time the Student is directly conditioned on the parameters of the Teacher, thus yielding an estimator $f_\theta$ of the family:
      \begin{equation*}
        \mathcal{F}_{M,S}: \mathcal{X} \times \mathcal{F}_{M,T} \rightarrow \mathcal{Y_M},
        % f_{\theta \mid \theta'}(X).
      \end{equation*}
      where we assume without loss of generality that the Teacher and Student are members of the same hypothesis space $\mathcal{F}_M$.
      The Student loss $\widetilde{\mathcal{L}_S}$ is thus augmented with an additional component $\widetilde{\mathcal{L}}_{T,S}(f_T, f_S)$ penalizing a distance between Teacher and Student:
      \begin{equation*}
        \widetilde{\mathcal{L}_S}(X, Y) = \frac{1}{n}\sum_{i=1}^n\mathit{\widetilde{\mathscr{L}}(Y_i,f_S(X_i))} + \lambda_S \widetilde{\mathcal{L}}_{T, S}(f_T, f_S),
      \end{equation*}
      for a predefined weight $\lambda_S$ balancing task performance and distance from the Teacher.
      In neural models, where Knowledge Distillation is prevalent, the Teacher-Student loss $\widetilde{\mathcal{L}}_{T, S}$ is usually implemented as $d(\theta_T, \theta_S)$, e.g., Euclidean or Cosine distance, between the parametrization $\theta_T$ of the Teacher and the parametrization $\theta_S$ the Student.
      In both Learning with Privilege and Knowledge Distillation settings, the final goal is to transfer information between Teacher and Student.
      
    % Finally, in
    \textit{\textbf{Transfer Learning}}
      In this the stpe where the Students tries to leverage the knowledge of the Teacher, to then adapt it to the specific task at hand~\cite{DBLP:journals/pieee/ZhuangQDXZZXH21}.
      % [CIT] Transfer learning the Teacher is a stepping stone for the Student, which first tries to leverage the knowledge of the Teacher, and only then adapts it to the task at hand~\cite{DBLP:journals/jbd/WeissK016}.
      This paradigm is typical of dynamic settings in which tasks or data distributions change rapidly, yet they are all strongly related.
      In this context, learning from agents which have already learned to solve related task is of great benefit.
      
    \textit{\textbf{Active Learning.}}
      The aforementioned approaches all provide a straightforward and direct way to inject learning capabilities from the Teacher to the Student. 
      However, the latter remains a passive agent in this process, resulting in a unidirectional connection between the two.
      %yet the latter is still \textcolor{red}{a passive agent}, thus the connection between the two is one-directional.
      Furthermore, knowledge is injected one-shot at training time, that is, these paradigms are not designed for subsequent Teacher-Student interactions.
      Active learning~\cite{settles2009active} builds on this paradigm by empowering the Student to actively seek the most relevant information.
      Like a student that poses questions to a teacher, an Active Student \textit{queries} the Teacher (in our formulation, the human agent) to maximize its own performance.
      Formally, the Student has at its disposal a pool of unlabeled data $X_U \in \mathbb{P}(\mathcal{X})$ that can, in principle, be labeled with the predictions of the Teacher.
      Querying the Teacher is feasible but costly, thus one wishes to maximize Student performance while staying under a given budget.
      To tackle this problem, we aim to learn a \emph{query policy}
      %
     % \begin{equation*}
       $ \pi: \mathcal{F}_M \times \mathrm{P}(\mathcal{X}) \mapsto \mathrm{P}(\mathcal{X})$
      %\end{equation*}
      % \begin{equation*}
      % \;\;\;\;(f_S, \;\;\; X_U) \mapsto \hat{X}_U
      % \end{equation*}
      %
      that, given the Student $f_S$ and the unlabeled data $X_U$, selects a subset $\hat{X}_U \subset X_U$ for which the Teacher then provides labels $\hat{Y}_U = \langle f_T(x_U) \rangle_{x_U \in \hat{X}_U}$.
      Training the Student involves an iterative process of querying and training, and in each iteration the Student seeks to best select the subset of instances $\hat{X}_U \subset X_U$ that will maximize its performance.
      $X_U$ can vary in its nature: it can be static or dynamic~\cite{settles2009active}, it can be updated with a handful~\cite{DBLP:conf/nldb/MullerPBF22} or a large set of novel unlabeled samples~\cite{DBLP:journals/ml/CohnAL94}, and the Student may even populate $X_U$ itself.
      Initial formulations of Active learning~\cite{DBLP:journals/ml/CohnAL94} strongly resembles Learn to Reject solutions,as the goal of both is to identify regions in the input space $\mathcal{X}$ where the machine is less accurate.
    The Teacher-Student paradigm is limited to one-way learning between agents, which makes it extremely flexible, as it does not have to model the interaction between agents, nor the agents themselves.
    This introduces several layers of complexity, each compounding on the previous, making for novel approaches to Learning Together, which takes inspiration from the aforementioned paradigms.

  \subsection{Learning Together}
    \begin{table}
      \centering
      \footnotesize
      \begin{tabular}{ @{} V{3.2cm} V{6cm} l @{} }
        \toprule
        % & \textbf{Symbol} & \textbf{Values} & \textbf{Description} \\
        % \midrule
        \textbf{Communication language.} & \multicolumn{2}{c}{\textit{The communication language that allows human and machine to interact.}} \\
        Hard reasoning      & First-Order logic. & \cite{DBLP:conf/iui/GuoDAMCK22,DBLP:conf/pldi/EllisWNSMHCST21} \\
        Soft reasoning      & Logic-like languages with soft reasoning engines. & \cite{DBLP:conf/emnlp/MishraTC22,DBLP:conf/emnlp/BostromZCD21,DBLP:conf/emnlp/MurtyMLR22,wang2017fvqa} \\
        Explanations        & Feature relevance, decision rules, and concepts. & \cite{DBLP:journals/aim/AmershiCKK14,MalandriMMN23,teso2019toward,DBLP:conf/aies/TesoK19,DBLP:conf/ijcai/RossHD17,DBLP:conf/kdd/WangKNSNCVV022,Popordanoska2020,DBLP:journals/corr/abs-2012-02898,DBLP:journals/datamine/BontempelliGPT22} \\
        \midrule
        
        \textbf{Interaction artifact.} & \multicolumn{2}{c}{\textit{The type of interaction artifact.}} \\
        Intrinsic           & Artifact internal to the machine, the human agent cannot directly influence the machine. & \cite{teso2019toward,DBLP:conf/aies/TesoK19,DBLP:conf/ijcai/RossHD17,DBLP:conf/kdd/WangKNSNCVV022,Popordanoska2020,DBLP:journals/corr/abs-2012-02898,DBLP:journals/datamine/BontempelliGPT22,DBLP:journals/aim/AmershiCKK14,MalandriMMN23}\\
        Extrinsic           & Artifact external to the machine, the human agent can directly influence the machine. & \cite{DBLP:conf/iui/GuoDAMCK22,DBLP:conf/pldi/EllisWNSMHCST21,DBLP:conf/emnlp/MishraTC22,DBLP:conf/emnlp/MurtyMLR22} \\
        \midrule
        
        \textbf{Time of interaction.} & \multicolumn{2}{c}{\textit{The machine phase in which the interaction occurs.}} \\
        Training            &  The human agent influences the machine directly in its training process. & \cite{teso2019toward,DBLP:conf/aies/TesoK19,DBLP:conf/ijcai/RossHD17,DBLP:conf/kdd/WangKNSNCVV022,Popordanoska2020,DBLP:journals/corr/abs-2012-02898,DBLP:journals/datamine/BontempelliGPT22,DBLP:journals/aim/AmershiCKK14,MalandriMMN23} \\
        Inference           &  The human agent influences the machine at inference time. & \cite{DBLP:conf/iui/GuoDAMCK22,DBLP:conf/pldi/EllisWNSMHCST21,DBLP:conf/emnlp/MishraTC22,DBLP:conf/emnlp/BostromZCD21,DBLP:conf/emnlp/MurtyMLR22,wang2017fvqa} \\
        \midrule

        \textbf{Learning cost.} & \multicolumn{2}{c}{\textit{The cost incurred by the machine to integrate the feedback by the human agent.}} \\
        Training            & The correction prompts an additional training phase of the machine. & \cite{DBLP:journals/aim/AmershiCKK14,teso2019toward,DBLP:conf/aies/TesoK19,DBLP:conf/ijcai/RossHD17,DBLP:conf/kdd/WangKNSNCVV022,Popordanoska2020,DBLP:journals/corr/abs-2012-02898,DBLP:journals/datamine/BontempelliGPT22} \\
        Null                & The correction does not cause any significant cost to the machine. & \cite{DBLP:conf/iui/GuoDAMCK22,DBLP:conf/pldi/EllisWNSMHCST21,DBLP:conf/emnlp/MishraTC22,DBLP:conf/emnlp/BostromZCD21,DBLP:conf/emnlp/MurtyMLR22,wang2017fvqa,MalandriMMN23} \\
        \bottomrule
      \end{tabular}
      \caption{Properties of systems in the Learning Together paradigm.}
      \label{tbl:hdms4:axes}
    \end{table}
    Unlike the aforementioned Teacher-Student paradigm, the \emph{Learning Together} paradigm aims to
    % properly\mattialightnote{meh} transfer\mattialightnote{meh again} decision-making capabilities among different agents.\mattialightnote{questa va pensata per bene}
    create a two-way street in which human and machines can communicate effectively, one learning from the other: the machine explaining its prediction to the human, and the human explaining its prediction to the machine, improving the overall performance of the system.
    % both improving their ability to make decisions.
    In this context, improvement can take many forms.
    For the machine, it can
    % be more
    provide more
    accurate predictions, a better ability to generalize, a shallower learning curve, or higher level of transparency. 
    On the other hand, human agents can derive additional advantages from exerting
    %For the human, we also have the added benefits of having a
    a tighter grip on \emph{how} the machine solves the task, rather than simply solving the task
    % for
    in place of the machine.
    % and discovering novel solutions to the task.
    %
    % TODO: done
    \begin{tcolorbox}[colback=blue!5!white,colframe=blue!50!white,title=Example of melanoma detection: Learning Together,fonttitle=\small]
    \small
        In Learning Together systems, the dermatologist may interact with an explanation of the melanoma classification.
        This would allow them to adjust the reasoning of the model based on their own judgement through, e.g., correcting the reasoning on the size, shape, and color of the potential melanoma.
        After receiving corrections from humans, the machine then integrates them to enhance the learning process and better perform on future images.
    \end{tcolorbox}
    Generally, a machine implementing this paradigm partially integrates a human agent with expertise $Z$, thus yielding an estimator of the form:
    $f_{\theta}(X, Z)$.
    For the sake of simplicity, here we model a single human agent, but the formulation can be extended to multiple agents.
    We characterize systems in the Learning Together paradigm according to four properties that we summarize in Table~\ref{tbl:hdms4:axes}: communication language, which guides the communication between agents; interaction artifacts, which guide the feedback mechanism; time of interaction, which defines \textit{when} the interaction occurs; and learning cost, which defines what \textit{cost} the interaction creates.
  
    \subsubsection{Communication Languages and Interaction Artifacts}
    \label{subsec:commandart}
      The communication language plays a critical role in facilitating successful high-level human-machine interactions.
      On one hand, its level must be sufficiently low to allow the machine to understand and interface with it; on the other hand, its level should be high enough for the human agent to understand it.
      Communication languages are comprised of two intertwined components: a \textit{language}, which defines how the agents communicate, and a set of \emph{interaction artifacts}, which comprise the atomic unit of interaction, and enable the feedback of one agent to be integrated into the other.

      \textit{\textbf{Interaction Artifacts.}} At the core of a communication language lies a set $A$ of \emph{interaction artifacts} that enable effective communication between humans and machines, and enable the embedding of human knowledge $Z$ into the machine.
      % In other words,
      That is, the human encodes
      % its
      their knowledge by \emph{acting} on the interaction artifacts $A \in \mathcal{A}$ presented by the machine.
      Then, the machine acknowledges the human action on $A$ and responds consequently.
      % the human feedback and readjusts its operations accordingly.
      %and elaborates the human by elaborating the action of the human agent on $A$.
      In the reviewed papers (see Table \ref{tbl:hdms4:axes}), typical artifacts include logic rules, relevant features, and decision rules.
      The human action can also take several forms, and is inherently dependent on the type of the interaction artifact.
      
      Regardless of their nature and formulation, artifacts are:
      \begin{itemize}
        \item \textit{understandable} by both the human and the machine, since they must enable communication;
        
        \item \textit{malleable} by the human, as they transfer expertise $Z$ to the machine  by acting on the artifact;
        
        \item \textit{embeddable} into the machine, to integrate human action on the artifact into the machine.
        \end{itemize}

      The last property puts a strong constraint on machines, which, unlike artifacts, tend to employ subsymbolic, e.g., neural, rather than symbolic models.
      This often results in systems where the family and architecture of the machine are all but determined by the choice of artifacts, thus communication language and machine end up being tightly coupled.
      % \foscanote{ ma qua a cosa fai riferimento? questa formulazione di human action va bene per tutti I paradigm, forse questa parte potrebbe andare nei concetti di base}
      
      Interaction artifacts come in one of two forms: \emph{intrinsic} and \emph{extrinsic}, yielding intrinsic and extrinsic machines, respectively.
      Intrinsic artifacts are machine-oriented, thus after the human acts upon them, the integration step is hidden to the human agent, who has no direct control of how the machine integrates the action.
      This often allows easier definition of the interaction artifact set at the cost of the effectiveness of the action.
      The result of an interaction between an intrinsic machine parametrized by $\theta$ and a human agent acting on an artifact $a$ is in an update of the model parameters $\theta$, which yields an estimator of the form
      % %
      % \begin{equation*}
      $f_{\theta \mid a}(X)$,
      % \end{equation*}
      % %
      where $\theta$ is conditioned on the artifact $a$ acting as proxy feedback for the human agent.
      Intrinsic artifacts are most often employed for systems using \textit{explanations} as communication language.
      % These artifacts are typically employed in systems explanations,
      %i.e., systems where humans and machines communicate through, e.g.,  decision rules or feature relevance vectors.

      Extrinsic artifacts instead, are human-oriented, thus once the human acts upon them, the integration is straightforward for the machine, which does not require additional integration steps.
      Typically, such artifacts are organized in an artifact bank $B$ that the user can inspect and act upon.
      %by introducing novel artifacts, removing or modifying existing ones.
      With an artifact bank to attend to, the interactive machine $M$ implements an estimator of the form
      % %
      % \begin{equation*}
      $f_\theta(X, B)$.
      % \end{equation*}
      % %
      Note that the whole artifact bank is a parameter of the machine, hence the parameters $\theta$ of the machine are independent of the artifact bank, 
      %thus allowing one to easily steer the machine by exclusively acting on the artifacts.
      Extrinsic artifacts are typically more complex than intrinsic ones, and so are the communication languages built upon them.

      \textit{\textbf{Communication languages.}}
      Communication languages can be broadly categorized in two families: \emph{reasoning} languages, both \textit{hard} and \textit{soft}, and \emph{explanation} languages.
      
      \textit{\textbf{Hard reasoning languages:} Logic.}
        Turning to languages, logic natively offers languages and artifacts that satisfy all three requirements. In literature we observe two approaches: those based on logic and those that combines logic with a neural component.
        % \mattialightnote{sarebbe da mettere xai prima ma c'e' solo FI praticamente} are a natural candidate to satisfy all three properties.
        Logic programs are naturally readable by humans due to their transparency and similarity to human reasoning, and thus are a suitable candidate for communication language.
        They are also naturally malleable but it is not straightforward to embed them in the machine due to their symbolic nature,
        which is in stark contrast with the neural machines. Therefore, another interesting family of Learning Together systems using logic as a communication language combines subsymbolic and symbolic components \cite{DBLP:conf/iui/GuoDAMCK22,DBLP:conf/pldi/EllisWNSMHCST21}.
        The subsymbolic component is typically a neural one, e.g., a neural network, which fully embodies machine reasoning.
        The symbolic component is geared towards the human, and thus employs a logic language.
        Neuro-symbolic systems aim to combine and integrate these two components to create an integrated framework that leverages the reasoning capabilities of symbolic AI and its adaptability of sub-symbolic AI \cite{Renkhoff2024}.
        
        Generally,\textbf{ \textit{logic and reasoning-like languages}} define a set of clauses or rules $\mathcal{R}$, and facts $\mathcal{T}$ that allow a
        logic engine to reason and present its logical derivations to the human agent as extrinsic artifacts on which they can act.
        % logical reasoning engine to perform different forms of reasoning which can be presented to the user as extrinsic artifacts on which the user can act.
        Possible actions on logic artifacts, i.e., logic rules, facts, and compositions of the two, include \textit{addition} and \textit{deletion} of new/existing rules or facts.
        Hard logic enjoys strong theoretical properties, which makes it suitable for highly compliant systems where providing feedback to the machine is highly likely to result in a successful integration in the machine.
        % On top of this, hard logic has a long history of successful applications, which have made it  
        Logic as a communication language is expert-driven, rather than data-driven.
        This means that artifacts in these languages are often hand-crafted, domain-specific, and automating their definition
        % or transferring them to related tasks or datasets
        usually involves a significant effort on the human side.
        \begin{figure}[t!]
          \includegraphics[width=0.8\textwidth]{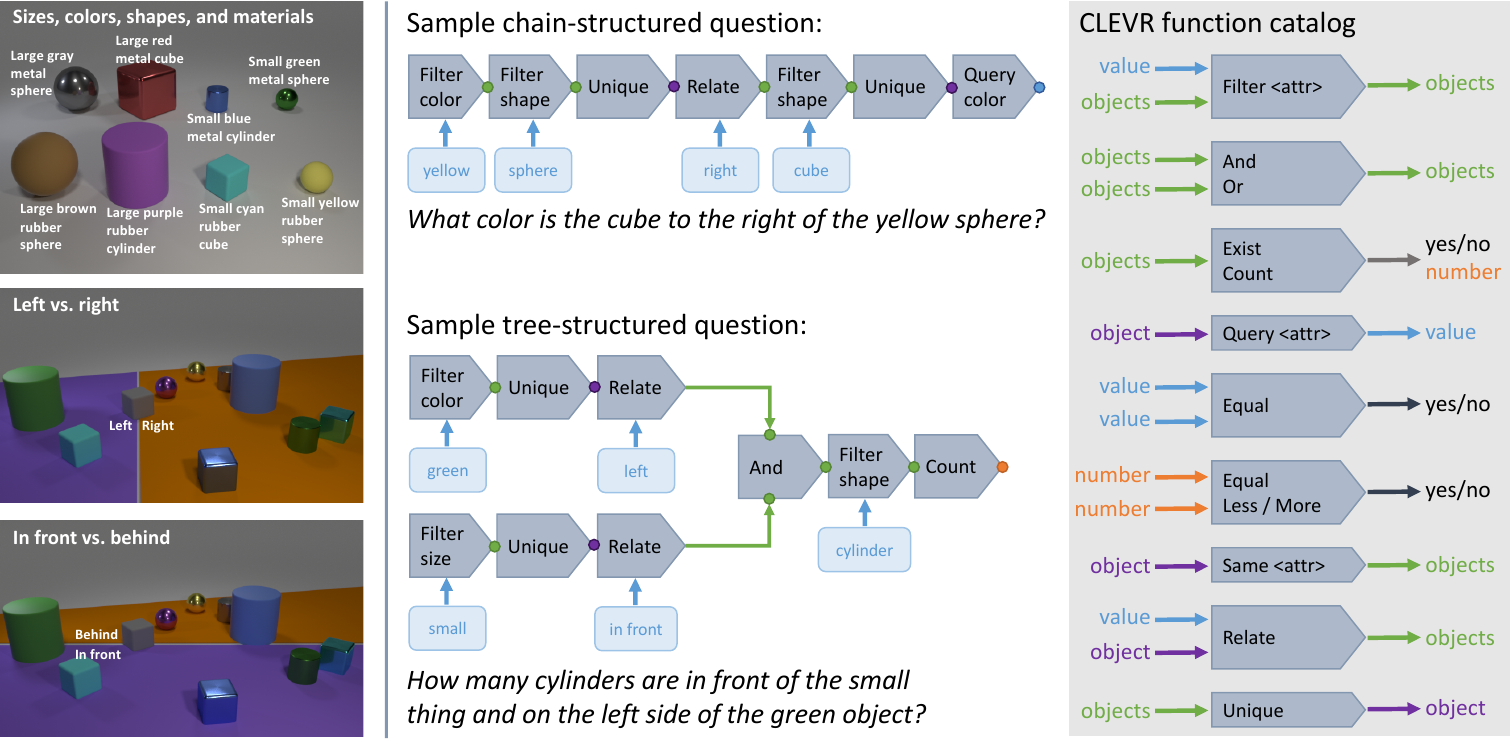}
          \caption{An example of Question Answering machine with a hard reasoning language. The agent maps the question to a program using a set of primitives%(right-hand side of the picture)
          , and then executes the program on the input
          %(left-hand side of the picture)
          , providing the human with both a prediction, and a malleable program that they can correct.}
          \label{fig:hdms4:clever}
        \end{figure}
        Figure~\ref{fig:hdms4:clever} provides an example extracted from the CLEVR dataset~\cite{DBLP:conf/cvpr/JohnsonHMFZG17}.
        Here, the task is to answer questions about the provided image, e.g., ``What is the color of the cube to the right of the yellow sphere?''
        The interaction artifact consists of a program solving the task by first identifying the yellow sphere, then finding objects to its right and finally filtering them by shape.
        The intermediate result of this program is then used to extract the color of such object.
        When presented with it, the human agent can inspect and appropriately modify the program to solve the task.
        We can find another classical example of hard reasoning in~\cite{DBLP:conf/iui/GuoDAMCK22}, where the machine is tasked to predict the final outcome of a Tic-Tac-Toe board by leveraging a set of logic rules, which the human can inspect and change at their leisure.

        \textit{\textbf{Neurosymbolic}} models also offer another related subfield, \emph{neural program synthesis}.
        Like inductive logic programming, neural program synthesis tries to solve tasks by learning interpretable programs, only in this case the program is not expressed in a logic language, but in an actual programming language.DreamCoder~\cite{DBLP:conf/pldi/EllisWNSMHCST21} is a neurosymbolic model to solve tasks by writing computer programs in a given minimal computer language.
        Notably, DreamCoder progressively grows a library of functions, small reusable snippets of computer code re-used throughout the program.
        A human agent can then inspect and change functions in the library that they deem incorrect.

        % \textcolor{red}{
        % Still, logic is not the only language of choice.
        % Recent advances in program synthesis~\cite{} have shown promising results in a number of tasks~\cite{}, thus opening the door to a host of applications, including human interaction with the generated programs.
        % The complexity of the programs is directly induced by the language defining them, and, as it is the case for logic languages, on the capability of the program induction algorithm to find concise and simple programs among the ones solving the task.
        % Languages 
        % allowing one to leverage program languages as a 
        % in recent years they have been 
        % Hard reasoning
        % }
      
      \textit{\textbf{Soft reasoning languages.}}
        Soft reasoning communication languages improve the flexibility of Logic languages at the cost of their strong theoretical properties by replacing logic rules and facts with \textit{logic-like} rules, and the symbolic reasoning engine with a subsymbolic approximate one.
        Like in hard reasoning languages, possible actions on artifacts include \textit{addition} and \textit{deletion} of new/existing rules or facts.
        These languages are almost exclusively Natural Languages, and are themselves applied to Natural Language models.
        % There are several practical reasons for this:
        % \textit{i)} Natural Language models can interact almost seamlessly with Natural Language, thus requiring little to no encoding effort from the human agent;
        % % \textit{ii)} even if sometimes domain-specific, several Natural Language-related tasks require similar foundational skills\mattialightnote{da cambiare?}, thus general foundational models, which are almost exclusive to Natural Language~\cite{zhou2023comprehensive};
        % % \textit{ii)} real-world datasets are often noisy or small in size, and Natural Language foundational models are uniquely able to handle noise, and be effective on small datasets;
        % \textit{ii)} Natural Language models already show some, even though limited, promise in reasoning tasks~\cite{dasgupta2022language};
        % \textit{iii)} Natural Language models show improvement when presented with reasoning-like feedback~\cite{wei2022chain}, and show promise in following explicit human instructions~\cite{DBLP:conf/nips/Ouyang0JAWMZASR22} and integrating them into the machine~\cite{talmor2020leap,pi2022reasoning}.   
        % Originating from the work by~\citeauthor{DBLP:conf/emnlp/KassnerTSC21}, this family of machines builds on top of two other families, knowledge injection models~\cite{DBLP:conf/nips/LewisPPPKGKLYR020} and soft reasoners~\cite{DBLP:conf/emnlp/Abzianidze17,DBLP:conf/ijcai/ClarkTR20}, the former providing models able to integrate external possibly human-sourced knowledge in their inference, and the latter providing soft reasoning engines for Natural Language.
        Originating from the work by~\citeauthor{DBLP:conf/emnlp/KassnerTSC21}, this family of machines builds on top of two other families, knowledge injection models~\cite{DBLP:conf/nips/LewisPPPKGKLYR020} and soft reasoners~\cite{DBLP:conf/ijcai/ClarkTR20}, the former providing models able to integrate external possibly human-sourced knowledge in their inference, and the latter providing soft reasoning engines for Natural Language.
        %TODO: uncomment to add fix,  add citation here
        \textcolor{black}{Knowledge injection models find a successful application in Large Language Models (LLMs), which can inject verbalized artifacts through prompting~\cite{DBLP:journals/corr/abs-2402-07927}.}
        
        \textcolor{black}{\textit{\textbf{On Large Language Models.}}
        Large Language Models (LLMs) are rapidly evolving and capable of solving various natural language tasks. In Learning Together systems, their interactive nature warrants consideration. Research on LLMs \cite{llms_agents} often focuses on self-sufficient models in specific environments, sometimes with human input to enhance performance. LLM prompting, i.e., creating tailored inputs to guide outputs, is a Natural Language-based interaction where humans provide context or set boundaries for AI, but this does not change the model's structure, only its contextual output. Unlike Learning Together systems, LLM interactions are purely linguistic and stateless, limited by context size \cite{llms_agents}. In contrast, Learning Together systems involve active human engagement to modify models through interaction.} 

        \textit{\textbf{Question answering.}}
        Due to the strong emphasis on flexibility,
        % \foscanote{ cosa significa soft qua? come soft constraints...o altro?}
        soft reasoning languages are usually loosely defined in terms of a set $\mathcal{T}$ of facts and informal rules $\mathcal{R}$.
        % escape formal definitions
        On inference, the machine presents the human agent with a subset $T \subset \mathcal{T}$ of facts, optionally accompanied by a set of derivation rules $R \subset \mathcal{R}$ it leveraged.
        According to the complexity of the task, facts may be enriched with intermediate reasoning steps that the machine has derived or generated to solve the task.
        \begin{figure}[t!]
            \centering
            \resizebox{0.75\textwidth}{!}{%
            \begin{tikzpicture}[node distance=1cm, auto,->,shorten >=2pt,
            database/.style={
              shape=rectangle,
              fill=mygreen!25,
              aspect=0.25,
              align=center,
              draw
            }
            ]
            % nodes
            \node[shape=rectangle,draw=black,fill=myblue,fill opacity=0.25,text opacity=1] (h1) at (-3,0) {Can a magnet pick up a penny?};            
            \node[shape=rectangle,draw=black,align=center,fill=myred,fill opacity=0.25,text opacity=1] (m1) at (0,-1) {Yes: i) pennies are made of metal,\\ii) metals are magnetic.};
            \node[shape=rectangle,align=center,draw=black,fill=myblue,fill opacity=0.25,text opacity=1] (h2) at (-3.0,-2.2) {Not all metals are magnetic and\\copper is not magnetic.};
            
            \node[shape=rectangle,draw=black,fill=myblue,fill opacity=0.25,text opacity=1] (h3) at (-3,-3.2) {Can a magnet pick up a penny?};            
            \node[shape=rectangle,draw=black,fill=myred,align=center,fill opacity=0.25,text opacity=1] (m2) at (0.4,-4.2) {No! Pennies are made of copper,\\ which is not magnetic.};

            \node[database] (db1) at (6,-0.6) {Pennies are made\\of metal\\Metals are magnetic};
            \node[database] (db2) at (6,-2.1) {Updating...};
            \node[database] (db3) at (6,-4.25) {...\\Copper isn't magnetic};

            % \path [->] (db1) edge node {} (m1);
            \draw [->] (db1.west) |- (m1.east);
            \draw [->] (h2.east) |- (db2.west);
            % \path [->] (h2) edge node {} (db2);
            \draw [->] (db2.south) -- (db3.north);
            \draw [->] (db3.west) |- (m2.east);
    
        \end{tikzpicture}
          }
          \caption{An example of a Question Answering hybrid system with a soft reasoning language. The \textcolor{myblue}{human} can interact with the \textcolor{myred}{machine} through Natural Language.
          First, the machine provides a prediction and a rationale to the user, who then is able to correct the rationale and feed it back to the machine, updating its \textcolor{mygreen}{artifact bank} (``Feedback memory'' in~\cite{DBLP:conf/emnlp/MishraTC22}).
          On a subsequent interaction, the machine provides the correct prediction and rationale to the same question by leveraging the previous feedback.}
          \label{fig:hdms4:feedback}
        \end{figure}
        Figure~\ref{fig:hdms4:feedback} shows an example of a Question Answering task where the hybrid system is tasked to answer basic physics questions by the user~\cite{DBLP:conf/emnlp/MishraTC22}.
        The machine is asked whether ``A magnet can pick up a penny'', and generated both an answer, ``Yes'', and an artifact supporting it in the form of syllogistic reasoning.
        The human agent notices a flaw in the premises, corrects the artifact by stating that there are non-magnetic materials, one of which is copper.
        Here, the integration step is minimal, and consists in adding the correction by the human agent to the artifacts bank.
        Then, on a successive iteration, the machine is presented with the same question, and successfully answers by leveraging the claims introduced by the user, without incurring any additional training cost.        
        % Other works operate on similar terms by providing full reasoning trees~\cite{DBLP:conf/emnlp/MishraTC22}, improving on the artifacts by artificial generation~\cite{DBLP:conf/emnlp/DalviJTXSPC21,DBLP:conf/emnlp/BostromZCD21,DBLP:conf/emnlp/Yang0C22}, or by providing ad-hoc artifacts to correct the machine~\cite{DBLP:conf/emnlp/MurtyMLR22}.
        Other works operate on similar terms by providing full reasoning trees~\cite{DBLP:conf/emnlp/MishraTC22},
        % improving on the artifacts by artificial generation~\cite{DBLP:conf/emnlp/BostromZCD21,DBLP:conf/emnlp/Yang0C22},
        improving on the artifacts by artificial generation~\cite{DBLP:conf/emnlp/BostromZCD21},
        or by providing ad-hoc artifacts to correct the machine~\cite{DBLP:conf/emnlp/MurtyMLR22}.
        The human agent can act upon them by correcting any number of facts and/or rules used in inference, yielding updated facts $\mathcal{T}'$ and $\mathcal{R}'$.    
        % \foscanote{ dubbio basico: ma se consideriamo hybrid systems anche I QA, perchè non lo sono anche I query system di un database?...non è che stiamo andando fuori tema?..e perchè c'è interazione? dove è il loop? intendiamoci, questa digressione sul linguaggio di comunicazione mi piace, ma..}
        
        \textit{\textbf{Knowledge Graphs.}}
        Another clear advantage of soft reasoning languages is the ability to integrate artifacts from multiple sources to aid reasoning.
        Knowledge Graphs are a primary source for several reasons:
        % \paragraph{Knowledge Graphs and Knowledge Bases.}\mattialightnote{weaken this to some evidence-only?}
        % Still, logic can be a bit too constricting for both agents, thus it is often replaced by looser alternatives such as Knowledge Graphs.\mattialightnote{replace intro sentence to say that it is a subset of logic-like languages where the knowledge base is factual and can be mined}
        they are easily available, enjoy widespread use, can be mined from Natural Language, cover a wide range of domains, and are immediately understandable to human agents, thus providing high understandability and malleability of the artifacts.
        Knowledge Graphs can also be easily turned into Natural Language through a process of \textit{verbalization} in which the Knowledge Graph is encoded into a set of Natural Language claims.
        Improving on Natural Language, they are usually verified by a wide pool of users, hence they make for reliable sources which require minimal to no verification.
        This last property is particularly of interest because it allows the system to scale to more human agents than the ones who are directly designing the system.
        To further strengthen the case for Knowledge Graphs, their integration into machines has been a well-studied problem for several years, and already presents several effective solutions~\cite{DBLP:conf/nips/LewisPPPKGKLYR020}.
        % [CIT] Knowldege Graphs also allow some basic form of reasoning which can be easily extended: they provide a strict reasoning, both commonsense~\cite{DBLP:conf/aaai/HwangBBDSBC21} and factual~\cite{DBLP:journals/cacm/VrandecicK14}, all of which can be integrated in Natural Language~\cite{weber2019nlprolog,yasunaga2021qa}.
        Knowldege Graphs also allow some basic form of reasoning which can be easily extended: they provide a strict reasoning, both commonsense~\cite{DBLP:conf/aaai/HwangBBDSBC21} and factual~\cite{DBLP:journals/cacm/VrandecicK14}, all of which can be integrated in Natural Language.
        % %
        % \begin{figure}[t!]
        %     \centering
        %     \includegraphics[width=0.75\textwidth]{media/hydrant.png}
        %     \caption{An example of a Question Answering hybrid system employing a soft reasoning language and leveraging Knowledge Graphs. Here, the machine consults its artifact bank to retrieve a supporting fact for its prediction, and provides it to the human agent alongside its prediction.}
        %     \label{tbl:hdms4:kg}
        % \end{figure}
        % %
        %
        \begin{figure}[t!]
        \centering
        %\begin{table}
           % \centering
            \begin{tabular}{p{6cm} p{5cm}}
                \raisebox{-.75\height}{\includegraphics[width=0.3\textwidth]{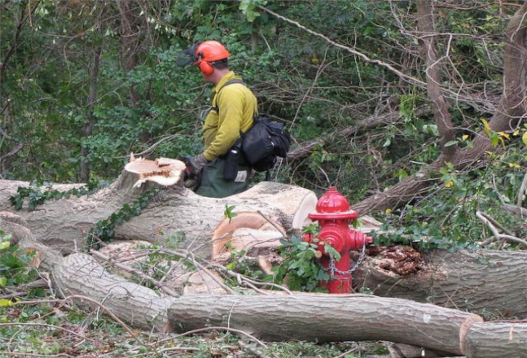}}
                & \textbf{Question:} What can the red object on the ground be used for?
                \newline
                \textbf{Answer:} Firefighting.\newline
                \textbf{Support fact:} Fire hydrants can be used to fight fires.
                \\
            \end{tabular}
            \caption{An example of a Question Answering hybrid system employing a soft reasoning language and leveraging Knowledge Graphs. Here, the machine consults its artifact bank to retrieve a supporting fact for its prediction, and provides it to the human agent alongside its prediction.}
            %\label{tbl:hdms4:kg}
        %\end{table}
        \label{fig:hdms4:kg}
        \end{figure}
        Figure~\ref{fig:hdms4:kg} presents an example from~\cite{wang2017fvqa}.
        Here, the goal is to understand what the purpose of the red object (the hydrant) in the photo is.
        We ought to remark the importance of the context, since even though the machine is aware of the concept of ``fire hydrant'', it is highly unlikely to have seen either a fire hydrant or a fireman in a forest.
        Moreover, since the image itself does not provide any information regarding the purpose of the hydrant, one should expect that the machine could not solve the task without the additional information provided by the human agent in the Knowledge Graph.

      %Evidenziare che in questa sezione si ritorna ad una fase di learning. La spiegazione ha ruolo di mediazione. Ridiciamo perché non si considera il conversational. Dividerlo in più paragrafi: 1 post-hoc expl, 2 interpretable
      \textit{\textbf{Explanation Languages.}} 
      Explanations are designed to \emph{explain} the machine, which makes them highly understandable artifacts out of the box.
    Common families of explanations include counterfactuals, prototypes, feature relevance, and decision rules, the last two leveraged in Learning Together systems.
        Feature relevance provides the human agent with the estimated influence that each input feature has on the prediction of the machine: the higher the relevance, the higher the sensitivity of the machine to changes in that feature.
        Decision rules, instead, provide a descriptive understanding by
        % describing with logic rules
        articulating with logic rules the predictions of the machine.
      In all the following approaches, the key point is that explanations are used to trigger an internal change by machine: the human interacts through the explanations, the machine modifies the model based on the explanations.
      
      \textit{\textbf{Post-hoc Explanations.}}     
        \textcolor{black}{Post-hoc and model-agnostic} explanations are extracted \textit{after} the machine has been trained and regardless of its form.
        Jointly, these two characteristics minimize coupling between the interaction artifacts and the machine, thus granting more flexibility in the design of the machine.
        % Still, explanations are not necessarily designed to be malleable, and even when they are, they may be too complex or brittle to yield a significant benefit to the machine~\cite{}.
        Some more recent proposal propose conversational agents that are able to provide an interface for the human agent to best query the machine for explanations~\cite{MalandriMMN23}.
        On this account, explanations are major artifacts of interest, particularly for systems in which the interaction is
        % a major point of interest
        the focal point, such as
        % Further tightening the link between human and machine,
        Interactive Machine Learning~\cite{DBLP:journals/aim/AmershiCKK14} and eXplainable Interactive Learning (XIL)~\cite{DBLP:conf/aies/TesoK19} systems.
        In particular, the latter allow the human agent to inspect and provide feedback to the machine 
        % through explanation correction
        by correcting its explanation.
        A XIL machine is based on a simple algorithmic kernel comprised of five steps:
        \begin{enumerate}
          \item the machine performs a \textit{learning} step by optimizing its parametrization $\theta$;
          \item the machine generates explanations $A_M$ of (a subset of) its predictions;
          \item the human examines $A_M$, and provides a (optionally) corrected explanation $A_H$;
          \item the machine performs a learning step by optimizing $\theta$ according to the corrected explanation;
          \item if no stopping criterion is met, the machine returns to step (1).
        \end{enumerate}
        By iteratively querying the human, the machine parametrization is thus conditioned on the corrected explanations $A_H$, that is, a XIL machine ends up implementing an estimator of the form:
        % %
        % \begin{equation*}
          $f_{\theta \mid A_H}(X),$
        % \end{equation*}
        % %
        where the correction $A_H$ acts as a proxy for the human knowledge $Z$.

        Integration is directly dependent on the family of explanations.
        By far the most widespread one is feature importance, which assigns a relevance score to each element of the input data $X$..
        The interaction then consists in a possible correction of the relevance scores, with the human agent activating or deactivating each feature according to their judgement.
        Integration follows either a \textit{learning} approach, in which the correction is directly encoded in the machine training objective~\cite{DBLP:conf/ijcai/RossHD17}, or a \textit{generative} one, where the correction is implemented via training on additional synthetic data~\cite{teso2019toward,DBLP:conf/aies/TesoK19}.
        In a learning approach, the corrections of the human agent are encoded in a correction matrix $C \in \{0, 1\}^{n \times m}$ that states what feature relevance have been corrected for each single instance.
        In a generative approach, $C$ is instead used to generate synthetic data $\widetilde{X}$ to further train the machine.
        Features with low relevance have randomized or copied values, while features with high relevance are kept as-is~\cite{DBLP:conf/aies/TesoK19}.
        A small subset of machines are designed to explicitly encode feature relevance in their architecture, thus allowing direct manipulation by human agents~\cite{DBLP:conf/kdd/WangKNSNCVV022}.We may also have a generative approach in which the human agent is tasked with directly creating the additional synthetic data themselves, working on global explanations~\cite{Popordanoska2020}.
        In all the aforementioned approaches, the key point is that explanations are used to trigger a new learning phase for the machine: the human interacts through the explanations, the machine re-trains based on the explanations.

        % While commonly the features presented to the human agent are the very same features provided to the machine, there are some more complex cases in which additional high-level features, such as concepts, are extracted.
        % These features tend to be directly encoded in the agent~\cite{}, in which case the correction simply consists in weight adjustment.
\textcolor{black}{\textit{\textbf{Interpretable explanatory machine learning.}}} A similar approach to interaction through explanations, can be employed with systems that are built to be interpretable by desing.
        % In the latter, we have a learning approach in which the explanation is itself a component of the architecture, thus the correction is again an added component of the machine objective~\cite{DBLP:journals/corr/abs-2012-02898,bontempelli2021toward}.
        For example in~\cite{DBLP:journals/corr/abs-2012-02898}, we have a learning approach in which the explanation is itself a component of the architecture, thus the correction is an added component of the machine objective.
        An emerging approach, mainly aimed at Concept explanations, is the \emph{structured explanation} approach, where explanations are provided as complex structures that the human can act upon.
        ~\cite{DBLP:journals/datamine/BontempelliGPT22} presents an application on concept hierarchies, where concepts are laid on a tree-like hierarchy such that the concept of a parent node, e.g., ``Animal'', is a generalization of the concepts in its children, e.g., ``Dog'' and ``Cat''.
        The machine, based on a k-NN model, is tasked to solve two tasks: a downstream task, and a \emph{concept drift} task, that is, to identify if the relationships within the structure have changed.
        Once detected, the machine presents the concepts of interest to the human, who in turn corrects their structure, e.g., by removing or adding concepts, or by acting on the structure itself, i.e., removing or adding parent-child relationships between concepts.
        The correction is integrated by removal/addition of appropriate instances from the training set of the machine, thus directly impacting the k-NN model.

    \subsubsection{Time of interaction and Cost Of Learning}
      Artifacts are integrated at different times in the lifetime of the machine, hence either at training or at inference time.
      In the former case, the machine integrates the artifact at training time, and cannot be interacted with at inference time.
      In this case, the human agent is effectively providing feedback only on training \cite{teso2019toward,DBLP:conf/aies/TesoK19,DBLP:conf/ijcai/RossHD17,DBLP:conf/kdd/WangKNSNCVV022,Popordanoska2020,DBLP:journals/corr/abs-2012-02898,DBLP:journals/datamine/BontempelliGPT22}.
      In the latter case, the human agent has more control on the machine, and receives and acts upon interaction artifacts at inference time \cite{DBLP:conf/iui/GuoDAMCK22,DBLP:conf/pldi/EllisWNSMHCST21,DBLP:conf/emnlp/MishraTC22,DBLP:conf/emnlp/BostromZCD21,DBLP:conf/emnlp/MurtyMLR22,wang2017fvqa}.
      Machines based on explanation languages, such as XIL, tend to provide training-time interaction, while more recent approaches based on hard or soft reasoning languages tend to provide inference-time interaction.

      A critical distinction in Learning Together systems is the cost of learning, that is, the cost the system has to pay to properly integrate the actions of the human within the machine.
      Systems tend to fall to the two ends of the spectrum.
      In traditional approaches such as XIL, the interaction triggers a costly training step for the machine.
      Here, the cost varies with respect to both the magnitude of the human agent correction and the intrinsic features of the machine \cite{teso2019toward,DBLP:conf/aies/TesoK19,DBLP:conf/ijcai/RossHD17,DBLP:conf/kdd/WangKNSNCVV022,Popordanoska2020,DBLP:journals/corr/abs-2012-02898,DBLP:journals/datamine/BontempelliGPT22}.
      Conversely, more recent approaches \cite{DBLP:conf/iui/GuoDAMCK22,DBLP:conf/emnlp/MishraTC22,DBLP:conf/emnlp/BostromZCD21,DBLP:conf/emnlp/MurtyMLR22,wang2017fvqa}, such as extrinsic artifact-based systems, have no additional cost due to the nature of the machine itself.
      Why then rely on traditional approaches if they incur in an inevitable additional cost?
      Extrinsic artifacts have to be generated in the first place: the cost of populating the artifact bank is as inevitable as the training cost for traditional approaches.
      As previously mentioned, when such banks already exist, e.g., in knowledge graphs, this cost can be greatly reduced.

      % \begin{table}
      %   \begin{tabular}{ @{} l l l @{} }
      %     \toprule
      %     \textbf{Model} & \textbf{Artifact} & \textbf{Interaction time} \\
      %     \midrule
      %     & Decision rule & Learning \\
      %     & Feature relevance & Learning \\
      %     & Feature relevance & Learning \\
      %     & Proof trees & Inference \\
      %     & Proof trees & Inference \\
      %     & Proof-like trees & Inference \\
      %     & Proof-like trees & Inference \\
      %     & Proof-like trees & Inference \\
      %     & Facts & Inference \\
      %     & Facts & Inference \\
      %     \bottomrule
      %   \end{tabular}
      %   \caption{Models enabling interaction }
      %   \label{tbl:hdms4:models}
      % \end{table}

    \subsubsection{Strengths and limitations of Learning Together Systems}
    Unlike Human overseers and Learn to Abstain, Learn Together systems integrate humans and machines in a hybrid, bidirectional system. The agents are tightly coupled and task, domain, data, and user-specific. Machines are limited by the chosen language, meaning \emph{i)} system design requires significant effort, and \emph{ii)} communication does not easily transfer across languages. This heterogeneity hinders progress, with each system offering insights mainly for similar tasks, domains, and interactions. Thus, Learn Together systems improve \emph{vertically}, with unpredictable success.
      % Unlike the Human overseers and Learn to Abstain paradigms, Learn Together systems integrate humans and machine in a fully hybrid system with bidirectional communication.
      % Thus, the two agents are rarely truly decoupled, their design and use being often task, domain, data and user specific.
      % Machines are limited by and tightly coupled with the language of choice, thus \emph{i)} designing such systems requires significant ground-up effort, and \emph{ii)} effective communication in one language does not appear to transfer to other languages. 
      % This extreme heterogeneity partially hinders progress, each system often yielding valuable insight only for future systems operating in similar tasks, domains, data type and desired user interaction.
      % In other words, Learn Together systems tend to improve \emph{vertically}, their success being often unpredictable.
      Currently, Learn Together systems are largely static: they cannot switch languages on demand, adapt to different human agents, or defend against incorrect human feedback. As a recent development in hybrid systems, they also lack tailored validation measures. While traditional validation applies to the system as a whole, little effort has been made to assess how well a machine complies with human corrections or to provide guarantees on the effects of those corrections.
    % At the moment, Learn Together systems are largely static: they are not able to switch language on demand, nor to adapt to different human agents, or defend themselves against possibly incorrect feedback provided by the human agents.
    %   As a relatively recent development in Hybrid systems, they also tend to lack proper measures of validation tailored to them.
    %   While classical validation measures apply to the system as a whole, as a predictive system, there appears to be little effort in properly assessing the compliance of a machine with the corrections provided by the human agent, and provide guarantees on the effects of such corrections.

\section{Literature gaps and concluding remarks}
  \label{sec:gaps}
  Hybrid Systems are a still novel topic, and there are several problems that are open to new research.
  % offer a lot of opportunities for development.
  Following our taxonomy, we identify three categories of open challenges:

      \textit{\textbf{Machine-related problems.}} Unlike Learning Together systems, Human Oversight and Learning to Abstain systems still lack meaningful means of communication to engage with the human agent.
      % \robernote{richiamare il discorso di andrea sui goal del sistema HDM}
      When presented with a prediction to oversee, or with an uncertainty estimate of such prediction, human agents are rarely also presented with suggestions on why the prediction should be accepted or rejected, or why the machine is uncertain of its prediction, let alone how to tackle the uncertainty itself.
      We have highlighted some first steps in tackling this problem in Subsection~\ref{sec:hdms1:explanations}, but this is far from a solved problem.
      
      \textit{\textbf{Human-related problems.}} Both Learning to Defer and Learning Together systems require a considerable human effort, i.e., a high volume of labels or artifacts, to be implemented.
      This is a particularly taxing for the human agent, and while current solutions aim to tackle the problem with a given fixed budget, there are no clear solutions as to how to reduce such high cost. 
      When it comes to Learning to Defer, human-AI ``collaboration'' heavily depends on a global data annotation industry, wherein a vast and often invisible human labor force is engaged with tedious and taxing jobs.
      Yet, the prevailing labor standards for data annotators are mostly characterized by lax regulations, low wages and few legal protections \cite{Sarkar2023}.
      
      \textit{\textbf{Interaction-related problems.}} Related to the interaction itself, there is still little experimentation on understanding what language is best suited to which task, and how to adapt the language to different users within the same Hybrid System.
      In particular, current hybrid systems are \textit{monolingual}, and are developed with one language and one type of human agent in mind, thus lack the ability to adapt to different humans.
      Underlying the development of such systems is the assumption that the human agents in the system and their capabilities and understanding are static, which is rarely the case.
      As a result, hybrid systems lack flexibility and have little ability to adapt to the heterogeneity of the humans they may interact with.
      When Hybrid Systems interact with several users, e.g. in Multiple-Expert Learning to Defer and in Learning Together systems, machine agents are rarely able to seamlessly handle higher multiplicity of agents, and basic concepts such as malicious agents who wish to poison the system, steer it in an undesirable direction, or simply introduce conflicts with existing artifacts are yet to be properly defined and studied.
  %

%\section{Conclusions}
%  \label{sec:conclusion}
\noindent Hybrid Systems, where humans and machines collaborate on predictive tasks, mark a new AI paradigm. Synergistically integrating different agents allows us to leverage their strengths and mitigates their respective weaknesses. This survey introduces a taxonomy of key Hybrid Systems literature, categorized into three paradigms: Human Oversight, Learn to Abstain, and Learn Together. In Human Oversight the human verifies the AI's output, in Learn to Abstain the human's knowledge is represented in the data for AI adaptation, finally in Learn Together, humans actively correct the AI's reasoning. We outlined significant works for each paradigm, discussing strengths and weaknesses, aiming to support future Hybrid Systems research.We hope that this survey can serve as a strong foundation for future research on Hybrid Systems.
  
% \section{Citations and Bibliographies}

\section*{Acknowledgements}
This work has been funded by PNRR - M4C2 - Investimento 1.3, Partenariato Esteso PE00000013 - “FAIR - Future Artificial Intelligence Research” - Spoke 1 “Human-centered AI” and by
ERC-2018-ADG G.A. 834756 “XAI: Science and technology for the eXplanation of AI decision making” and Prot. IR0000013. 
This work was also funded by the European Union under Grant Agreement no. 101120763 - TANGO. Views and opinions expressed are however those of the author(s) only and do not necessarily reflect those of the European Union or the European Health and Digital Executive Agency (HaDEA). Neither the European Union nor the granting authority can be held responsible for them. 
The work has also been realised thanks to NextGenerationEU - National Recovery and Resilience Plan, PNRR) - Project: “SoBigData.it - Strengthening the Italian RI for Social Mining and Big Data Analytics” - Prot. IR000001 3 - Notice n. 3264 of 12/28/2021.

%\bibliographystyle{ACM-Reference-Format}
%\bibliography{references}

\printbibliography[heading=subbibintoc]
\end{refsection}

% %\clearpage
% \begin{refsection}
% %\section*{Appendices}
%   \appendix
%   % \input{appendices}
%   \include{appendices}

% \printbibliography[heading=subbibintoc]
% \end{refsection}

\end{document}

%\title{AI, Meet Human: A Survey on Hybrid Systems }
%\title{AI, Meet Human: A Survey on Hybrid Systems for Decision-Making Processes}
\title{Supplemental Materials.
AI, Meet Human: Learning Paradigms for Hybrid Decision-Making Systems}

\author{Clara Punzi}
\email{clara.punzi@sns.it}
\orcid{??}
\affiliation{%
  \institution{Scuola Normale Superiore}
  \streetaddress{Piazza dei Cavalieri, 7}
  \city{Pisa}
  \country{Italy}
  \postcode{56126}
}

\author{Roberto Pellungrini}
\email{roberto.pellungrini@sns.it}
\orcid{??}
\affiliation{%
  \institution{Scuola Normale Superiore}
  \streetaddress{Piazza dei Cavalieri, 7}
  \city{Pisa}
  \country{Italy}
  \postcode{56126}
}

\author{Mattia Setzu}
\email{mattia.setzu@unipi.it}
\orcid{??}
\affiliation{%
  \institution{University of Pisa}
  \streetaddress{Largo Bruno Pontecorvo, 3}
  \city{Pisa}
  \country{Italy}
  \postcode{56127}
}

\author{Fosca Giannotti}
\email{fosca.giannotti@sns.it}
\orcid{??}
\affiliation{%
  \institution{Scuola Normale Superiore}
  \streetaddress{Piazza dei Cavalieri, 7}
  \city{Pisa}
  \country{Italy}
  \postcode{56126}
}

\author{Dino Pedreschi}
\email{dino.pedreschi@unipi.it}
\orcid{??}
\affiliation{%
  \institution{University of Pisa}
  \streetaddress{Largo Bruno Pontecorvo, 3}
  \city{Pisa}
  \country{Italy}
  \postcode{56127}
}
%% of authors' names for this purpose. in
\renewcommand{\shortauthors}{Punzi, Pellungrini, Setzu, et al.}


% \section{Citations and Bibliographies}

%\section*{Acknowledgments}
  % \begin{acks}
  %   % \begin{verbatim}
  %   %   ...
  %   % \end{verbatim}
  % \end{acks}

%\bibliographystyle{ACM-Reference-Format}
%\bibliography{references}
\maketitle

% %\clearpage
%\section*{Appendices}
  \appendix
  \section{Properties of surrogate loss functions in classification tasks}
\label{sec:surrogate}
  % \foscanote{FOSCA: questa è una sotsottosezione di LTD? valutiamo bene see questa parte possa essere sintetizzata e messa come material ausiliario}  
  Shifting the learning objective from the empirical risk minimization of a loss function to that of a \textit{surrogate} loss function has valuable computational gains. However, a number of statistical properties need to be assessed in order to guarantee the theoretical robustness of optimization results. This section provides a succinct overview of the statistical implications associated with the key conditions that are deemed desirable for a surrogate loss function. 
    \subsection{Pointwise Fisher consistency}
    \label{sec:consistency}
    \textit{Fisher Consistency} (FC) is a desirable property of statistical estimators, which posits that if an estimator were computed using the complete population instead of a sample, it would yield the true value of the estimated parameter \cite{Fisher1922}.
    When the estimation procedure is the minimization of the risk associated to a loss function, FC is a necessary condition for reasonable performance, since it guarantees that the loss represents the correct objective function \cite{Lin2004}. Furthermore, consistency results ensure that the optimization of a surrogate function does not hinder the search for a function that attains the Bayes risk. Consequently, it provides a theoretical foundation for the use of computationally efficient techniques that are specifically suited to convex functions, such as surrogate losses, and not to discrete losses \cite{Bartlett2006}.
   % The concept of \textit{pointwise FC} has been introduced by Bartlett et al. \cite{Bartlett2006} as the weakest possible condition that a surrogate loss function $\widetilde{\mathscr{L}}$ should satisfy to give non-trivial upper bounds on the excess risk (i.e., the quantity $\mathscr{R}(f)-\mathscr{R}^*$) in terms of excess $\widetilde{\mathscr{L}}$-risk (i.e., the quantity $\mathscr{R}_{\widetilde{\mathscr{L}}}(f)-\mathscr{R}_{\widetilde{\mathscr{L}}}^*$). 
    %In other words, the property of pointwise FC is the minimal requirement a surrogate loss function should satisfy so that the minimization of the $\widetilde{\mathscr{L}}$-risk may provide a reasonable surrogate that minimizes the expected risk.  
    %Informally, a surrogate loss function is said to be \textit{Fisher consistent} if its optimization over all measurable functions results in a classifier whose accuracy is close to that of the Bayes optimal classifier. 
    In the context of Learning to Abstain, the formal definition FC with respect to surrogate losses follows the one stated for $K$-class classification, where the optimization objective is to minimize the average loss function $\mathscr{L}(y, g_1(x),\ldots,g_k(x))$ in order to learn the real-valued scoring function $g_i$, for $i\in\{1,\ldots,K\}$. These are then used to classify the samples as $\hat{y} = \arg\max_i g_i(x)$ \cite{Long2013}. Formally:

    \begin{definition}\label{def:consistency}[Fisher Consistency in L2D \cite{Mozannar2023,Verma22}] A surrogate loss $\widetilde{\mathscr{L}}$ is a consistent loss function with respect to another loss $\mathscr{L}$ if minimizing the surrogate loss over all measurable scoring function is equivalent to minimizing the original loss.
    \end{definition}

    The existence of classes of loss functions, whether convex or not, that can attain FC has been proven for many learning problems, including binary classification \cite{Lin2004} and multiclass classification \cite{Liu2007,Tewari2007,Zou2008}. To date, in the L2D setting the surrogate loss functions that meet the Fisher consistency criteria are only the two proposed in \cite{Mozannar20,Verma22}.
    For a more complete treatment of Fisher consistency the reader can refer to \cite{Fisher1922,Gerow1989} and for its application in the context of surrogate loss function to \cite{Bartlett2006,Lin2004,Neykov2016,Steinwart2007,Tewari2007}.  Note that this notion of consistency has been also called \textit{classification-calibration} \cite{Bartlett2006} or simply \textit{consistency} throughout the L2D literature. However, it should not be confused with other concepts commonly referred to with the same name, such as the notion of \textit{asymptotic consistency}, which instead refers to the property that an estimator has if it converges in probability to the true value.
    
    \subsection{Confidence calibration}
    \label{sec:calibration}
    The notion of \emph{confidence calibration} refers to the property of an estimator (e.g., a probabilistic classifier) to generate a predictive distribution that is consistent with the empirical frequencies observed from realized outcomes \cite{Dawid1982,Verma22,Vaicenavicius2019}.
    %accurately reflects the true predictive uncertainty 
    %(e.g., the empirical performance of the machine and human agent in the Learning to Abstain setting) . 
    That is, a calibrated estimator correctly quantifies the level of uncertainty or confidence associated with its predictions \cite{SilvaFilho2023}.
    For instance, in the case of a HS, if a human decision-makers has a 80\% probability of being correct, then the system should forecast that the human agent will be correct around 80\% of the time.
    In the context of decision-making, 
    %hybrid decision-making systems
    this property is desirable as it guarantees that the outcomes of a model can be interpreted as real-world probabilities (i.e., the true uncertainties of both human and machine), thus promoting the trustworthiness of the system \cite{Vaicenavicius2019}. 
    When considering a HS, such as a Learning to Abstain model, a properly calibrated machine loss is not enough to ensure optimal performance, as a faulty deferral policy can severely and negatively impact the system performance. Indeed, while the calibration of the underlying classifier can be accomplished by employing post-hoc techniques such as temperature scaling to the classifier sub-component of the optimization objective function \cite{Kull2019,SilvaFilho2023}, it is equally necessary to calibrate the correctness of the human agent. This entails ensuring that the estimated conditional probability, as determined by the deferral function, of the $i^{\text{th}}$ human agent being correct on a specific input aligns with their actual probability of being correct on similar inputs. Formally:

    \begin{definition}[Confidence calibration \cite{Verma2022Multi,Verma22}]
    Let $t:\mathcal{X}\rightarrow(0,1)$ be an estimator of the correctness of a human $H$, i.e., $t(\cdot)=\mathbb{P}[Y_H=Y \mid X=\cdot]$. Then $t$ is said to be \textit{calibrated} if, for any confidence level $c\in(0,1)$, the actual proportion of times $H$ is correct is equal to $c$:
    $$\mathbb{P}[Y_{H,i}=Y_i|t(X_i)=c]=c \;\;\; \forall c\in(0,1) \;\forall i\in\{1,\ldots, n\}.$$
    Moreover, the level of calibration of $t$ can be computed through the \textit{Expected Calibration Error} (ECE), which is defined as follows:
    $$ \text{ECE}(t)=\mathbb{E}_X\left[\mathbb{P}(Y_H=Y \mid t(X)=c)-c \right].$$
    \end{definition}

    %As already pointed out in \cite{Verma2022Multi,Verma22} and shown in \cite{Vaicenavicius2019}, since the problem of expert correctness can be reduced to binary classification, there is no need to distinguish between \textit{distribution calibration, confidence calibration}, and \textit{classwise calibration}, as they all match the definition given above.
    The explicit formulation of $t$ has been derived in \cite{Verma22} for the case of Single-Expert L2D, while it has been derived in \cite{Verma2022Multi} for the case of Multiple-Expert L2D. 
    %Specifically, the learning problem is modelled as a multiclass classification task with an augmented label space $\mathcal{Y}^\perp=\mathcal{Y}\cup\{\perp\}$, with the class $\perp$ corresponding to the action of deferral. For each class index $y\in\mathcal{Y}$, let $g_y:\mathcal{X}\rightarrow\mathbb{R}$ be scoring functions such that $f(x)=\arg\max_{y\in\mathcal{Y}}g_y$, and, similarly, let $g_\perp:\mathcal{X}\rightarrow\mathbb{R}$ such that $\rho_M(x)=\mathbbm{1}_{\max_{y\in\mathcal{Y}}\leq g_\perp}$. The probability that an expert is correct is then given by $\mathbb{P}[Y_H=Y \mid X]$
    
    \subsection{Realizable consistency}
    The notion of consistency of a surrogate loss with respect to another loss introduced in Definition \ref{def:consistency} assumes the optimization to be performed over the entire set of measurable functions. However, in realistic scenarios, the minimization will be very likely carried out only on restricted classes of function, resulting in near-optimal performance whenever the Bayes predictor does not belong to the selected classes.
    %it is reasonable to expect that restricting the hypothesis set may result in a significant improvement in performance. 
    In the context of multiclass classification, this issue has been observed and discussed by \citet{Long2013}, who firstly introduced the terminology of $\mathcal{F}$-\textit{realizable consistency} when referring to consistency with respect to a restricted class $\mathcal{F}$ of scoring functions. In particular, the empirical findings in \cite{Long2013} indicate that the property of $\mathcal{F}$-realizable consistency holds greater significance and applicability compared to FC in scenarios where learning algorithms are constrained to a specific class $\mathcal{F}$ of scoring functions, as it is more closely related to classification accuracy. 
    Successively, the same concept has been recalled and adapted to the frameworks of L2R \cite{cortes2016l2r} and L2D \cite{Mozannar20,Mozannar2023}. Notably, in these settings the hypothesis spaces of both the classifier (i.e., $\mathcal{F}_M$) and rejector (i.e., $\Rho_M$) are taken into consideration.

    \begin{definition}
        [realizable $(\mathcal{F}_M,\Rho_M)$-consistency \cite{cortes2016l2r,Mozannar2023,Mozannar20}]
        Let $\widetilde{\mathscr{L}}$ be a surrogate loss function of the loss $\mathscr{L}$. 
        Then $\widetilde{\mathscr{L}}$ is said to be realizable $(\mathcal{F}_M,\Rho_M)$-consistent if, for all distributions \textbf{P} over $X\times Y_M$ and any human agent $H$ for which there exist $f^*\in\mathcal{F}_M$ and $\rho_M^*\in\Rho_M$ such that $\mathscr{L}(f^*,\rho_M^*)=0$, 
        then for any $\epsilon>0$ there exists $\delta>0$ such that if $(\hat{f}, \hat{\rho}_M)$ satisfies:
        $$ \mid \widetilde{\mathscr{L}}(\hat{f}, \hat{\rho}_M) - \inf_{(f,\rho_M)\in(\mathcal{F}_M,\Rho_M)} \widetilde{\mathscr{L}}(f, \rho_M) \mid \leq \delta \quad \Rightarrow \quad \mathscr{L}(\hat{f}, \hat{\rho_M})\leq \epsilon.$$
        %optimizing $\widetilde{\mathscr{L}}$ returns the zero error solution:
        %$$\widetilde{f},\widetilde{\rho}_M\in \underset{{(f,\rho_M)\in\mathcal{F}_M}\times\mathcal{D}}{\arg\inf}\widetilde{L}(Y,Y_H,f(X),\rho_M(X)) \Rightarrow \mathscr{L}_{\text{defer}}(Y,Y_H,\widetilde{f}(X),\widetilde{\rho}_M(X))=0$$
    \end{definition}
    
    % The formalization of $(\mathcal{F}_M,\Rho_M)$-realizable consistency can be split in two parts, namely, the definition of a realizable setting and that of a realizable $(\mathcal{F}_M,\Rho_M)$-consistent surrogate loss:
    % \begin{definition}
    %     [Realizable setting w.r.t. $(\mathcal{F}_M,\mathcal{D})$, \cite{Mozannar20,Mozannar2023}] 
    %     Denote with $\mathcal{F}_M$ and $\mathcal{D}$ the classifier and deferral function hypothesis spaces, respectively. Then, a  setting in L2D is said to be \textit{realizable} in $(\mathcal{F}_M,\mathcal{D})$ if there exists a classifier $f\in\mathcal{F}_M$ and a deferrer $\rho\in\mathcal{D}$ that achieve zero system loss $\mathscr{L}_\text{defer}$. 
    % \end{definition}
    For instance, \citet{Mozannar2023} propose a differentiable and realizable $(\mathcal{F}_M,\Rho_M)$-consistent surrogate loss for the L2D setting suitable for hypothesis classes $\mathcal{F}_M$ and $\Rho_M$ of scoring functions that are close under scaling\footnote{A class $\mathcal{F}$ of scoring functions is closed under scaling if $f\in\mathcal{F}\Rightarrow \alpha f\in\mathcal{F}\;\forall\alpha\in\mathbb{R}$.}, e.g., linear functions and feedforward neural networks.

\printbibliography[heading=subbibintoc]